\documentclass[twoside]{article}

\usepackage[accepted]{aistats2022}
\usepackage[round]{natbib}

\usepackage{hyperref}       
\usepackage{url}            
\usepackage{booktabs}       
\usepackage{nicefrac}       
\usepackage{microtype}      
\usepackage{xcolor}         
\usepackage{graphicx}
\usepackage{subcaption}
\usepackage{float}
\usepackage{pgfplots}
\usepgfplotslibrary{fillbetween}
\usetikzlibrary{arrows}
\usetikzlibrary{backgrounds}
\usetikzlibrary{positioning}
\usepackage{amsmath}
\usepackage{amssymb}
\usepackage{mathtools}
\usepackage{bbm}
\usepackage{dsfont}

\newcommand{\bd}[1]{\boldsymbol{\mathrm{#1}}}
\newcommand{\R}{\mathbb{R}}

\newcommand{\normal}{\mathcal{N}}
\newcommand{\expect}{\mathbb{E}}

\newcommand{\res}[2]{$#1$ \small $\pm #2$}
\definecolor{highlightBlue}{RGB}{58,146,222}
\definecolor{highlightOrange}{RGB}{243,165,54}
\definecolor{highlightRed}{RGB}{208,2,27}
\definecolor{highlightGreen}{RGB}{65,117,5}
\definecolor{highlightPurple}{RGB}{144,19,254}
\definecolor{darkGray}{RGB}{50,50,50}
\definecolor{lightGray}{RGB}{200,200,200}

\begin{document}

%

%
\runningauthor{Felix L. Opolka, Yin-Cong Zhi, Pietro Li\`o, Xiaowen Dong}

\twocolumn[

\aistatstitle{Adaptive Gaussian Processes on Graphs via Spectral Graph Wavelets}

\aistatsauthor{ Felix L. Opolka$^*$ \And Yin-Cong Zhi$^*$ \And  Pietro Li\`o \And Xiaowen Dong }

\aistatsaddress{ University of Cambridge \\\& Invenia Labs \And University of Oxford \And University of Cambridge \And University of Oxford } ]

\begin{abstract}
Graph-based models require aggregating information in the graph from neighbourhoods of different sizes. In particular, when the data exhibit varying levels of smoothness on the graph, a multi-scale approach is required to capture the relevant information. In this work, we propose a Gaussian process model using spectral graph wavelets, which can naturally aggregate neighbourhood information at different scales. Through maximum likelihood optimisation of the model hyperparameters, the wavelets automatically adapt to the different frequencies in the data, and as a result our model goes beyond capturing low frequency information. 
We achieve scalability to larger graphs by using a spectrum-adaptive polynomial approximation of the filter function, which is designed to yield a low approximation error in dense areas of the graph spectrum.
Synthetic and real-world experiments demonstrate the ability of our model to infer scales accurately and produce competitive performances against state-of-the-art models in graph-based learning tasks.

\end{abstract}

\section{INTRODUCTION}

Many modern day data sets come in the form of graphs or networks, such as social networks, brain graphs, and protein-interaction networks, where additional information is represented by connective structures between data points. While node features on their own can be used by a variety of machine learning algorithms, the graph structure can often crucially enrich the model further. With the rise of graph signal processing \citep{shuman2012emerging}, graph neural networks \citep{wu2019comprehensive}, and geometric deep learning \citep{bronstein2021geometric}, there is now a rich library of tools to build models for graph structured data, making it possible to tackle a range of complex graph-based modelling tasks.

In particular, Gaussian process (GP) models are popular tools for taking into account the probabilistic nature of the data. A GP model on graphs would allow for modelling uncertainty associated with the nodes in the graph and making predictions on unlabelled nodes. A key requirement in building GPs on graphs is incorporating the graph information into the design of the GP kernel, for example using convolution-like operations~\citep{Ng18,Walker19,Opolka20,Li20} or following the separable kernel design of multi-output GPs~\citep{Venkitaraman20,Zhi20}.

\begin{figure*}[t]
\centering
\begin{subfigure}[b]{0.33\linewidth}
\includegraphics[width=\linewidth, height = 3.8cm]{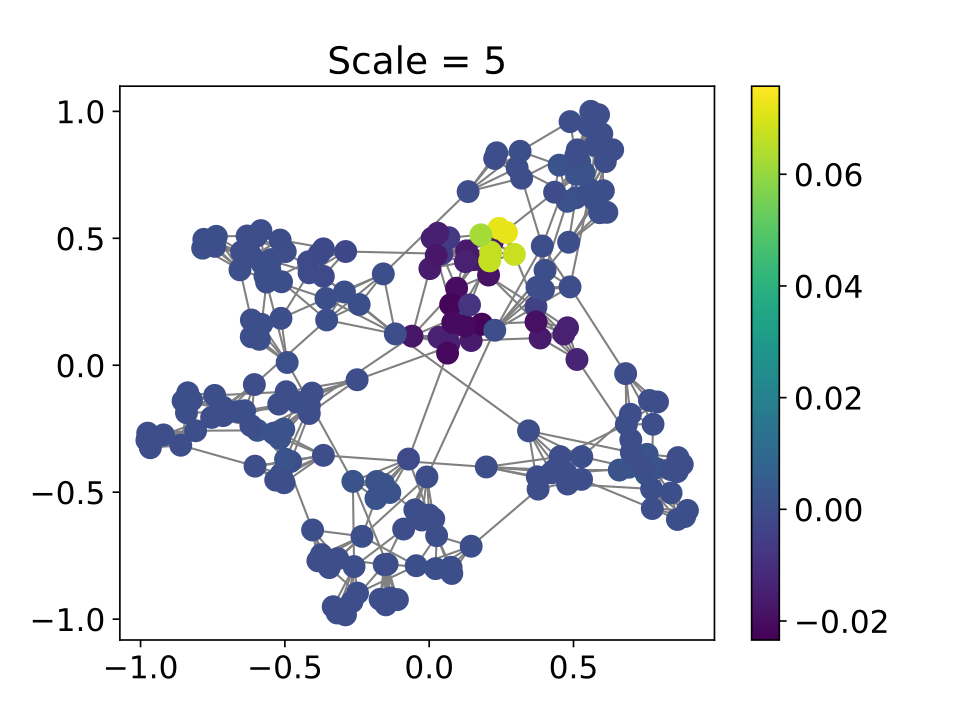}
\caption{}\label{wave_illus1}
\end{subfigure}
\hspace{-3mm}
\begin{subfigure}[b]{0.33\linewidth}
\includegraphics[width=\linewidth, height = 3.8cm]{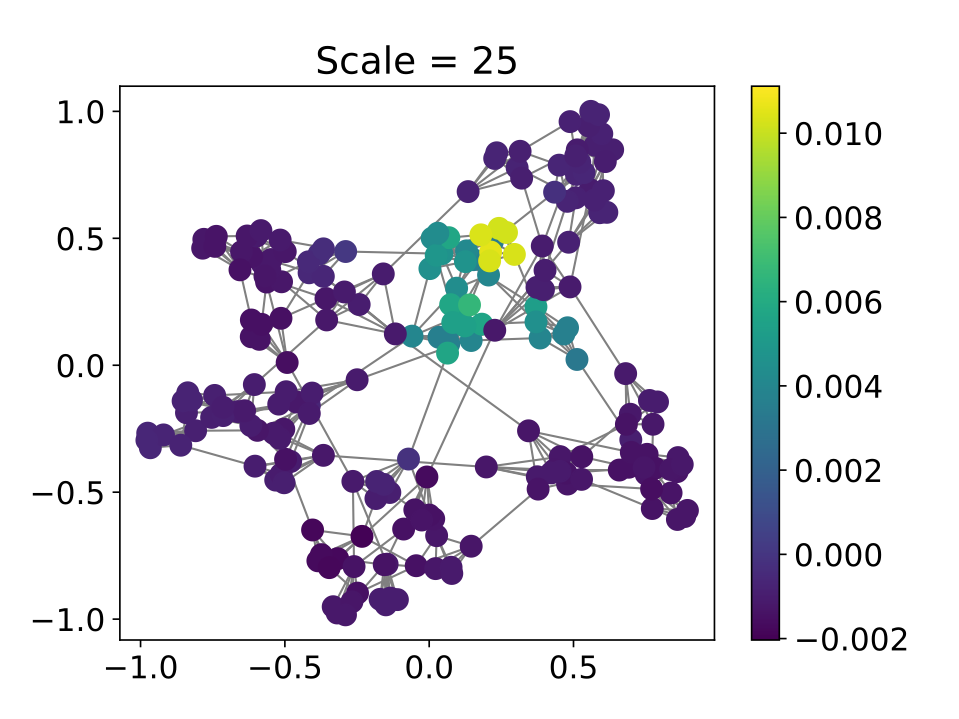}
\caption{}\label{wave_illus2}
\end{subfigure}
\hspace{-3mm}
\begin{subfigure}[b]{0.33\linewidth}
\includegraphics[width=\linewidth, height = 3.8cm]{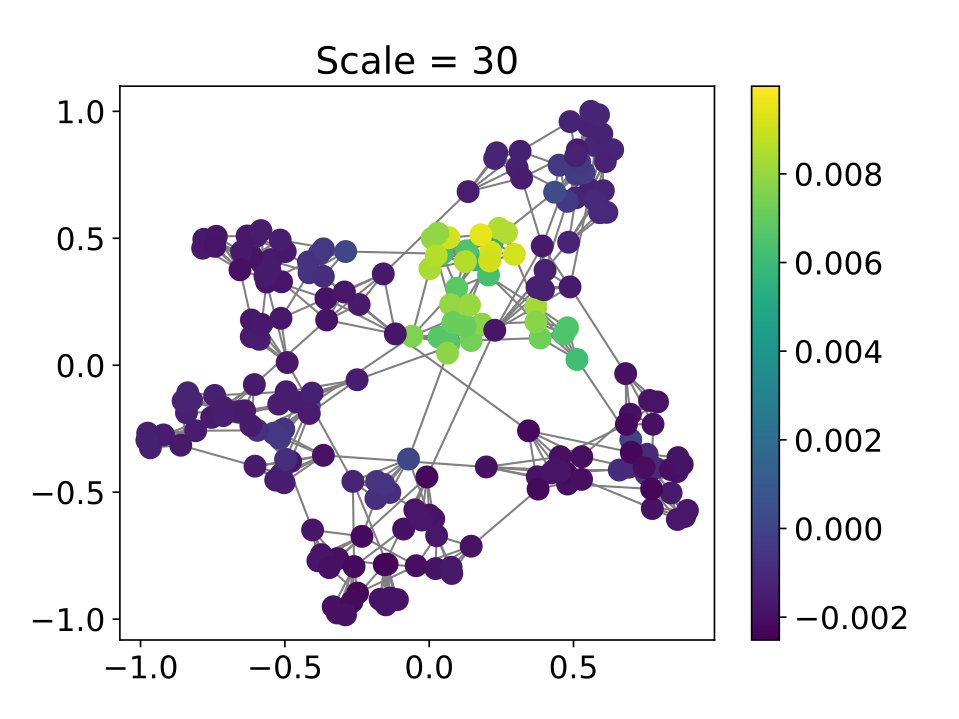}
\caption{}\label{wave_illus3}
\end{subfigure}
\caption{The Mexican Hat wavelet transform of a $\delta$ signal on the focal node. With different scales, the wavelet is able to capture different neighbourhood information weighted in a continuous manner.}\label{wavelet_transforms}
\end{figure*}

The core consideration when incorporating graph structure into a model design is how much neighbours at varying distances should influence the prediction at a certain node. Early spectral approaches rely on the Fourier basis when designing graph-based operators~\citep{bruna2014graphcnn,defferrard2016chebnet}, which is fully localised in the frequency domain but not in the spatial domain, 
hence requiring a polynomial approximation of the graph Laplacian to enforce spatial localisation. 
We instead propose an approach using wavelets, which offer a natural way of trading off between spectral and spatial resolution---and thus localisation---in both domains. The degree of spatial localisation is implicitly controlled by a single wavelet scale parameter defined in the spectral domain (visualised in Figure~\ref{wavelet_transforms}), which makes graph wavelets a natural tool to enable a more flexible notion of neighbourhood of varying size. Moreover, the single scale parameter enables the model to adjust the effective neighbourhood sizes to the properties of the data when incorporated into a model that allows learning hyperparameters, such as a GP.

\begin{figure*}
    \centering
    \begin{tikzpicture}
    \node[anchor=west] at (0.0, 0.0) {\includegraphics[width=0.45\textwidth]{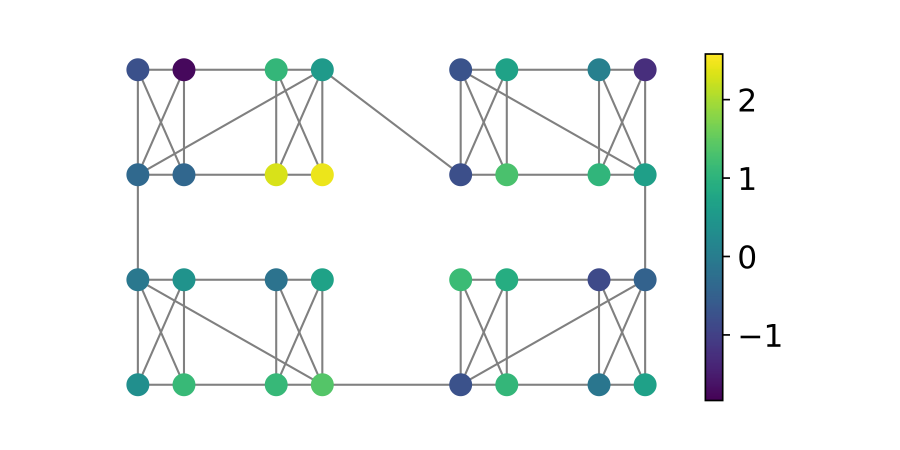}};
    \node at (0.25\textwidth, -2.25) {(a) full multi-scale signal};
    
    \node[anchor=east] at (\textwidth, 0.0) {\includegraphics[width=0.45\textwidth]{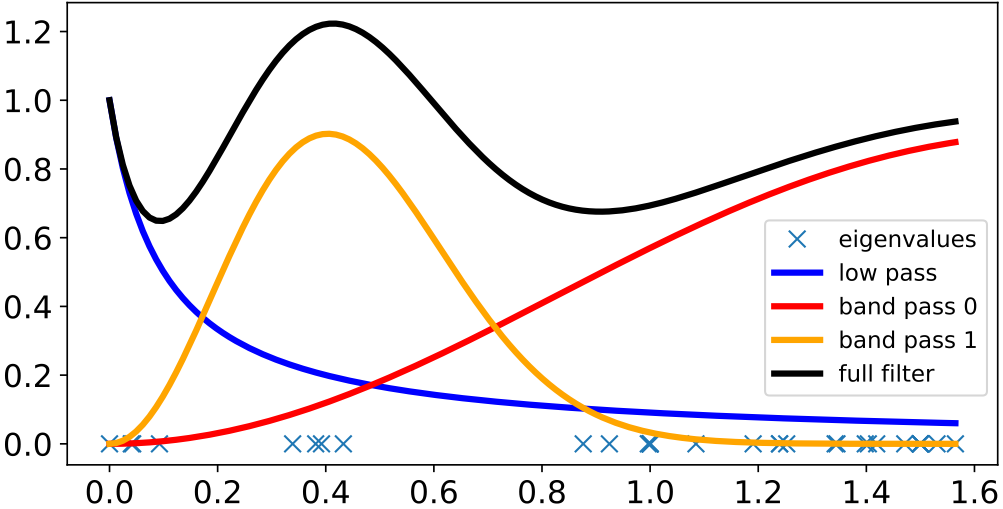}};
    \node at (0.75\textwidth, -2.25) {(b) signal spectrum};
    
    \node[anchor=west] at (0.0, -4.25) {\includegraphics[width=0.275\textwidth]{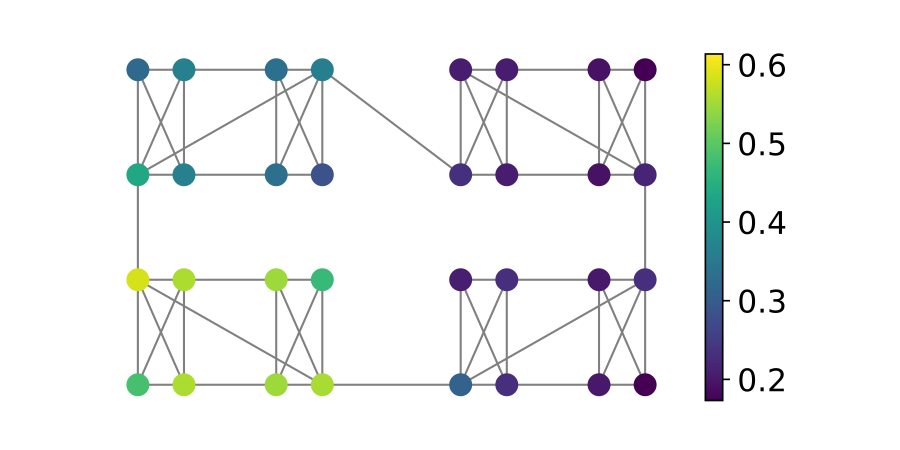}};
    \node at (0.15\textwidth, -5.75) {(c) low-pass filtered signal};
    
    \node at (0.5\textwidth, -4.25) {\includegraphics[width=0.3\textwidth]{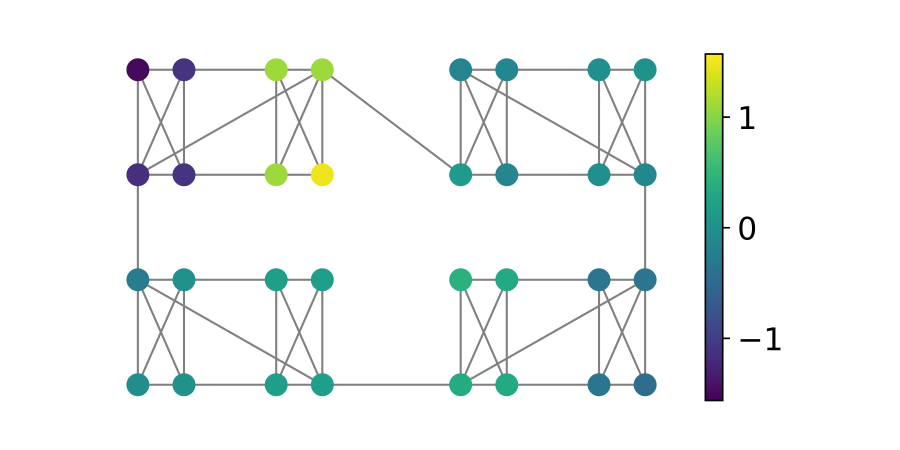}};
    \node at (0.5\textwidth, -5.75) {(d) band-pass filtered signal};
    
    \node[anchor=east] at (\textwidth, -4.25) {\includegraphics[width=0.3\textwidth]{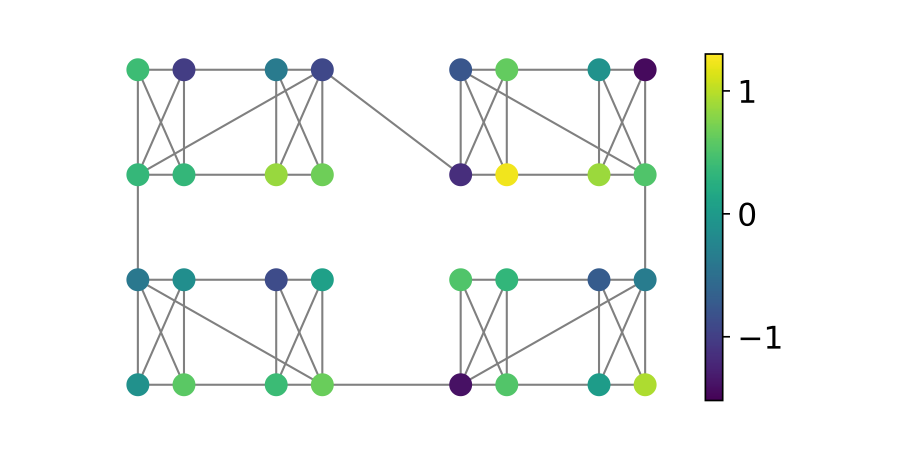}};
    \node at (0.85\textwidth, -5.75) {(e) high-pass filtered signal};
    \end{tikzpicture}
    
    \caption{Visualisation of how wavelet filters can be used to capture multi-scale properties in both the graph structure and the graph signal. Figure (a) shows a graph with two levels of clusters (4-node clusters and 8-node clusters). These clusters are reflected in the gaps (around $0.2$ and $0.6$) in the spectrum of the graph in Figure (b). The signal is obtained by filtering a random signal with the filter in (b), purposefully highlighting the three eigenvalue clusters. Figures (c) - (e) show how the full signal from (a) decomposes into the three filter components. As expected, the low-pass signal varies mostly on the highest cluster level (between 8-node clusters), the band-pass signal mostly on the second cluster level (between 4-node clusters), and the high-pass signal from node to node.}
    \label{fig:overview}
\end{figure*}
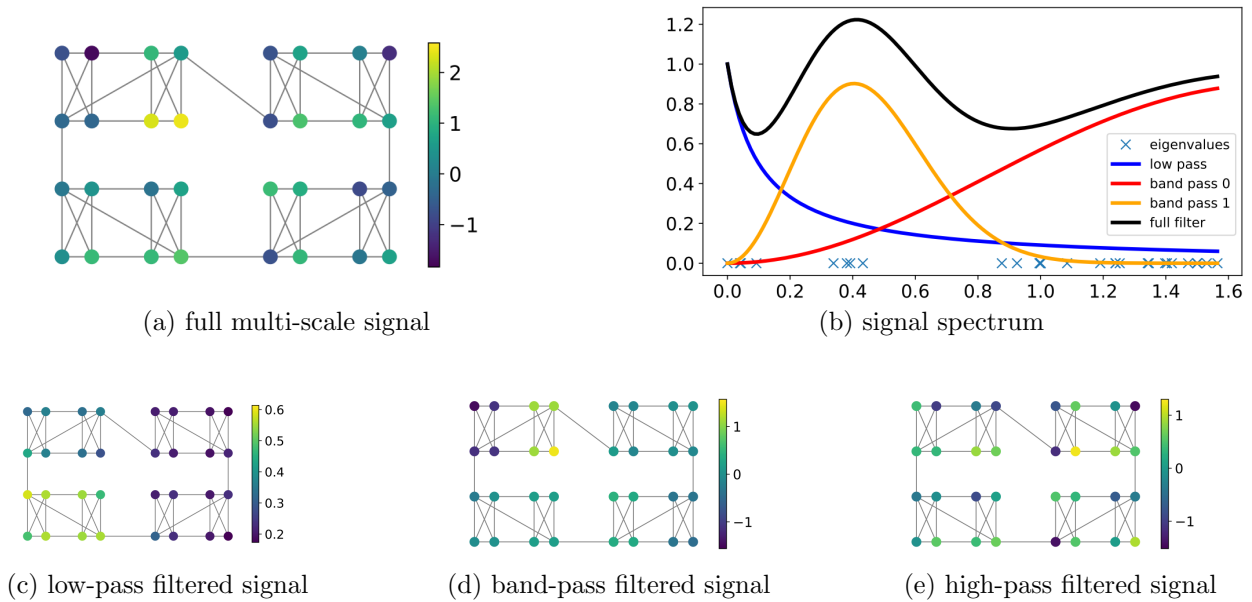

Beyond flexible control of neighbourhood size, using wavelets allows combining filters of different scales. Real-world networks such as connectivity patterns in the brain or metabolic or social interactions networks often exhibit such multi-scale community structure~\citep{ravasz2003hierarchical,clauset2008hierarchical,dutkowski2013gene}, where sets of densely connected nodes in turn form densely connected communities (see Figure~\ref{fig:overview}a for a synthetic example).
These graphs often naturally form the domain of multi-scale signals, which can be modelled through wavelets by combining filters of multiple scales. Figure~\ref{fig:overview}b shows an example of how low-pass and band-pass filters are combined into a more complex wavelet filter, which then captures signal components varying at different scales (Figures~\ref{fig:overview}c-e).

In this work, we introduce a novel graph GP model that uses spectral graph wavelets to incorporate graph structure into the GP kernel. Building on the convenient properties of the wavelet transform, the wavelet graph GP can naturally model continuous neighbourhoods of varying sizes and by extension multi-scale graph signals. The kernel filters are learnable such that their responses can adapt to the observed graph and data.
To bypass the expensive eigen-decomposition of the graph Laplacian, we develop a fast approximation to the wavelet-based filtering, which still allows us to directly optimise the wavelet scales and reduces the approximation error on parts of the spectrum with most eigenvalues. We show that our approximation is more suitable for wavelet filters than the Chebyshev polynomial approximation commonly used for existing low-pass filtering approaches. Through experiments, we demonstrate accurate recovery of scales on a synthetic graph and evaluate our model on benchmark data sets, showing model performance is competitive against state-of-the-art graph-based models.

\section{PRELIMINARIES}
\label{sec:preliminaries}

\paragraph{Gaussian Processes.}

Consider data of the form $(\mathbf{X}, \mathbf{y})$ where we have inputs $\mathbf{X} = [\mathbf{x}_1, \dots, \mathbf{x}_N]^\top$ and labels $\mathbf{y} \in \mathbb{R}^N$, a GP $f$ is defined as
\begin{align}
    f(\mathbf{x}) \sim \mathcal{GP}\big(m(\mathbf{x}), \mathcal{K}_\theta(\mathbf{x},\mathbf{x'})\big)
\end{align}
for any inputs $\mathbf{x}, \mathbf{x'}$, where $m(\cdot)$ is the mean function, and $\mathcal{K}_\theta(\cdot,\cdot)$ is the symmetric and positive definite kernel function. GPs are Bayesian regression models known for the ability to incorporate prior information and having a closed form solution in computing the posterior. When predicting a new data point, the model provides both point predictions and confidence intervals. In addition, GPs provide a marginal log-likelihood on the training data, and in optimising this likelihood we can find the optimal hyperparameters based on the data.

The limitations of GPs are in the inference step where two problems can arise. If a non-Gaussian likelihood is assumed on the data, for classification tasks for example, then the posterior will be analytically intractable. The inference step will also be problematic if the number of data $N$ becomes large as it requires an expensive $\mathcal{O}(N^3)$ matrix inversion. Both problems can be addressed by approximating the posterior through a variational approach, here, a set of inducing points $\mathbf{Z} = [\mathbf{z}_1, \dots, \mathbf{z}_M]^\top$ are introduced and form the inducing random variables $\mathbf{u} = [f(\mathbf{z}_1), \dots, f(\mathbf{z}_M)]^\top$ that is a subset of the GP $f(\mathbf{x})$. Assuming the GP prior of $\mathbb{P}(\mathbf{u}) \sim \mathcal{N}(0, \mathbf{K_{zz}})$ where $[\mathbf{K_{zz}}]_{ij} = \mathcal{K}_\theta(\mathbf{z}_i, \mathbf{z}_j)$, the conditional GP has the following distribution
\begin{align}
    f(\mathbf{x}) | \mathbf{u} \sim \mathcal{GP}(\mathbf{k^\top_{zx}} \mathbf{K^{-1}_{zz}}\mathbf{u} \;, \; \mathcal{K}(\mathbf{x}, \mathbf{x}) - \mathbf{k^\top_{zx}} \mathbf{K^{-1}_{zz}} \mathbf{k_{zx}})
\end{align}
where $\mathbf{k_{zx}}$ are the cross covariances $[\mathcal{K}(\mathbf{z}_1, \mathbf{x}), \dots, \mathcal{K}(\mathbf{z}_M,\mathbf{x})]^\top$. The variational posterior distribution $q(\mathbf{u})$ is assumed to be a multivariate Gaussian with mean $\mathbf{m}$ and covariance matrix $\mathbf{S}$ to be found through maximising the Evidence Lower Bound (ELBO)
\begin{align}
    \mathcal{L}(\theta, \mathbf{Z}, \mathbf{m}, \mathbf{S}) = &\sum_{n=1}^N \mathbb{E}_{q(f(\mathbf{x}_n))} [\log \mathbb{P}(y_n | f(\mathbf{x}_n))]\\
    &- \text{KL}[q(\mathbf{u} || \mathbb{P}(\mathbf{u})]. \label{elbo}
\end{align}
Typically the parameters are optimised via stochastic gradient descent. We refer readers to \citep{rasmussen2005gp} for a more comprehensive overview.

\paragraph{Spectral Filtering and Wavelets on Graphs.}

We refer to the filtering of a signal as the process of highlighting specific frequency components in the signal while de-emphasising others with the aim of obtaining a function more suitable for the prediction task.
Let $\mathcal{G} = (\mathcal{V}, \mathbf{A})$ be a graph with vertex set $\mathcal{V} = \{v_1, \dots, v_N \}$ and adjacency matrix $\mathbf{A}$, we define the notion of spectral filtering on graphs \citep{shuman2012emerging} based on the graph Laplacian defined as $\mathbf{L} = \mathbf{D} - \mathbf{A}$, where $\mathbf{D}$ is the diagonal degree matrix. Additionally, the commonly used normalised graph Laplacian is computed as $\tilde{\mathbf{L}} = \mathbf{D}^{-\frac{1}{2}} \mathbf{L} \mathbf{D}^{-\frac{1}{2}}$.
This is often preferred due to the boundedness of its eigenvalues to the interval $[0, 2]$ and the scaling of the graph edge weights \citep{shuman2012emerging}, hence our model will make use of this normalised version throughout.

Assuming that $\mathcal{G}$ is undirected, the Laplacian is symmetric and admits the eigen-decomposition $\tilde{\mathbf{L}} = \mathbf{U\Lambda U}^\top$ where $\mathbf{U}$ contains the eigenvectors and $\mathbf{\Lambda}$ is the diagonal matrix of eigenvalues. For any function $\mathbf{f}$ on $\mathcal{G}$ (or graph signal), the graph Fourier transform is defined as $\mathbf{U}^\top \mathbf{f}$ and computes the amplitude of each eigenvector in the function $\mathbf{f}$. Filtering on graphs is then achieved in the graph spectral domain by applying a function on the eigenvalues as $g(\mathbf{\Lambda})$, and we write $\hat{\mathbf{f}} = \mathbf{U}g(\mathbf{\Lambda}) \mathbf{U}^\top \mathbf{f}$ as the filtered signal (or function), and $\mathbf{U}g(\mathbf{\Lambda}) \mathbf{U}^\top$ is referred to as the graph filtering matrix.

The graph Fourier transform $\mathbf{U}$ is localised in the graph spectral domain as each eigenvector only contributes a single frequency to the construction of $\mathbf{f}$. However, they are not localised in space as each eigenvector of $\mathbf{f}$ is on the entire spatial domain. Wavelet transform addresses this issue by decomposing a function $\mathbf{f}$ into a linear combination of basis function 
that are both localised in space and frequency. 
The definition of graph wavelets is derived from spectral graph theory by \citet{hammond2011wavelets} and will form the basis of the wavelets we utilise. The transform is an operator function of the graph Laplacian determined by a function $g$ as follows: $b_\beta(\tilde{\mathbf{L}}) = \mathbf{U}g(\beta\mathbf{\Lambda}) \mathbf{U}^\top$.
The function $g$ is applied in the graph spectral domain, but spatially it will also be localised if chosen from the library of mother wavelets. The scale parameter $\beta$ then plays the role of controlling the localisation of the transform. We make use of the Mexican Hat wavelet, which we will present later on along with our model formulation. The spatial localisation can be demonstrated by applying the wavelet transform to an impulse signal on the graph $b_\beta(\tilde{\mathbf{L}}) \delta_n$, where $\delta_n = 1$ at node $n$ and 0 elsewhere. This is presented in Figure \ref{wavelet_transforms} where the various scales $\beta$ lead to different proximity of neighbourhoods. For each scale, the different hop neighbourhoods are also weighted in a continuous manner that decays to 0 once far enough away from the centre node. This allows the aggregation to happen in a non-linear manner to extract additional information for each node.

\section{METHODOLOGY}
\paragraph{Graph wavelet GP.}

We describe a Gaussian process model for the task of semi-supervised node-level prediction on a graph $\mathcal{G} = (\mathcal{V}, \bd{A})$ with $N$ nodes.
The nodes of the graph are commonly associated with a set of features $\{\bd{x}_1, \ldots, \bd{x}_N\}$, which form the feature matrix $\bd{X} \in \R^{N \times K}$.
As we have seen in Section~\ref{sec:preliminaries}, given a graph signal $\bd{f} \in \R^{N}$ on the graph domain, we can apply a wavelet filter $g_\theta(\lambda)$ with scale parameters $\theta$ as follows:
    \begin{equation}
    \hat{\bd{f}} = \bd{U} g_\theta(\bd{\Lambda}) \bd{U}^\top \bd{f},\label{eq:wavelet_filtering}
    \end{equation}
where $\bd{U}$ and $\bd{\Lambda}$ are the eigenvectors and eigenvalues of the graph Laplacian of $\mathcal{G}$ such that $\tilde{\bd{L}} = \bd{U} \bd{\Lambda} \bd{U}^\top \in \R^{N \times N}$ and $\bd{\Lambda}$ is a diagonal matrix. The wavelet filter $g_{\theta}$ is applied element-wise to $\Lambda$. For brevity, we define $\bd{W}_\theta \coloneqq \bd{U} g_\theta(\bd{\Lambda}) \bd{U}^\top$ and refer to it as the \textit{wavelet filter matrix}.

For the sake of conducting Bayesian inference, we assign a Gaussian process prior to the function $\bd{f}$
    \begin{equation}
    \bd{f} \sim \mathcal{GP}\left(m(\bd{x}), \mathcal{K}_{\psi}(\bd{x}, \bd{x}')\right),\label{eq:node_feature_gp}
    \end{equation}
with the mean function $m$ and kernel function $\mathcal{K}_{\psi}$ with parameters $\psi$ operating on the node features. 
On domains described by graphs with a finite number of nodes this prior is equivalent to a multivariate normal distribution with mean $\bd{m} = m(\bd{X}) \in \R^{N}$ and covariance $\bd{K} = \mathcal{K}(\bd{X}, \bd{X}) \in \R^{N \times N}$ respectively.
As the wavelet filtering described in Equation~\ref{eq:wavelet_filtering} is a linear operation the filtered signal $\hat{\bd{f}}$ follows a Gaussian process prior
    \begin{equation}
    \hat{\bd{f}} \sim \mathcal{GP}\left(\bd{W}_\theta \bd{m}, \bd{W}_\theta \bd{K} \bd{W}_\theta^\top \right).\label{eq:filtered_prior}
    \end{equation}
When combined with a likelihood $p(\bd{y} \,\vert\, \hat{\bd{f}})$, the model is capable of Bayesian inference of an output signal $\bd{y} \in \R^{N}$ by computing the posterior distribution $p(\hat{\bd{f}} \,\vert\, \bd{y})$. In case of regression, the likelihood is commonly assumed to be a normal distribution $p(\bd{y} \,\vert\, \hat{\bd{f}}) = \normal(\bd{y} \,\vert\, \hat{\bd{f}}, \sigma^2 \bd{I})$ with observation noise $\sigma^2$ and the posterior distribution can be computed in closed form~\citep{rasmussen2005gp}. In case of classification, a categorical likelihood is assumed, leading to an intractable posterior. We then opt to approximate it with a variational posterior $q(\hat{\bd{f}})$ following~\citet{hensman2015gpc}.

\paragraph{Adaptive GP via learning wavelet scales.}

A key part of model design is the choice of wavelet filter $g_\theta$ (cf. Equation~\ref{eq:wavelet_filtering}). 
A wide variety of mother wavelet functions are available, 
here, we choose the Mexican Hat wavelet function for the band-pass filters, defined as
    \begin{equation}
    b_\beta(\lambda) = \frac{2 \sqrt{2}}{\sqrt{3}\pi^{\frac{1}{4}}} \left(\frac{\lambda}{\beta}\right)^2 \exp\left(-\frac{1}{2} \left(\frac{\lambda}{\beta}\right)^2\right)
    \end{equation}
with scale $\beta$.
A band-pass filters emphasises the frequencies in an interval (or band) of the spectral domain. The location of that interval is controlled by the scale $\beta$, which thereby controls the localisation of the transform in the spatial and frequency domain.
To model lower frequencies of the signal we choose a scaling function with a relatively fast decay as the low-pass filter, defined as
    \begin{equation}
    h_\alpha(\lambda) = \frac{1}{1 + \alpha \lambda}
    \end{equation}
with scale $\alpha$.
A low-pass filter emphasises the lower frequencies of a signal, corresponding to its smoother components, where smoothness is measured by the Dirichlet energy $\Vert \bd{f} \Vert_{\mathcal{G}} = \bd{f}^\top \bd{L} \bd{f}$. The scale $\alpha$ controls how much the filter smooths the signal.
To obtain the combined effect of the low-pass and all band-pass filters, we can compute a full filter function as the sum of the individual filters. For $L$ scales, this leads to the spectral filter function
    \begin{equation}
    g_\theta(\lambda) = h_{\alpha}(\lambda) + \sum\limits_{l = 1}^{L} b_{\beta_l}(\lambda)
    \end{equation}
with $\theta = \{ \alpha, \beta_1, \ldots, \beta_L \}$, which is used to compute the wavelet filter matrix $\bd{W}_\theta = \bd{U} g_\theta(\bd{\Lambda}) \bd{U}^\top$, where the subscript highlights the dependence of the filter matrix on the scale parameters. 

When the wavelet filter is applied to the GP prior as in Equation~\ref{eq:filtered_prior}, the scale parameters $\theta$ can be treated as kernel hyperparameters and can be optimised as part of the model fitting process. This is achieved by maximising the marginal log-likelihood $p(\bd{y} \,\vert\, \theta, \psi)$ with respect to both the scale parameters $\theta$ and the parameters $\psi$ of the node feature kernel $\mathcal{K}_{\psi}$ (cf. Equation~\ref{eq:node_feature_gp}):
    \begin{align}
    \theta, \psi &= \underset{\theta, \psi}{\text{arg max }} p(\bd{y} \,\vert\, \theta, \psi)\nonumber\\
    &= \underset{\theta, \psi}{\text{arg max}} \int p(\bd{y} \,\vert\, \hat{\bd{f}}) p(\hat{\bd{f}} \,\vert\, \theta, \psi) \,\text{d}\hat{\bd{f}},
    \end{align}
where we highlight the dependence of the GP prior $p(\hat{\bd{f}} \,\vert\, \theta, \psi)$ on the hyperparameters by explicitly conditioning on them.
In the case of classification, which prescribes a non-Gaussian likelihood, the marginal likelihood is intractable and we therefore resort to maximising a variational lower bound (Equation \ref{elbo}) on the marginal likelihood, again following~\citet{hensman2015gpc}. This setup enables the model to learn to emphasise frequencies in the data that best describe the output signal $\bd{y}$ at hand. In Section~\ref{sec:experiments}, we examine the model's ability to recover the correct scale in a synthetic data experiment.

\paragraph{Spectrum-adaptive polynomial approximation.}

The model formulation described in 
previous sections requires computing the eigen-decomposition of the Laplacian of the input graph $\mathcal{G}$, which has computational complexity in $\mathcal{O}(N^3)$ and is therefore intractable for larger graphs. To alleviate this limitation, we opt for choosing to approximate the wavelet filter $g_\theta(\lambda)$ with a polynomial $p_\theta(\lambda) = \gamma_0 + \gamma_1 \lambda + \ldots + \gamma_K \lambda^K \approx g_\theta(\lambda)$ of degree $K$, as previously suggested by~\citet{hammond2011wavelets}. This allows rewriting the filtering operation in Equation~\ref{eq:wavelet_filtering} as
    \begin{equation}
    \hat{\bd{f}} = \bd{U} g_\theta(\bd{\Lambda}) \bd{U}^\top \bd{f} \approx \bd{U} p_\theta(\bd{\Lambda}) \bd{U}^\top \bd{f} = p_\theta(\tilde{\bd{L}}) \bd{f}.
    \end{equation}
This formulation circumvents the expensive eigendecomposition of the graph Laplacian and furthermore allows exploiting the sparsity of the Laplacian by using sparse matrix-vector multiplication to compute $p_\theta(\tilde{\bd{L}}) \bd{f}$, which reduces the complexity of the filtering operation to $\mathcal{O}(K E)$, where $E$ is the number of edges in the graph. Existing approaches have relied on a truncated Chebyshev polynomial approximation of the filtering operation and freely optimising the polynomial coefficients $\bd{\gamma} \in \R^{K+1}$~\citep{hammond2011wavelets, defferrard2016chebnet}. In contrast, our approach is based on optimising the scale parameters (see previous sections) 
and we therefore require a polynomial approximation that is parameterised by the wavelet scales $\theta$. A natural choice is the least squares approximation to the filter function $g_\theta(\lambda)$
    \begin{equation}
    \bd{\gamma}_\theta = (\bd{V}_\xi^\top \bd{V}_\xi)^{-1} \bd{V}_\xi^\top g_\theta(\bd{\xi}),
    \end{equation}
where $\bd{\xi} \in \R^S$ is a set $\{\xi_i\}_{i=1}^{S}$ of linearly spaced points on the spectral domain in the interval $[0, 2]$ and $\bd{V}_\xi \in \R^{S \times (K+1)}$ is the Vandermonde matrix for $\bd{\xi}$ up to degree $K$. 

The above least-squares approximation minimises the approximation error uniformly on the spectral domain. However, in graphs with multi-scale characteristics, the eigenvalues are not uniformly distributed on the spectral domain but rather display \textit{spectral gaps} corresponding to the different scales in the data (cf. Figure~\ref{fig:overview}). As the filter function $g_\theta(\lambda)$ is only ever evaluated at the eigenvalues of the graph, a high approximation error of the polynomial approximation at those spectral gaps can be accepted in turn for a lower approximation error on parts of the spectrum with a higher density of eigenvalues. Following the ideas of~\citet{shuman2015adapted, fan2020approx}, we achieve this by computing a weighted least square approximation of the filter function $g_\theta(\lambda)$, where the weights are chosen to be proportional to the spectral density of the graph~\cite[Chapter 6]{piet2011spectra}, which is defined as
    \begin{equation}
    p_\lambda(z) \coloneqq \frac{1}{N} \sum\limits_{l=1}^{N} \mathds{1}_{\{\lambda_l = z\}}.
    \end{equation}
\textit{Spectral density estimation} aims to approximate this function without performing the expensive eigendecomposition of the graph Laplacian. We opt to employ the Kernel Polynomial Method~\citep{lin2016approximating, li2016scalable, silver1994densities, silver1996kernel, wang1994calculating} to find an estimate of the spectral density function by first finding an estimate for the cumulative spectral density function $P_\lambda(z) \coloneqq \frac{1}{N} \sum_{l=1}^{N} \mathds{1}_{\{\lambda_l \leq z\}}$. For each $\xi_i$ from the set $\{\xi_i\}_{i=1}^{S}$ of $S$ linearly spaced points on the spectral domain, we aim to find the number of eigenvalues less than or equal to $\xi_i$. This can be achieved via stochastic trace estimation~\citep{girard1989fast}, which provides us with a randomized algorithm for computing the trace of a matrix $\bd{B}$ and we use the Gaussian estimator
    \begin{equation}
    \text{tr}(\bd{B}) = \expect\left[\bd{z}^\top \bd{B} \bd{z} \right] \approx \frac{1}{R} \sum\limits_{r=1}^{R} \bd{z}^\top \bd{B} \bd{z}, \text{\hspace{0.5cm}}\bd{z} \sim \normal(\bd{0}, \bd{I}),
    \end{equation}
where $R$ is the number of Monte Carlo samples drawn for approximating the expectation.
We thus require a matrix function $\Theta_{\xi_i}$ that maps the Laplacian $\tilde{\bd{L}}$ to a matrix whose trace equals the number of eigenvalues less or equal to $\xi_i$. This mapping is trivially given by $\Theta_{\xi_i}(\lambda) = \mathds{1}_{\{\lambda \leq \xi_i\}}$.
While we are not able to cheaply compute $\Theta_{\xi_i}$ exactly, we can approximate it using a Jackson-Chebyshev polynomial $\tilde{\Theta}_{\xi_i}$ (details of this approximation can be found in~\citep{napoli2016numerical, puy2018structured}). We obtain an approximation $\tilde{P}_\lambda(z)$ to the cumulative spectral density function by interpolating between the estimates at points $\xi_i$ using monotonic piece-wise cubic interpolation $\mathcal{I}$
    \begin{align}
    \tilde{P}_\lambda(z) = \mathcal{I}\left(\left\{\left(\xi_i, \frac{1}{N} \left[ \frac{1}{R} \sum\limits_{r=1}^{R} \bd{z}_r^\top \tilde{\Theta}_{\xi_i}(\tilde{\bd{L}}) \bd{z}_r \right]\right) \right\}_{i=1}^{S}\right),\nonumber\\ 
    \text{\hspace{0.5cm}}\bd{z} \sim \normal(\bd{0}, \bd{I}).\nonumber
    \end{align}
Finally, differentiating $\tilde{P}_\lambda(z)$ with respect to $z$ gives an approximation $\tilde{p}_\lambda(z)$ to the spectral density.

Using this estimate of the spectral density, we can compute weights $\bd{\omega} \in \R^{S}$ for each of the $S$ sample points $\xi_i$ on the spectral domain. We can then compute the weighted least squares coefficients
    \begin{equation}
    \bd{\gamma}_\theta = \underbrace{\left(\bd{V}^\top \text{diag}(\bd{\omega}) \bd{V}\right)^{-1} \bd{V}^\top \text{diag}(\bd{\omega})}_{\text{projection matrix } \bd{P}} g_\theta(\bd{\xi})
    \end{equation}
to be used in the polynomial approximation $p_\theta(\tilde{\bd{L}})$. The spectral density weights $\bd{\omega}$ may be pre-computed before training and combined into the projection matrix $\bd{P} \in \R^{(K+1) \times S}$, which projects from the exact filter values $g_\theta(\bd{\xi})$ to the polynomial coefficients $\bd{\gamma}_\theta$. Finally, these coefficients are used to approximate the wavelet filter matrix $\bd{W}_\theta \approx \gamma_0 \bd{I} + \gamma_1 \tilde{\bd{L}} + \gamma_2 \tilde{\bd{L}}^2 + \ldots$, where we have dropped the coefficient's explicit dependence on the scale parameters $\theta$ for notational clarity.

\section{RELATED WORK}

Our work is first related to recent developments in developing GP models to handle graph-structured data, where the main challenge is to incorporate the graph information into the design of GP kernels. The first option is to directly encode the relational structure of nodes provided by the graph as an aggregation of the kernel matrix \citep{Ng18,Li20, liu2020uncertainty, cheng2020dynamic}. A second option is to leverage notions of graph convolutions for the same purpose \citep{Opolka20,Walker19}. 
From a slightly different perspective, the studies by \citet{Venkitaraman20,Zhi20} follow the literature of multi-output GPs with a separable kernel design. 
Finally, a Matérn GP on graphs has been proposed by \citet{Borovitskiy21}, although their model resembles kernels on graphs \citep{Smola03}. 
All of these studies, however, do not exploit the topological properties of the graph on which the GP is built; furthermore, in the context of graph GPs, only the recent work by \citet{Zhi20} has attempted to learn an adaptive graph filter via a polynomial design, and their work only focuses on the vector-output setting. Our work proposes an adaptive GP that utilises the spectral graph wavelets to adapt to the multi-scale properties of the graph domain as well as the data it supports.
The resulting kernel is semi-supervised for scalar-output GPs, but the graph wavelet can easily be adapted to vector-outputs.

Our study is more broadly addressing the recent attempts in incorporating signal processing concepts and tools into the design of graph-based learning models, especially the graph neural networks (GNNs) \citep{wu2019comprehensive}. One well-documented issue of these models is over-smoothing \citep{li2018deep,oono2020graph} which, from a signal processing perspective, may be interpreted as a result of merely low-pass filtering of the graph signals~\citep{wu2019simplifying}. As a consequence, they may also not be suitable for scenarios where the labels exhibit a low level of homophily \citep{zhu20beyond}. Several recent studies have attempted to address these issues by designing filters that go beyond low-frequency information \citep{min2020scattering,bo2021beyond, zheng2021framelets,liao2018lanczosnet,nt2020stacked,dong2020graph, he2021bern}. 
Although these frameworks offer the possibility to go beyond low-pass filtering, they are all neural network models which typically require a large amount of training data and lack a measure of predictive uncertainty. Our work proposes a GP model that address both limitations, where the learnable wavelet filters offer the flexibility in representing different types of signal spectra.

Our work is finally related to multi-scale analysis of graph data. Various wavelet transforms have been developed to analyse both the graph and the signals at different scales \citep{coifman2006diffusion,Hammond11,gama2019diffusion}.
For example, the work by \citet{tremblay2014graph} has adopted the spectral graph wavelets and tackled the problem of detecting community structure at multiple levels, while the application of wavelets by \citet{Xu2019GraphWN} is used as an alternative to the graph Fourier basis. In the latter, the bases are however limited to low-pass and as a result lacks the spectral multi-scale property of our model. In terms of data defined on graphs, recent studies have utilised scattering transforms, which are based on diffusion wavelets, for applications such as node and graph classification as well as dimensionality reduction \citep{gao2019geometric,min2020scattering}. To the best of our knowledge, our framework is the first that incorporates graph wavelets into GP design for the same purpose.






\begin{figure}[t]
\centering
\begin{subfigure}[b]{0.49\linewidth}
\includegraphics[width=\linewidth]{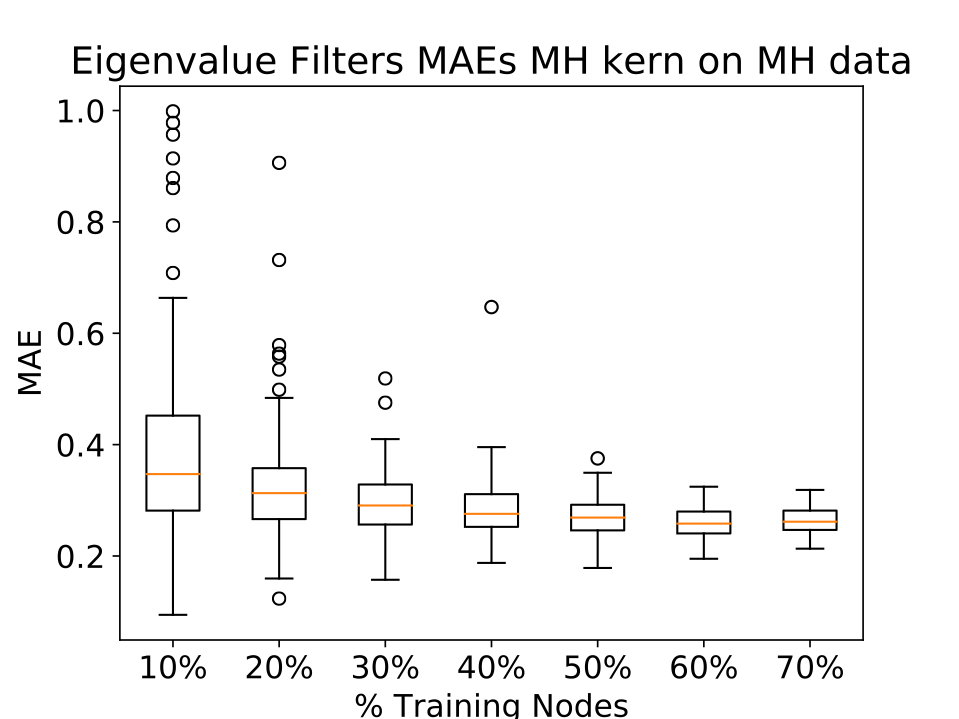}
\caption{}\label{mae1}
\end{subfigure}
\begin{subfigure}[b]{0.49\linewidth}
\includegraphics[width=\linewidth]{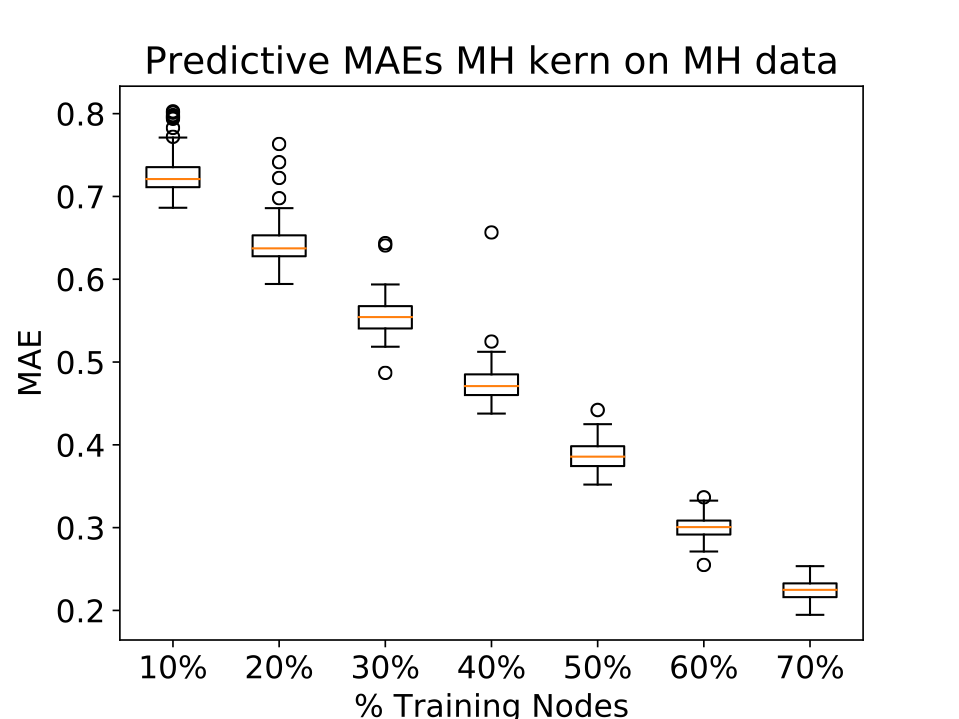}
\caption{}\label{mae2}
\end{subfigure}
\caption{(a) MAE between the recovered wavelet filter against ground truth at the eigenvalues. (b) MAE between predicted values at testing nodes and ground truth labels}\label{synthetic_maes}
\end{figure}

\section{EXPERIMENTS}
\label{sec:experiments}
\paragraph{Synthetic Multi-Scale Graphs For Scales Recovery and Predictions.} 


\begin{figure}[t]
\centering
\begin{subfigure}[b]{0.49\linewidth}
\includegraphics[width=\linewidth, height = 3cm]{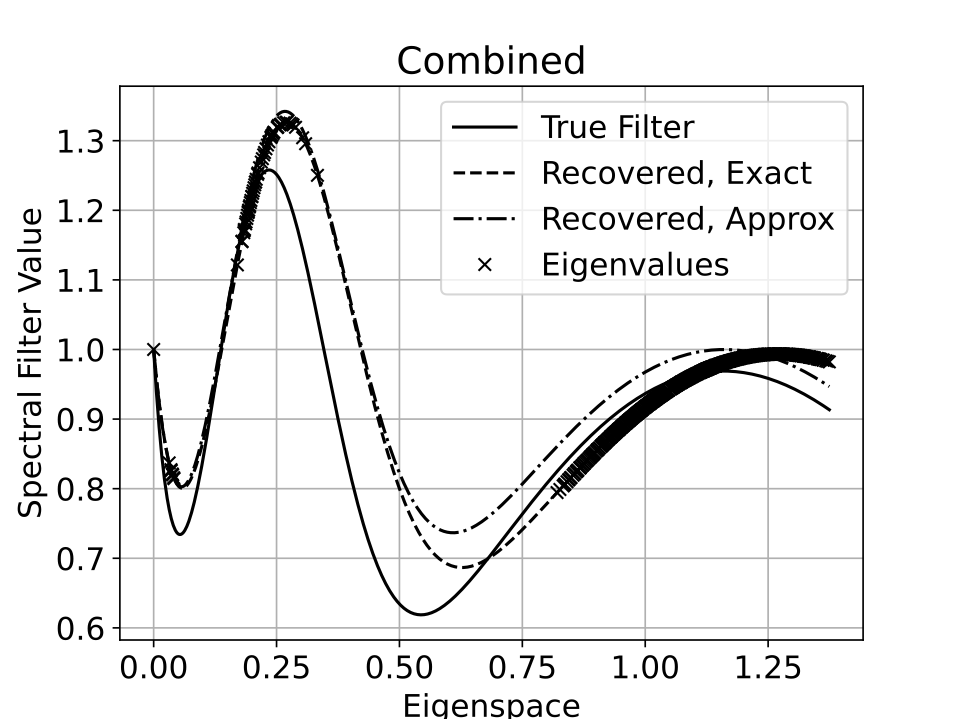}
\caption{}\label{mexicanhat4}
\end{subfigure}
\begin{subfigure}[b]{0.49\linewidth}
\includegraphics[width=\linewidth, height = 3cm]{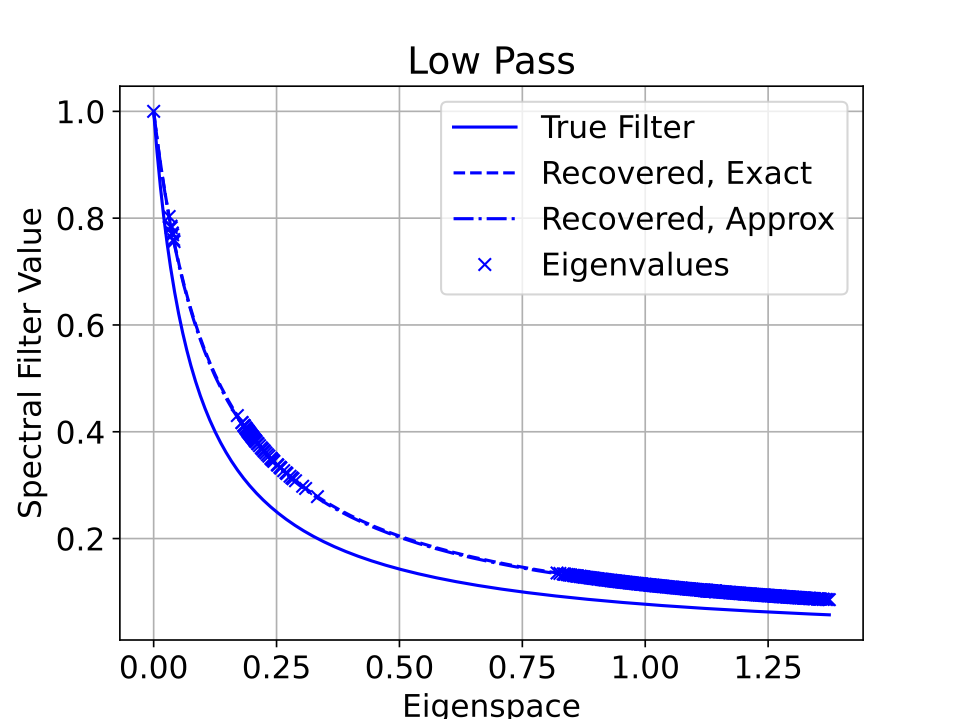}
\caption{}\label{mexicanhat1}
\end{subfigure}
\begin{subfigure}[b]{0.49\linewidth}
\includegraphics[width=\linewidth, height = 3cm]{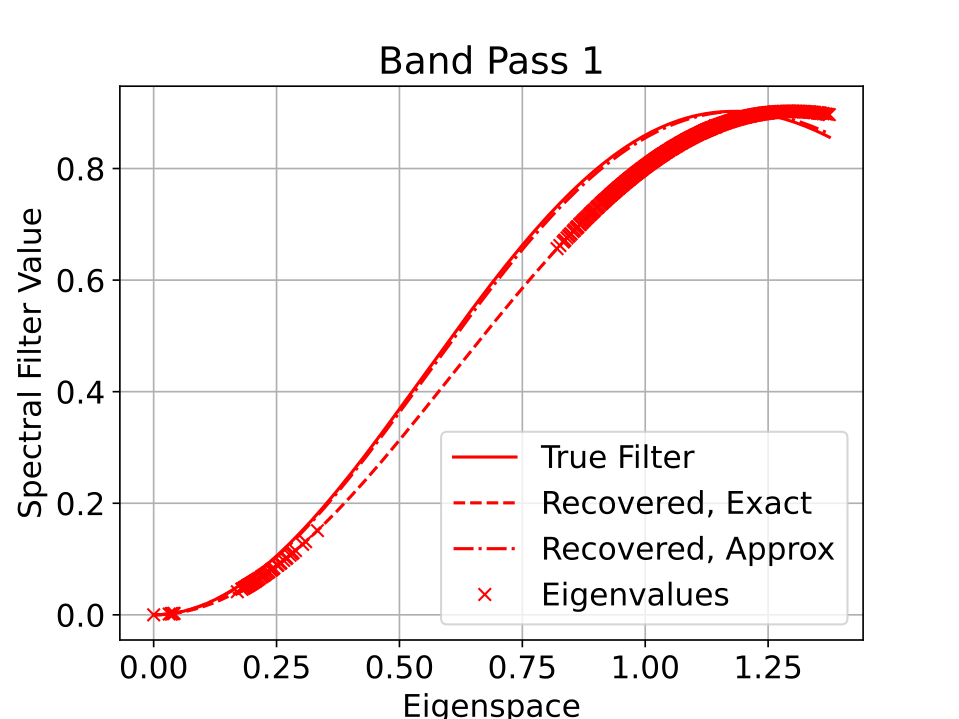}
\caption{}\label{mexicanhat2}
\end{subfigure}
\begin{subfigure}[b]{0.49\linewidth}
\includegraphics[width=\linewidth, height = 3cm]{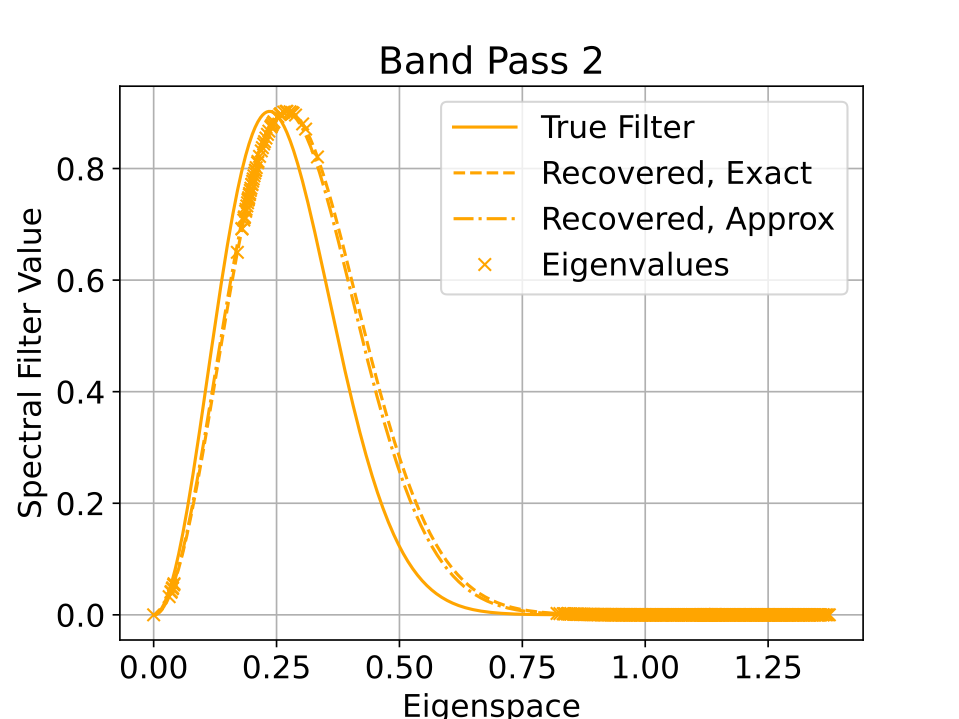}
\caption{}\label{mexicanhat3}
\end{subfigure}
\caption{Scale recoveries using exact wavelets and polynomial approximations on 50\% of data. The ground truth (a) is made of a low-pass $a = 12$ and two band-passes $s = 1.2$ \& $6$ shown in (b)-(d).}\label{fig:mexicanhat}
\end{figure}

\begin{figure}[t]
\centering
\begin{subfigure}[b]{0.49\linewidth}
\includegraphics[width=\linewidth, height = 3cm]{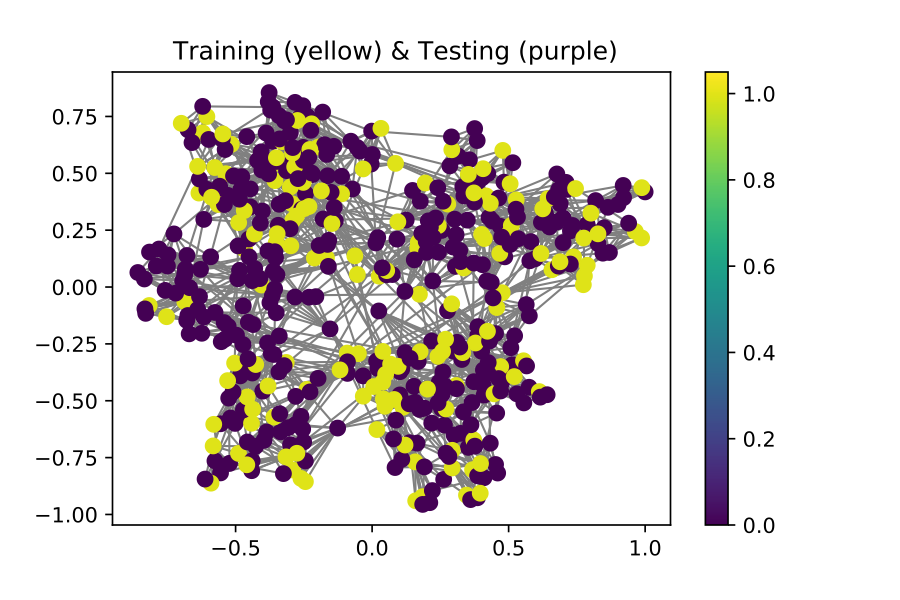}
\caption{}\label{graph_train_test}
\end{subfigure}
\hspace{-5mm}
\begin{subfigure}[b]{0.49\linewidth}
\includegraphics[width=\linewidth, height = 3cm]{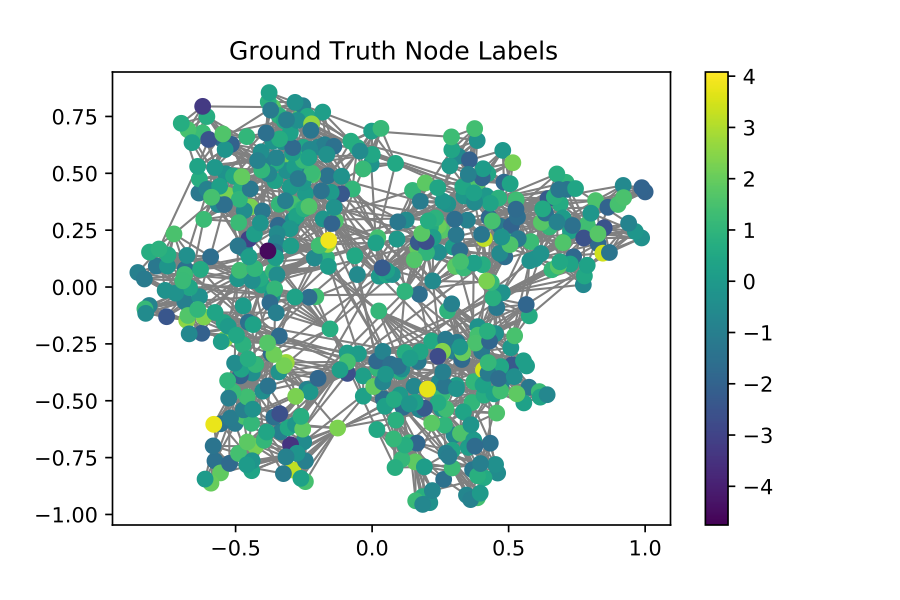}
\caption{}\label{graph_label}
\end{subfigure}
\hspace{-5mm}
\begin{subfigure}[b]{0.49\linewidth}
\includegraphics[width=\linewidth, height = 3cm]{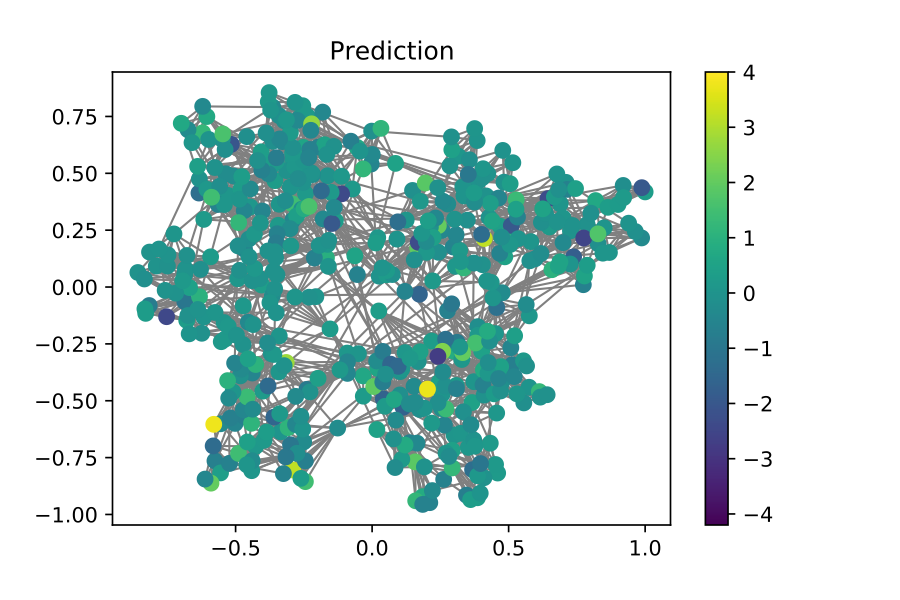}
\caption{}\label{graph_pred}
\end{subfigure}
\hspace{-5mm}
\begin{subfigure}[b]{0.49\linewidth}
\includegraphics[width=\linewidth, height = 3cm]{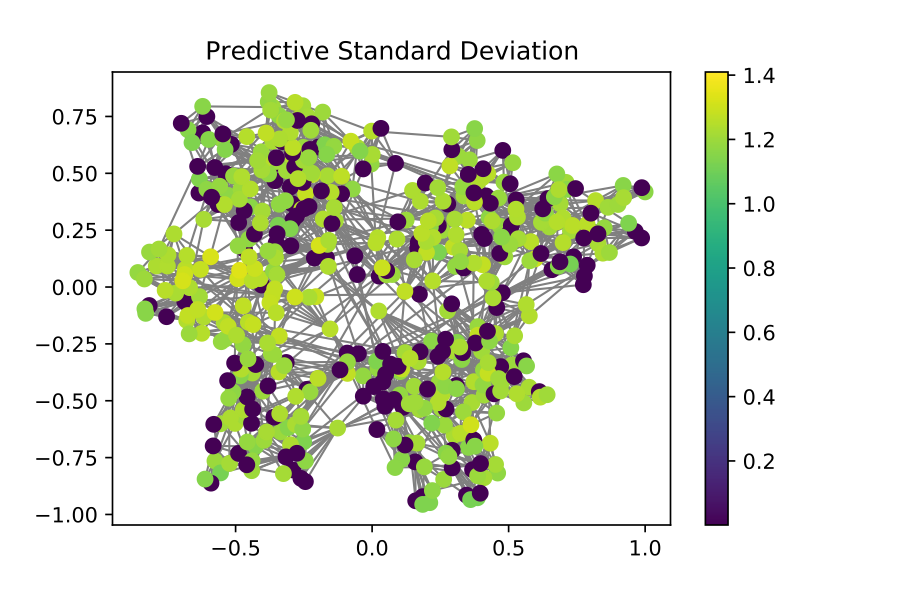}
\caption{}\label{graph_pred_std}
\end{subfigure}
\caption{(a) graph split into training (yellow) and testing (purple) nodes, only training node labels are made available to the model. (b) node labels of full graph. (c) prediction of full signal using only the training nodes. (d) Node standard deviation of posterior (these are 0 at training nodes).} \label{synth_pred}
\end{figure}

The concept of multi-scale corresponds to different things depending on if we are in the graph spatial or spectral domain. In the spectral domain, this is characterised by different dilation of the band pass filter, whereas spatially we often associate higher level scales as clustering of clusters. If the graph is spatially multi-scale, the different levels of clusters translate to gaps in the eigenvalues, which means we can apply certain characteristics to each level by adjusting a number of ground truth wavelet filters. In this synthetic setting we apply our model to graph data simulated to have both spectral and spatial multi-scale properties. We show that in optimising the GP prior for the model fitting process, we can accurately recover different scales in the wavelets of the ground truth.

We start by sampling a multi-scale graph through a specially designed algorithm. We use the Erd\H{o}s-R\'{e}nyi (ER) random graph as the base generator, and the algorithm involves repeatedly sampling ER graphs to replace the nodes in that level. Continuous labels are then generated for the nodes by sampling from a Gaussian prior with wavelets in the kernels. Let $\mathbf{W}_\phi$ represent a set of wavelets with pre-chosen set of scales $\phi = \{a = 12, s_1 = 1.2, s_2 = 6\}$ such that $\mathbf{W}_\phi = h_a(\tilde{\mathbf{L}}) + g_{s_1}(\tilde{\mathbf{L}}) + g_{s_2}(\tilde{\mathbf{L}})$. 
We do not specify any node attributes, hence an identity kernel is assumed for $\mathbf{K}$. To obtain the node labels, we sample from the Gaussian process
\begin{align}
    \mathbf{y} \sim \mathcal{GP}(0, \mathbf{W}_\phi \mathbf{W}_\phi^\top). \label{sample_label}
\end{align}
We split the labels $\mathbf{y}$ randomly into $\mathbf{y}_{\text{train}}$ and $\mathbf{y}_{\text{test}}$, with only $\mathbf{y}_{\text{train}}$ made available to the model for training. The model we use will take the form $\mathbf{f} \sim \mathcal{GP}(0, \mathbf{W}_\theta \mathbf{W}_\theta^\top)$ where $\mathbf{f}$ is the prior between the training and testing nodes and $\theta = \{\alpha, \beta_1, \beta_2\}$ are parameters to be found based on the training labels provided. As in the semi-supervised setting, the full graph will be made available to the model through computing the full $\mathbf{W}_\theta$ matrix, and $\theta$ is then found by maximising the marginal log-likelihood $\mathbb{P}(\mathbf{y}_{\text{train}}|\theta, \mathcal{G})$. Once the hyperparameters are found we can condition on the training data to obtain the predictive distribution $\mathbb{P}(\mathbf{y}_{\text{test}}|\mathbf{y}_{\text{train}}, \theta, \mathcal{G})$. This distribution provides us with the mean prediction and confidence intervals as shown in Figure \ref{synth_pred}.

We look at two particular performance measures: the mean absolute error (MAE) between the ground truth wavelet filter and the recovered filter at the eigenvalues, and the MAE between $\mathbf{y}_{\text{test}}$ and the posterior mean of $\mathbb{P}(\mathbf{y}_{\text{test}}|\mathbf{y}_{\text{train}}, \theta, \mathcal{G})$. For each selection and percentage of nodes used during training, we sample multiple labels as in (\ref{sample_label}) to recover the filters from. The MAEs can be found in Figure \ref{synthetic_maes}. The ground truth and recovered filters (via both exact formulation and approximation) for one specific example are presented in Figure \ref{mexicanhat4}, while the individual filters $h_a, g_{s_1}, g_{s_2}$ are also shown in Figure \ref{mexicanhat1} - \ref{mexicanhat3}. More results on scale recovery and predictions on synthetic data (including comparison against baselines) are presented in the Appendix.

\paragraph{Semi-Supervised Classification on Graphs.}

\begin{figure*}[t]
\centering
\begin{minipage}{.3\textwidth}
  \centering
  \includegraphics[width=\textwidth]{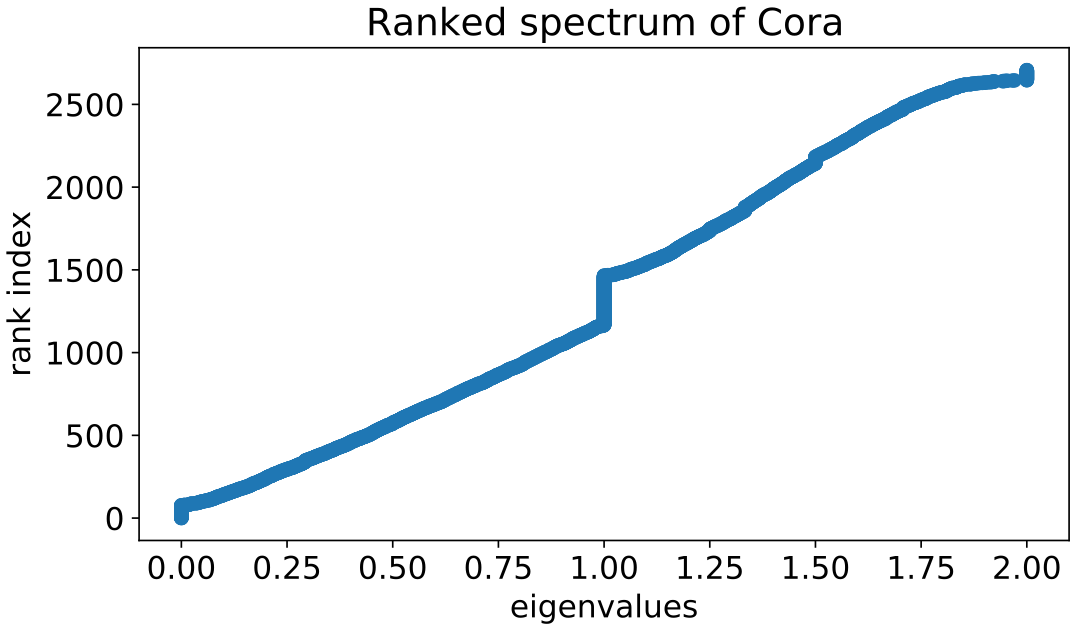}\\
  \includegraphics[width=\textwidth]{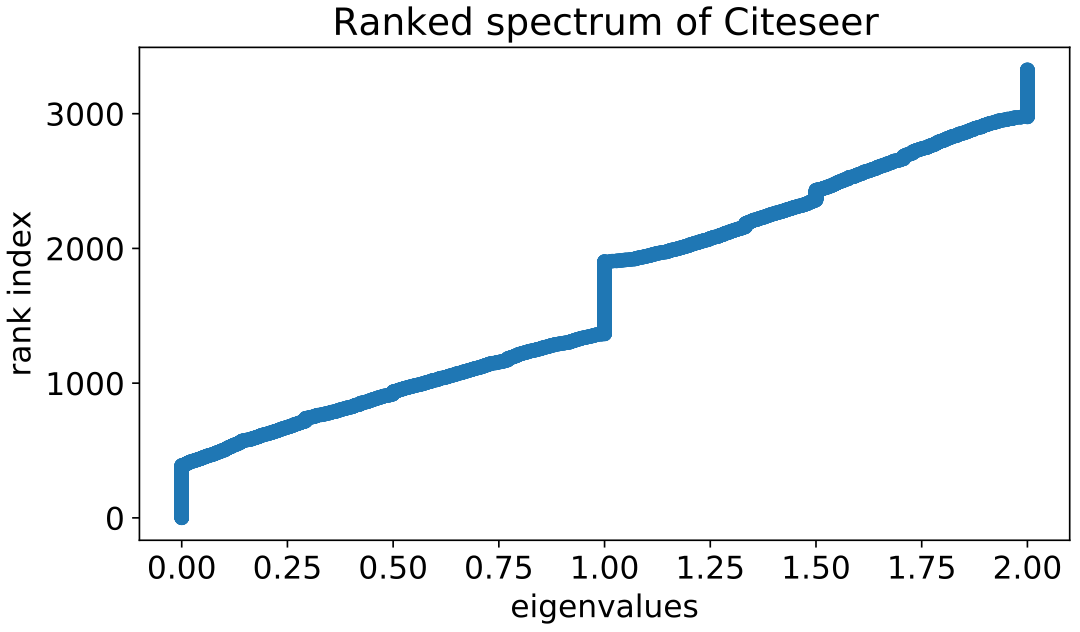}
  \captionof{figure}{Ranked spectra of Cora and Citeseer. Both present distinct ranges of eigenvalues, suggesting multi-scale graph structure.}
  \label{fig:ranked_spectra}
\end{minipage}\hfill%
\begin{minipage}{.65\textwidth}
  \centering
    \resizebox{\textwidth}{!}{%
    \begin{tabular}{lccccc}
        \toprule
        \textbf{Method} & \textbf{Cora} & \textbf{Citeseer} & \textbf{PubMed} \\
        \midrule
        \textbf{GCN} \citep{kipf2017gcn} & \res{80.5}{0.8} & \res{68.1}{1.3} & \res{77.8}{0.7} \\
        \textbf{GAT} \citep{velickovic2018gat} & \res{82.6}{0.7} & \res{72.2}{0.9} & \res{76.7}{0.5} \\
        \textbf{ChebNet} \citep{defferrard2016chebnet} & \res{78.0}{1.2} & \res{70.1}{0.8} & \res{69.8}{1.1} \\
        \textbf{LanczosNet} \citep{liao2018lanczosnet} & \res{79.5}{1.8} & \res{66.2}{1.9} & \res{78.3}{0.3} \\
        \textbf{AdaLanczosNet} \citep{liao2018lanczosnet} & \res{80.4}{1.1} & \res{68.7}{1.0} & \res{78.1}{0.4} \\
        \textbf{GP} \citep{Ng18} & 60.8 & 54.7 & 71.5 \\
        \textbf{GGP} \citep{Ng18} & 80.9 & 69.7 & 77.1 \\
        \textbf{GGP-X} \citep{Ng18} & 84.7 & 75.6 & 82.4 \\
        \textbf{ChebGP} (ours) & 79.7 & 66.5 & 77.2 \\
        \midrule
        \textbf{WGGP} (ours) & 84.7 & 70.8 & 78.4 \\
        \textbf{WGGP-X} (ours) & 87.5 & 76.8 & 90.0 \\
        \bottomrule
    \end{tabular}}
  \captionof{table}{Predictive accuracies of our proposed Wavelet Graph Gaussian Process model compared to a number of baselines. Results are reported with the mean and standard deviation over 10 runs except for Gaussian process models, which do not require random weight initialisations.}
  \label{tab:results}
\end{minipage}
\end{figure*}

\begin{figure*}[h]
\centering
\begin{subfigure}[b]{0.3\textwidth}
    \centering
    \includegraphics[width=\textwidth]{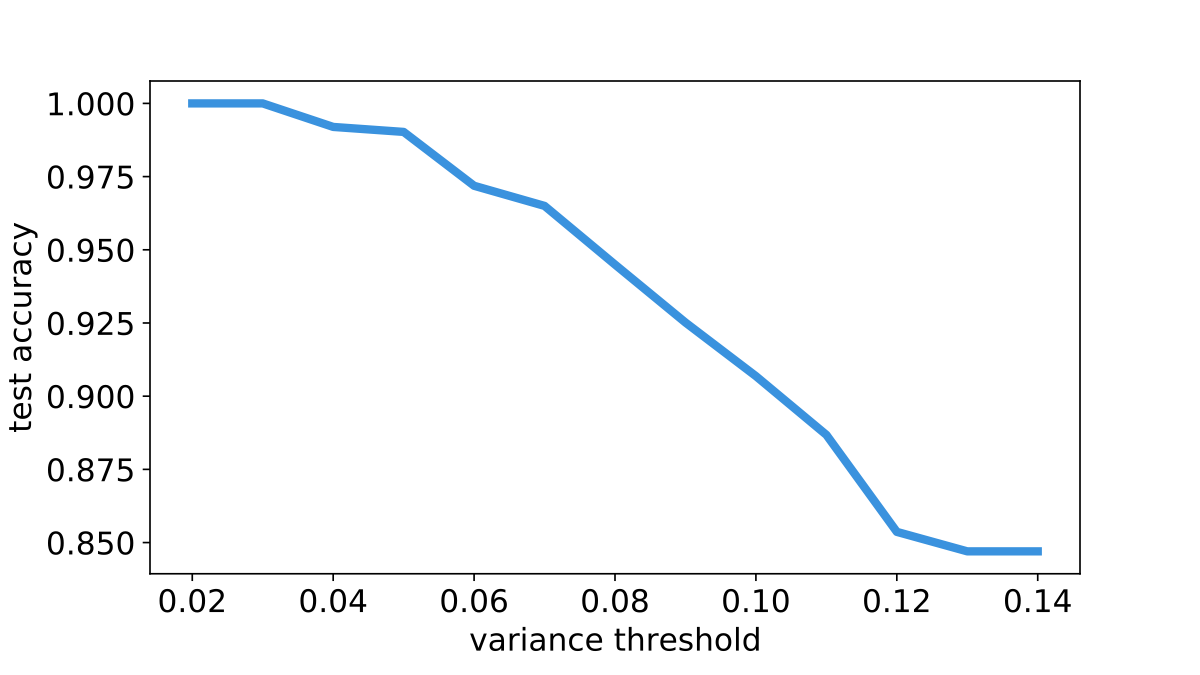}
    \caption{Cora}
    \label{fig:uncertainty_cora} 
\end{subfigure}
\begin{subfigure}[b]{0.3\textwidth}
    \centering
    \includegraphics[width=\textwidth]{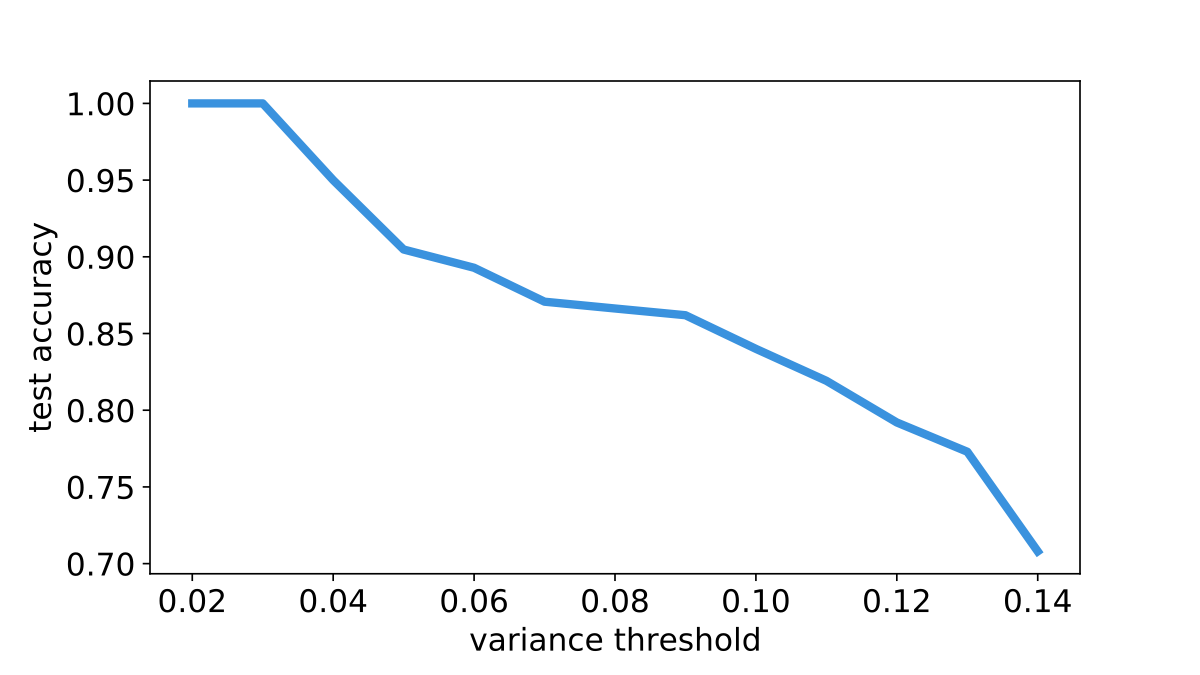}
    \caption{Citeseer}
    \label{fig:uncertainty_citeseer}
\end{subfigure}
\begin{subfigure}[b]{0.3\textwidth}
    \centering
    \includegraphics[width=\textwidth]{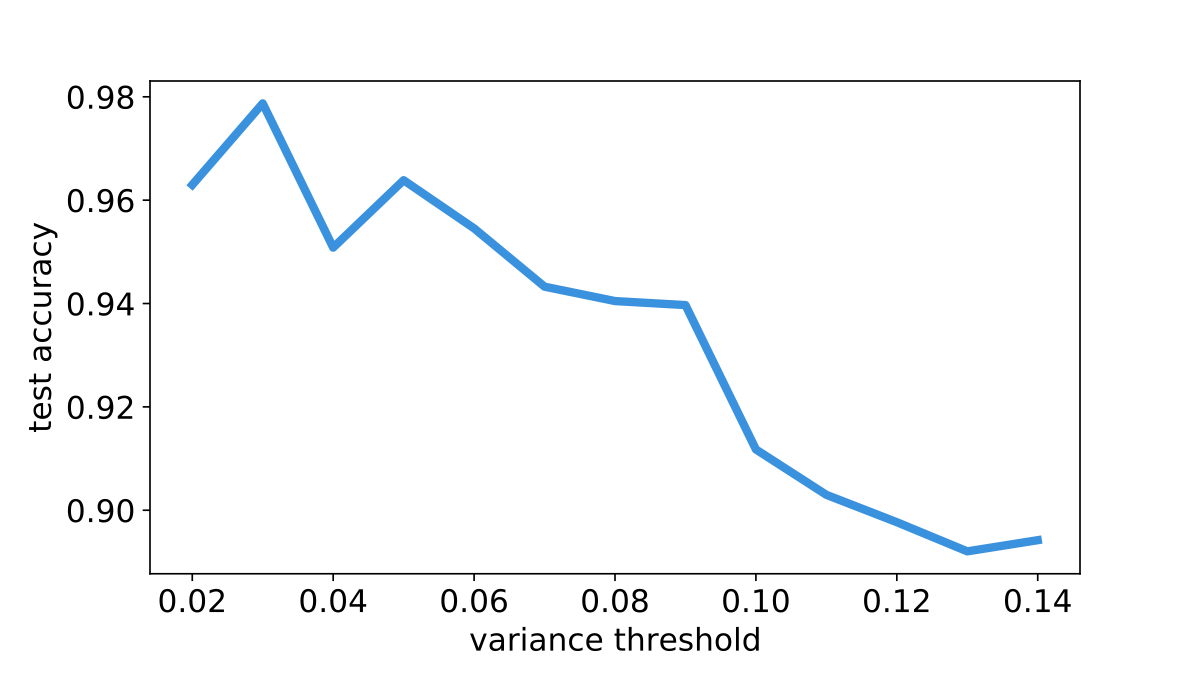}
    \caption{Pubmed}
    \label{fig:uncertainty_pubmed}
\end{subfigure}
\caption{We evaluate the performance of our WGGP model when rejecting samples with a high predictive variance, i.e. samples with high uncertainty. If the predictive variance estimates are well calibrated, as the variance threshold increases, fewer samples with high uncertainty are rejected and the accuracy should decrease.}
\label{fig:uncertainty_analysis}
\end{figure*}

We apply Wavelet Graph GP (WGGP) to three citation networks~\citep{sen2008citation}, which are commonly used as benchmark data sets for graph-based models. Here, the underlying graph consists of citations and the node features are bag-of-words (BOW) re-weighed using the popular term frequency-inverse document frequency (TFIDF) transformation. The prediction targets are the topics of the scientific papers in the networks. For the base kernel of the GP, we use a degree 3 polynomial kernel on the TFIDF features, which has been empirically shown to work well with similar models. The wavelet kernel uses two band pass-filters and a low-pass filter. 
The wavelet kernel is approximated with a degree 5 polynomial for Cora and Citeseer and with a degree 3 polynomial for PubMed. Moreover, for Cora and Citeseer, the kernel is used as part of a non-sparse variational GP, whereas for PubMed we use a sparse variational GP to enable stochastic optimisation of the ELBO using mini-batches.


The hyperparameters of the model are the initial band-pass scales and whether a low-pass filter should be included in the kernel.
We train all GP models for up to 300 epochs with a learning rate of $0.01$. 
To check convergence, we plot the ELBO curves in Figure~\ref{fig:elbo_plots} in the Appendix. Early-stopping and model selection are performed using the ELBO achieved on the training set and WGGP hence does not require a hold-out validation set. Similar to~\citet{Ng18}, we thus also report the result of WGGP trained on both the training and the validation set and refer to it as WGGP-X.
The results are presented in Table \ref{tab:results} where our model is very competitive against a set of state-of-the-art baselines including graph neural network and GP models. In particular, LanczosNet and AdaLanczosNet~\citep{liao2018lanczosnet} were included as, like the method proposed here, they are designed to extract multi-scale information from graphs. We also included a version of our model called ChebGP, which uses Chebyshev polynomials for the spectrum approximation method, to show the superiority of the polynomial approximation method we adopted. Our model outperforms both a vanilla GP model operating solely on the node features and the Graph Gaussian process (GGP) \citep{Ng18} aggregating information from the first-hop neighbourhood, thus highlighting the benefit of our multi-scale approach. Additional results and ablation studies are presented in the Appendix.

\paragraph{Uncertainty Estimates.}
Unlike the neural network baselines, our proposed GP model performs approximate Bayesian inference and therefore outputs confidence estimates for its predictions at each node $v_i$ via the variance of the variational predictive distribution $q(y_i)$.
We expect reliable variance estimates to be useful in deciding which samples to reject (and potentially send to a human labeller) because the model is unable to make a prediction with high enough confidence. We evaluate our model in this regard by computing its predictive accuracies for different variance thresholds. For a lower threshold, more low-confidence samples are rejected, which should lead to a higher predictive performance. We confirm that this property holds for the confidence estimates of our model via Figure~\ref{fig:uncertainty_analysis}.

\section{DISCUSSION}

In integrating wavelets with a GP, we have developed a model that is capable of capturing multi-scale information in the data. By including different wavelet scales, the model combines various levels of localisation on graphs to capture beyond low-frequency elements. Even though the function is defined in the graph spectral domain, by adopting a polynomial approximation we avoid an expensive eigen-decomposition, allowing the model to scale to larger graphs. We show on synthetically generated data that different scales can be recovered accurately, and the multi-scale approach leads to competitive performance on real graph data sets against state-of-the-art graph models. 

Applying the proposed wavelet model to a task at hand requires taking a number of practical considerations into account. Firstly, the number of scales in the wavelet kernel should ideally be chosen in a way such that the multi-scale graph data is captured by the different scales of the wavelets (although the model is robust to varying number of scales, cf. Appendix). For example, we may aim to match the number of scales in the kernel with the number of gaps in the spectrum by estimating the eigenvalue distribution of the graph Laplacian, which is already part of the wavelet transform approximation.
Secondly, given the nature of wavelets as dilated and shifted band-pass filters, an interesting question is which mother wavelet 
to choose for the GP model. While our model is robust to different choices of the mother wavelets (cf. Appendix)
certain options might be preferred for a given task based on their localisation properties in the spatial and spectral domain.
Finally, which nodes are selected for training can impact the learning process and final performance. If domain knowledge is available, one may look to find strategic ways to sample training nodes that will lead to the best possible characterisation of input data given a limited sampling budget.


\subsubsection*{Acknowledgements}

We would like to thank reviewers for their helpful comments. FLO acknowledges funding from the Huawei Hisilicon Studentship at the Department of Computer Science and Technology of the University of Cambridge. YCZ and XD would like to thank the support from Oxford Networks \& Society Group, Oxford-Man Institute, and an EPSRC studentship under grant EP/S515541/1.

\bibliographystyle{abbrvnat}
\bibliography{references.bib}


\clearpage
\appendix

\thispagestyle{empty}

\onecolumn \makesupplementtitle

\section{VISUALISATION OF WAVELET TRANSFORM ON A REGULAR GRID}

\begin{figure}[h]
\centering
\begin{subfigure}[b]{0.33\linewidth}
\includegraphics[width=\linewidth, height = 3.8cm]{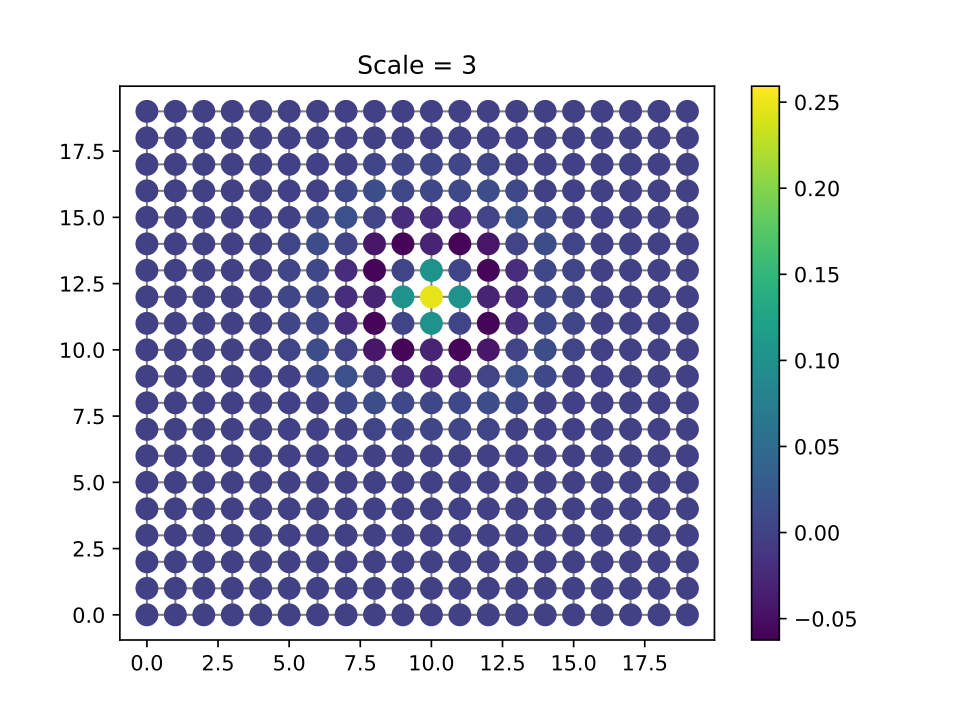}
\caption{}
\end{subfigure}
\hspace{-3mm}
\begin{subfigure}[b]{0.33\linewidth}
\includegraphics[width=\linewidth, height = 3.8cm]{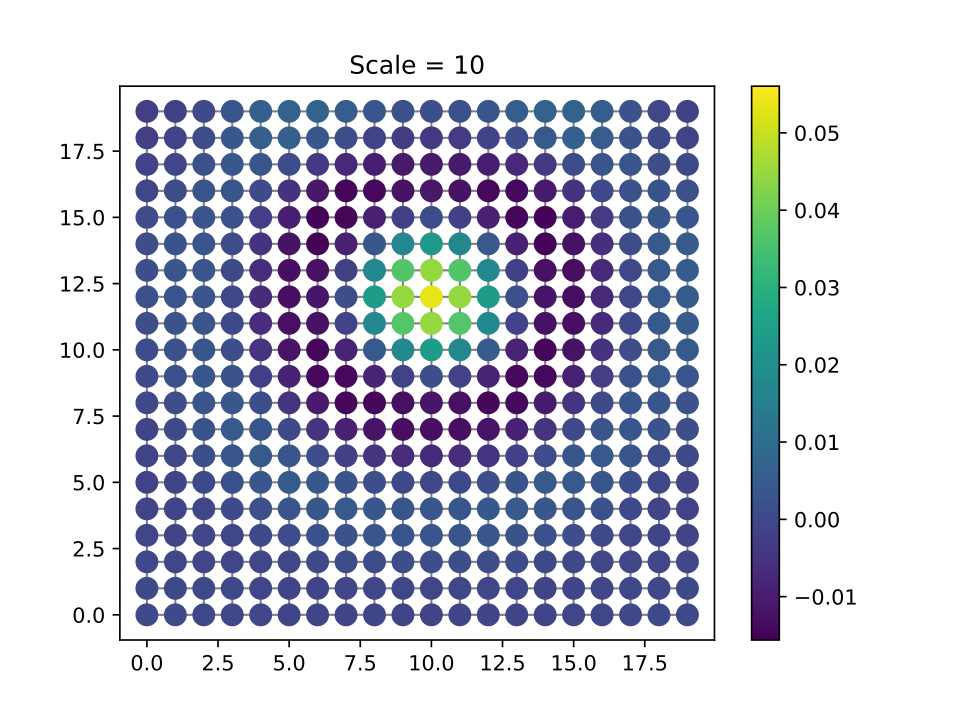}
\caption{}
\end{subfigure}
\hspace{-3mm}
\begin{subfigure}[b]{0.33\linewidth}
\includegraphics[width=\linewidth, height = 3.8cm]{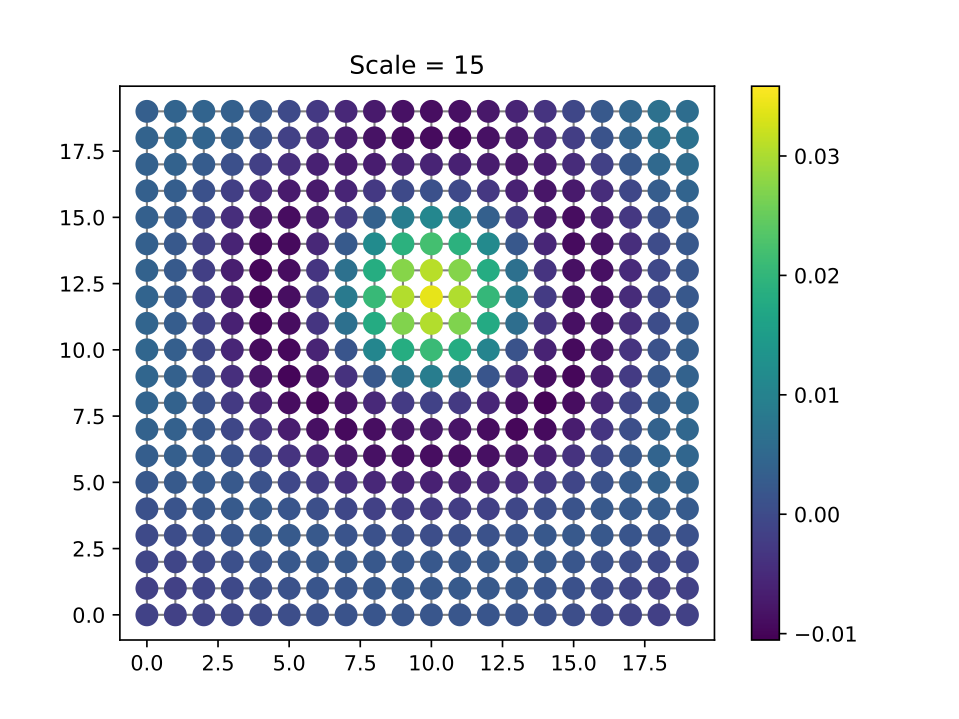}
\caption{}
\end{subfigure}
\caption{The Mexican Hat wavelet transform of a $\delta$ signal on a regular grid graph. The grid simulates a Euclidean domain to demonstrate the neighbourhoods more clearly at different scales.}\label{wavelet_transforms_grid}
\end{figure}

By applying the wavelet transforms to an impulse function centred around a certain node, we can visualise how wavelets of different scales spread around the centre node, capturing different ranges of neighbourhoods (cf. Figure \ref{wavelet_transforms}). When applied to a regular grid graph, the pattern resembles that of the Euclidean domain. This is shown in Figure \ref{wavelet_transforms_grid}, where we apply the Mexican Hat wavelet transform with various scales to show the different ranges of neighbourhoods. Neighbours are weighted continuously with intensity becoming zero once beyond a certain proximity. Thus, by using different scales, we can capture different ranges of neighbourhood information.

\section{ADDITIONAL EXPERIMENTAL RESULTS}

\subsection{Synthetic Scale Recovery Experiments and Implementation Details}

We run our synthetic experiments multiple times to show the overall behaviour of the model. We sample the labels for the nodes by Eq. (17) 
100 times; for each sample, we also randomly select a set of nodes to use for training. The hyperparameters are optimised as part of the training process. For each set of training labels, we test 20 different initializations and use the converged values that lead to the lowest loss.

\begin{figure}[h]
\centering
\begin{subfigure}[b]{0.24\linewidth}
\includegraphics[width=\linewidth, height = 3.2cm]{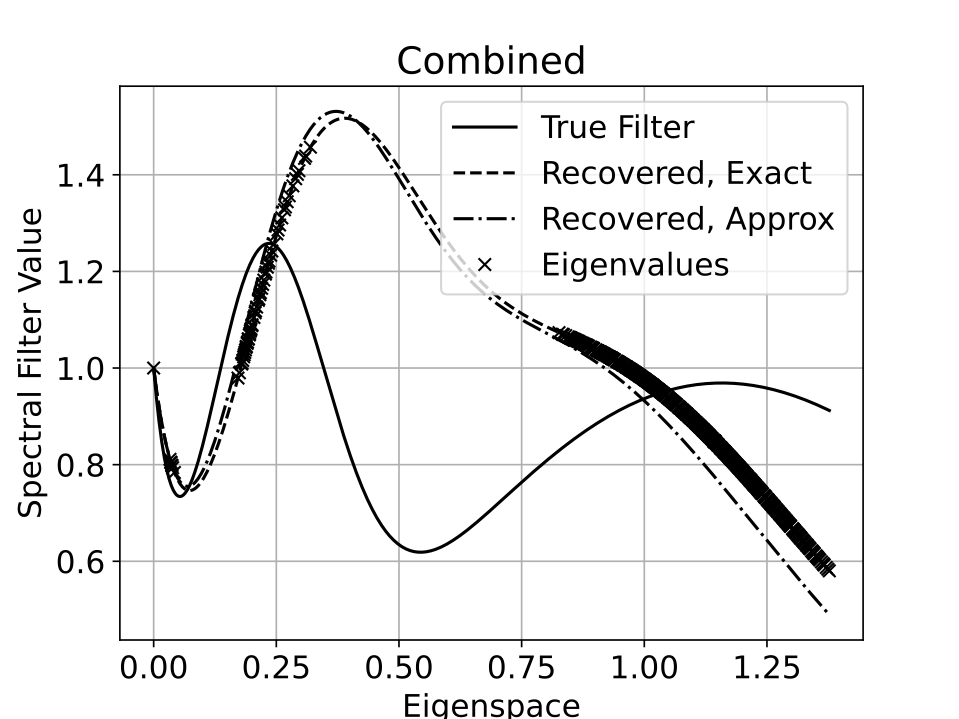}
\end{subfigure}
\begin{subfigure}[b]{0.24\linewidth}
\includegraphics[width=\linewidth, height = 3.2cm]{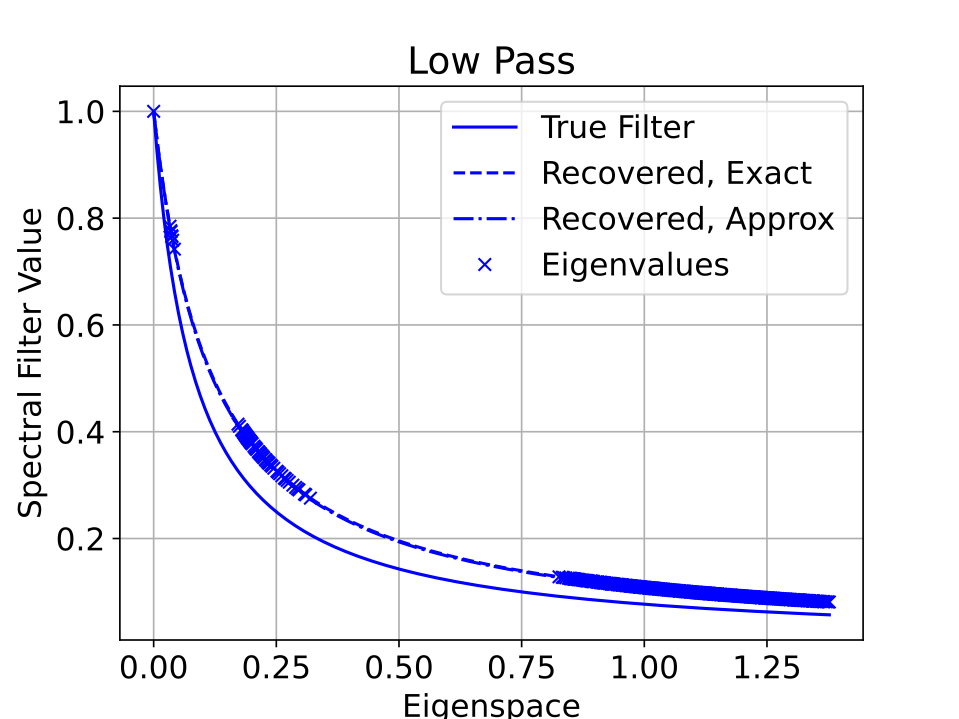}
\end{subfigure}
\begin{subfigure}[b]{0.24\linewidth}
\includegraphics[width=\linewidth, height = 3.2cm]{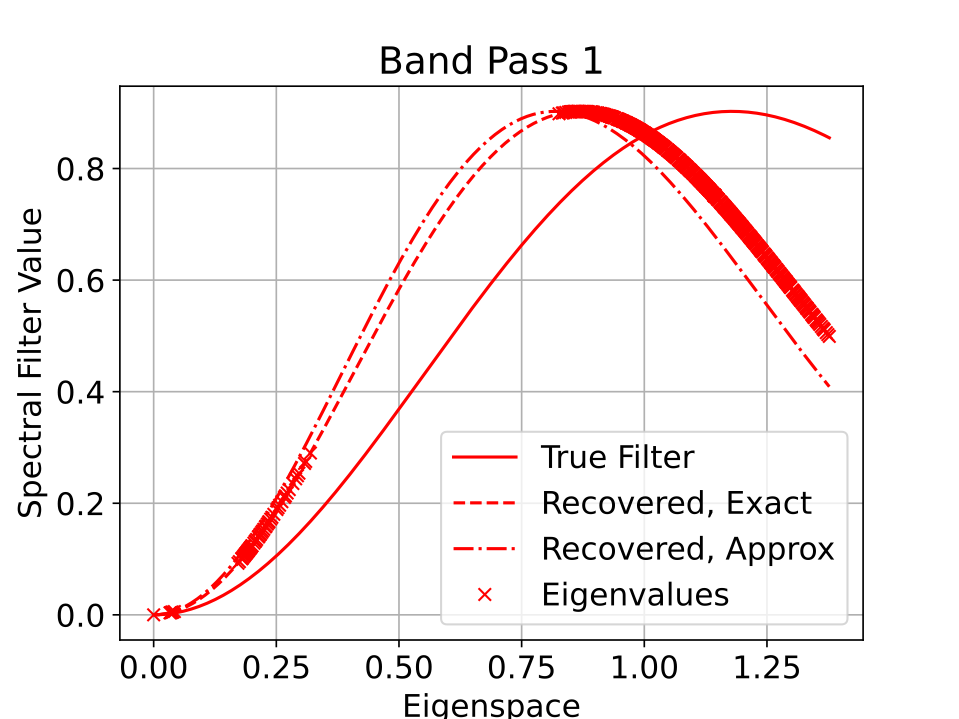}
\end{subfigure}
\begin{subfigure}[b]{0.24\linewidth}
\includegraphics[width=\linewidth, height = 3.2cm]{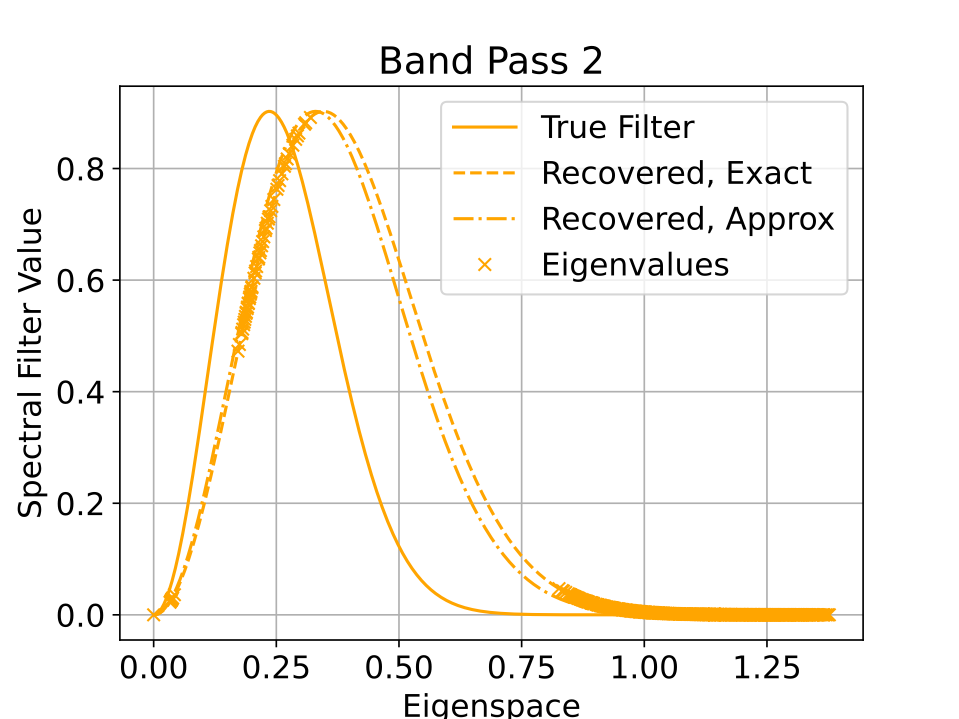}
\end{subfigure}
\begin{subfigure}[b]{0.24\linewidth}
\includegraphics[width=\linewidth, height = 3.2cm]{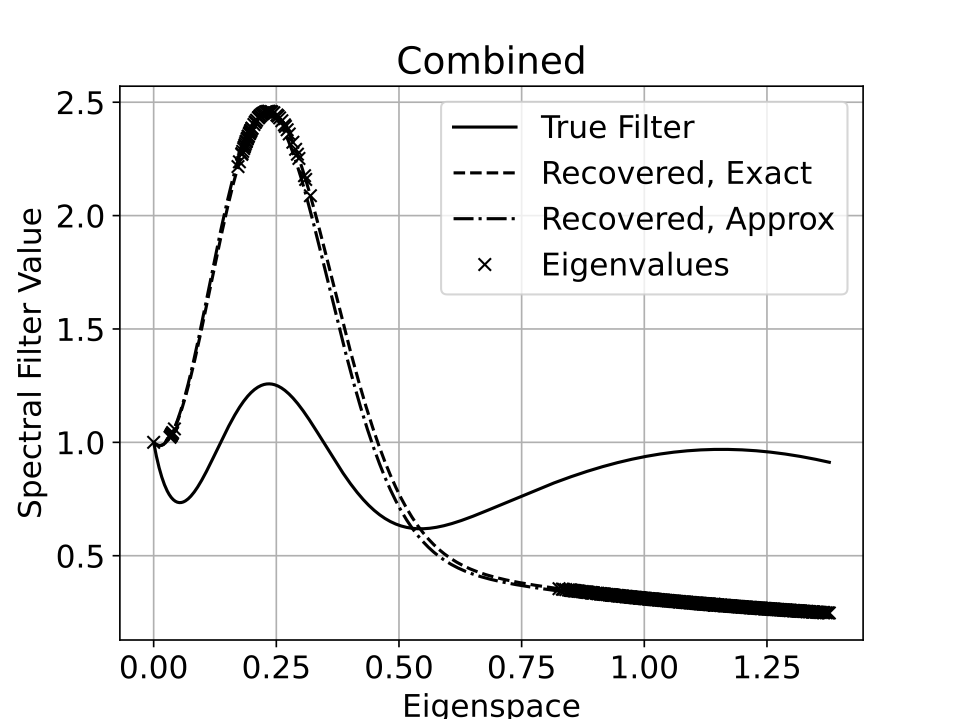}
\end{subfigure}
\begin{subfigure}[b]{0.24\linewidth}
\includegraphics[width=\linewidth, height = 3.2cm]{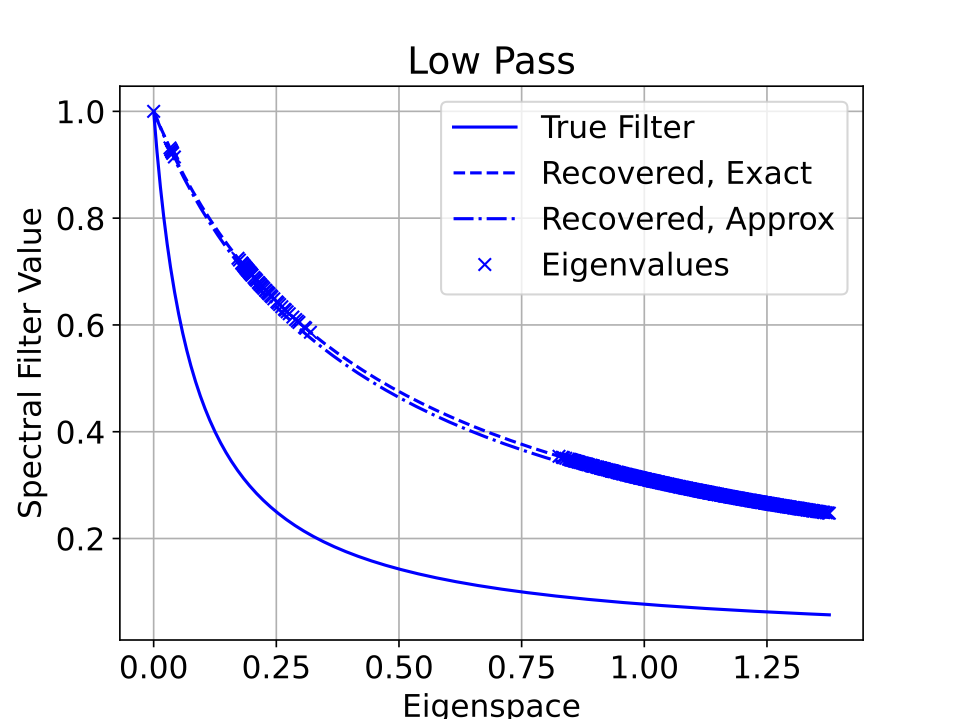}
\end{subfigure}
\begin{subfigure}[b]{0.24\linewidth}
\includegraphics[width=\linewidth, height = 3.2cm]{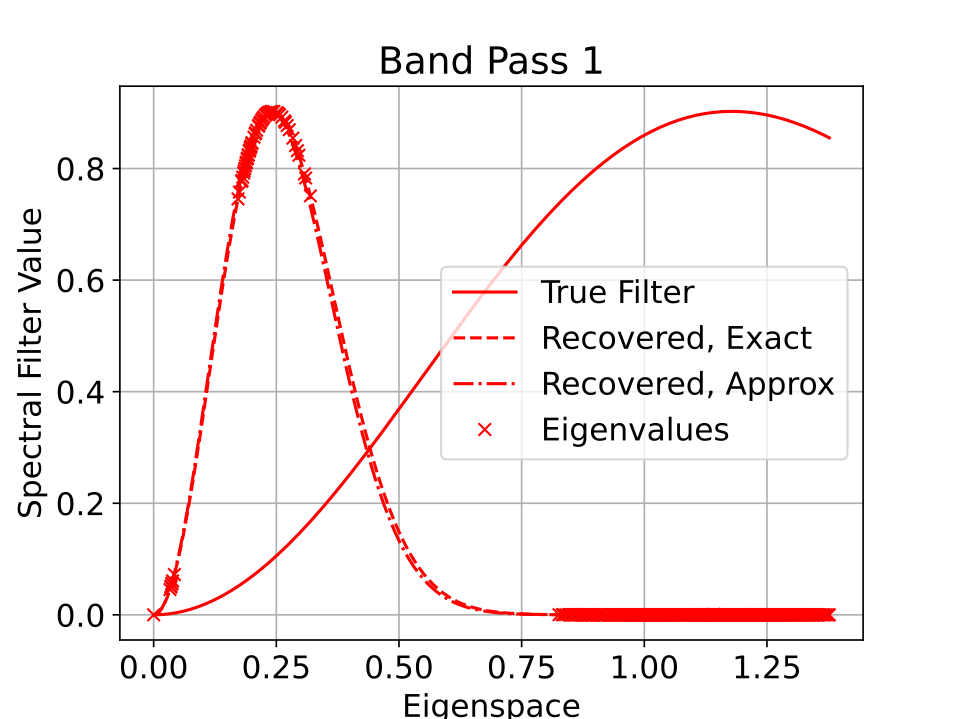}
\end{subfigure}
\begin{subfigure}[b]{0.24\linewidth}
\includegraphics[width=\linewidth, height = 3.2cm]{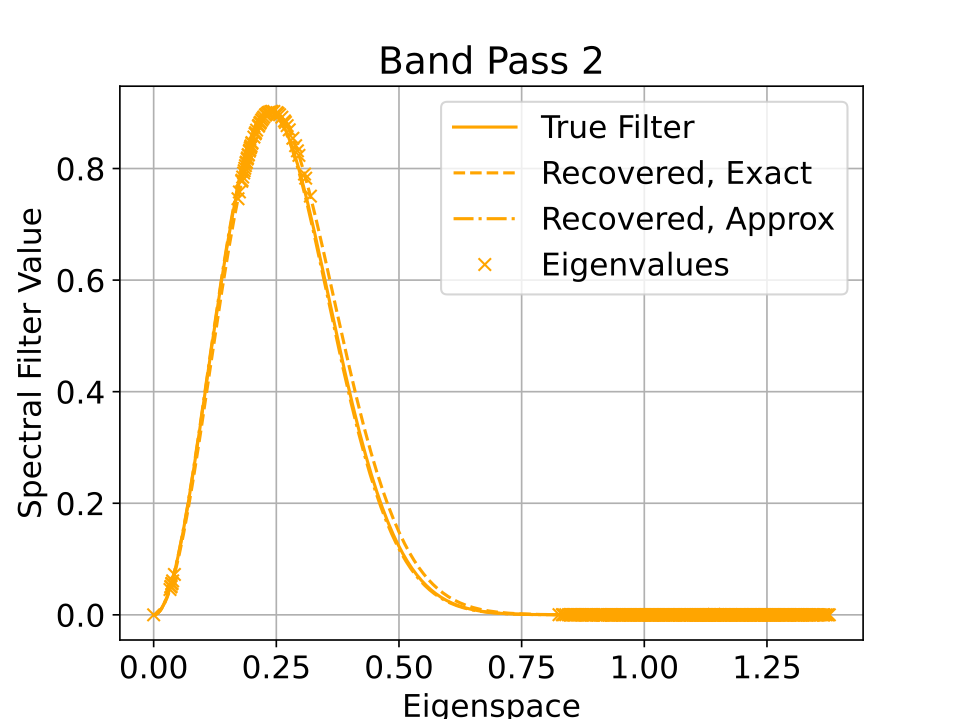}
\end{subfigure}
\begin{subfigure}[b]{0.24\linewidth}
\includegraphics[width=\linewidth, height = 3.2cm]{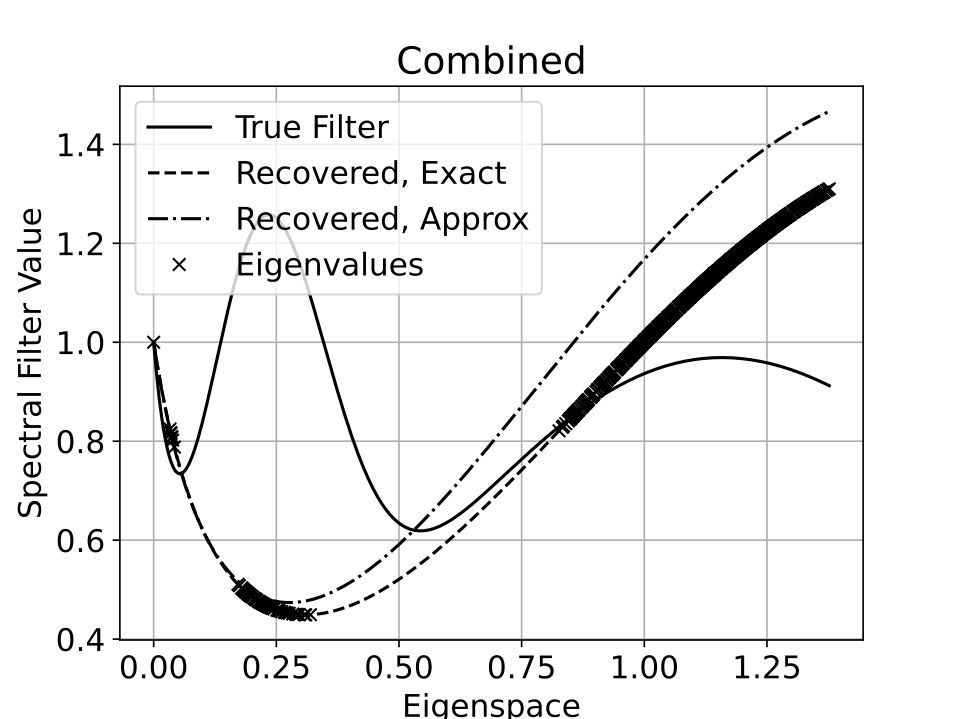}
\end{subfigure}
\begin{subfigure}[b]{0.24\linewidth}
\includegraphics[width=\linewidth, height = 3.2cm]{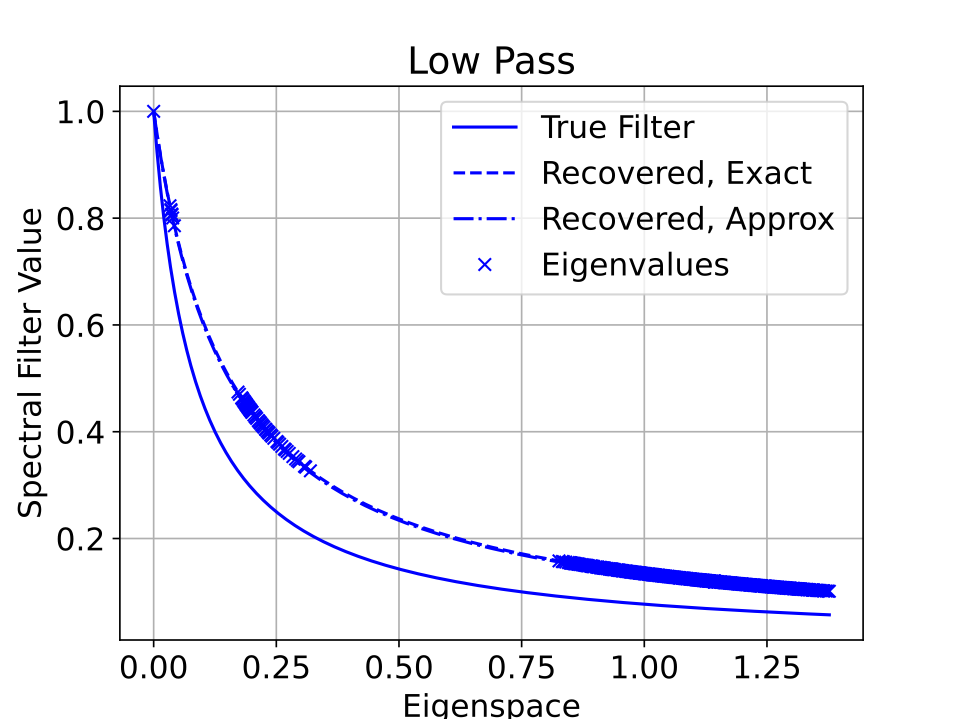}
\end{subfigure}
\begin{subfigure}[b]{0.24\linewidth}
\includegraphics[width=\linewidth, height = 3.2cm]{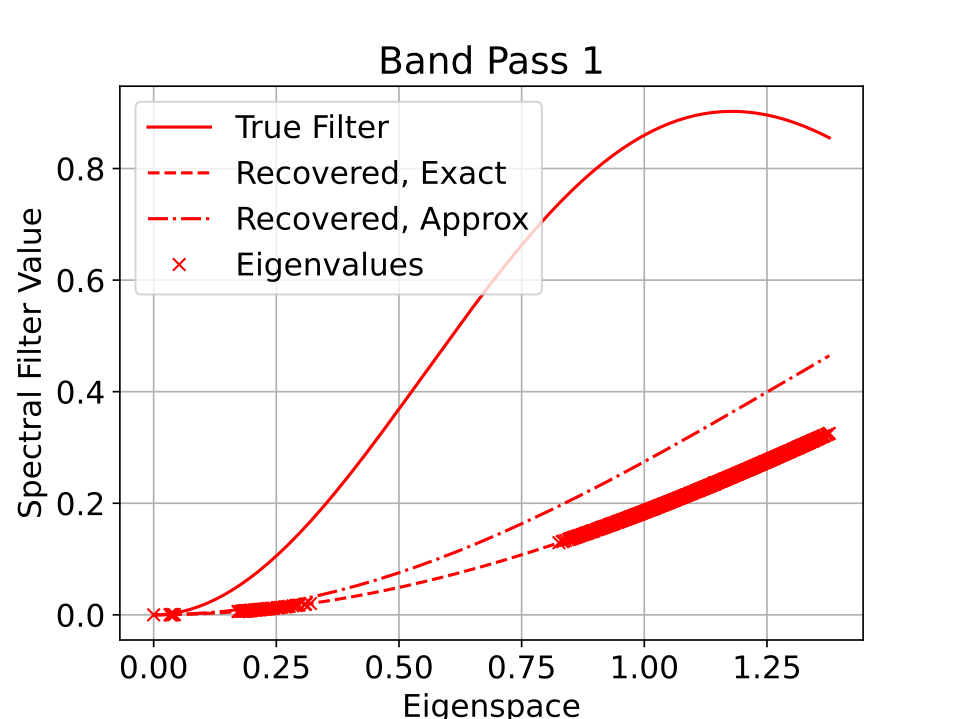}
\end{subfigure}
\begin{subfigure}[b]{0.24\linewidth}
\includegraphics[width=\linewidth, height = 3.2cm]{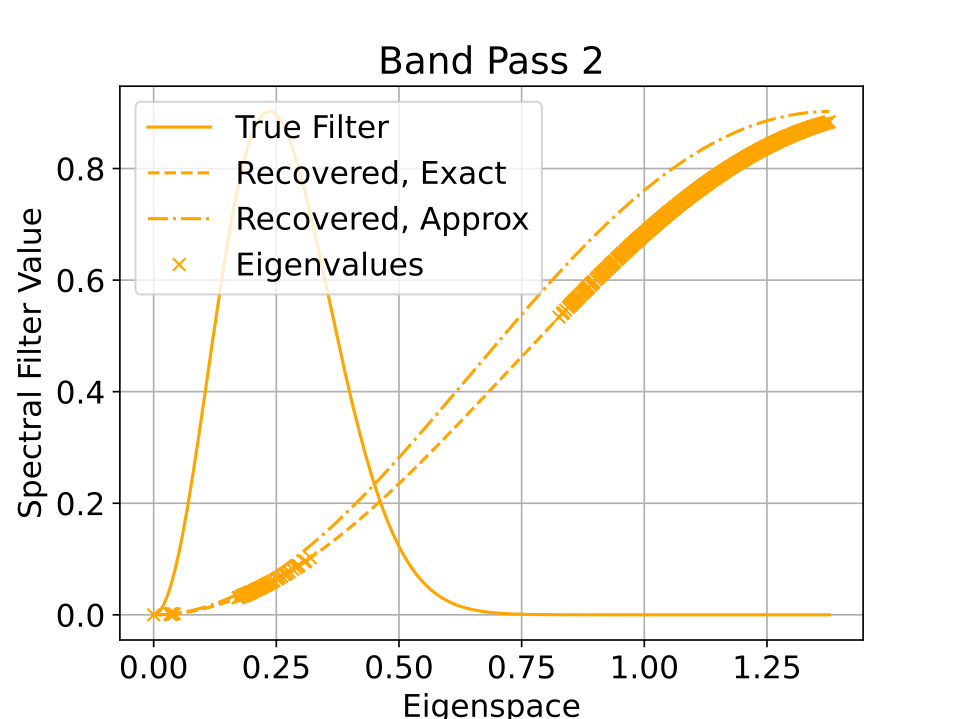}
\end{subfigure}
\caption{Scale recoveries from synthetic experiments using 10\% of nodes as training. Each row is a different random selections of training nodes.}\label{wavelet_transforms_add10}
\end{figure}

\begin{figure}[h]
\centering
\begin{subfigure}[b]{0.24\linewidth}
\includegraphics[width=\linewidth, height = 3.cm]{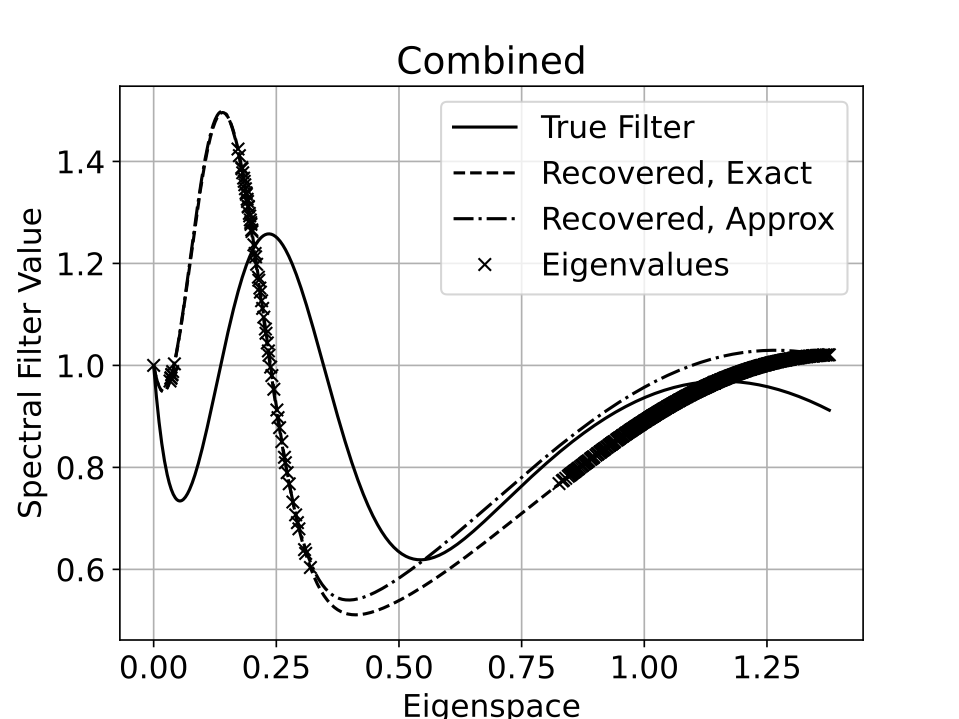}
\end{subfigure}
\begin{subfigure}[b]{0.24\linewidth}
\includegraphics[width=\linewidth, height = 3.cm]{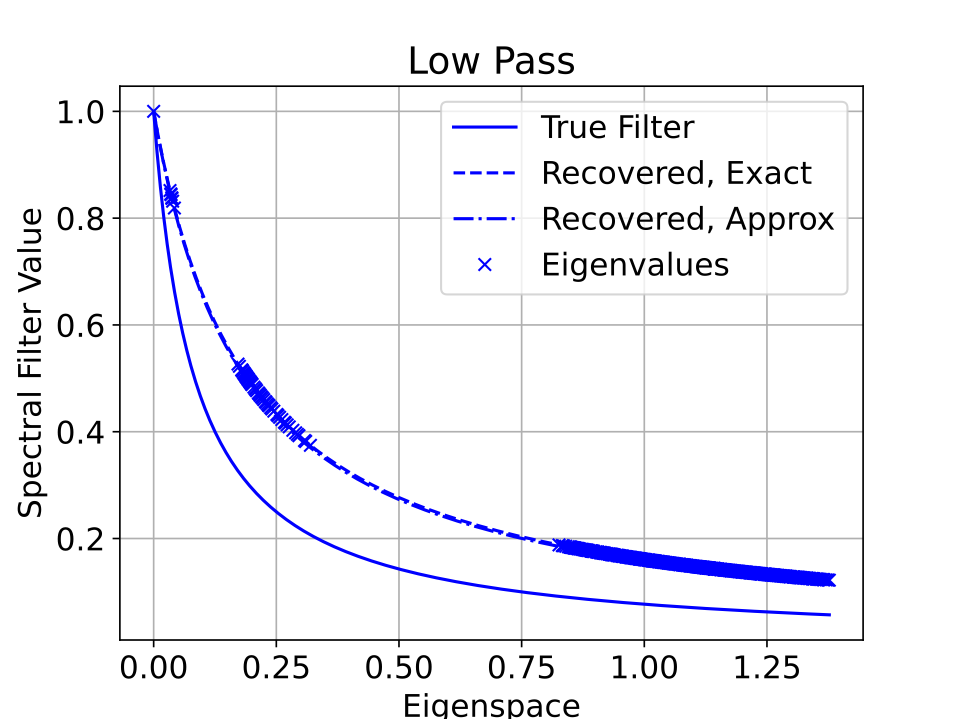}
\end{subfigure}
\begin{subfigure}[b]{0.24\linewidth}
\includegraphics[width=\linewidth, height = 3.cm]{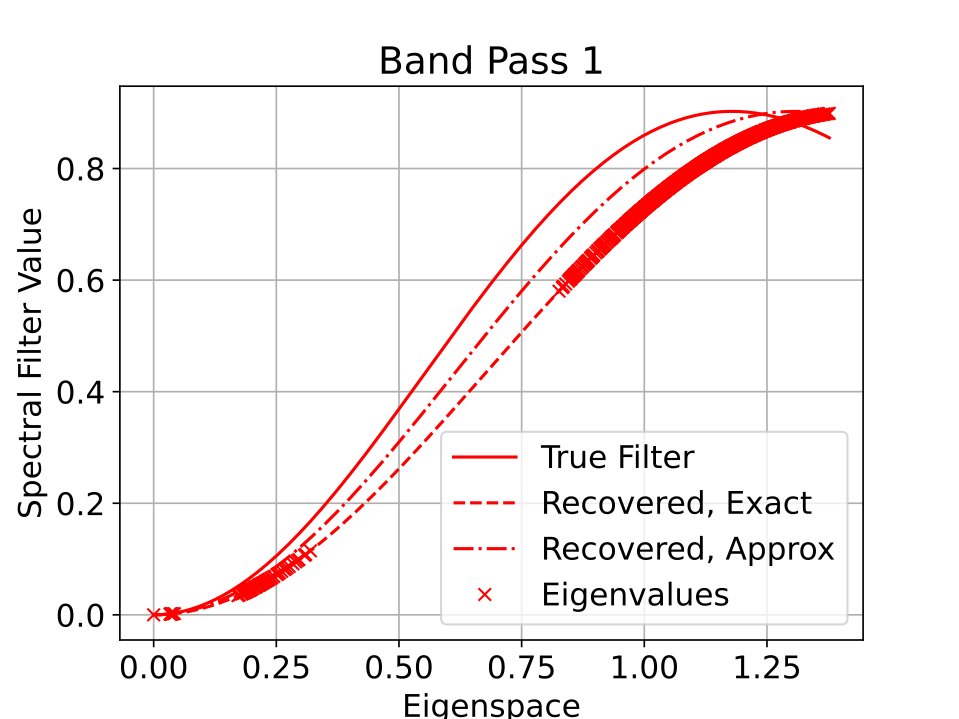}
\end{subfigure}
\begin{subfigure}[b]{0.24\linewidth}
\includegraphics[width=\linewidth, height = 3.cm]{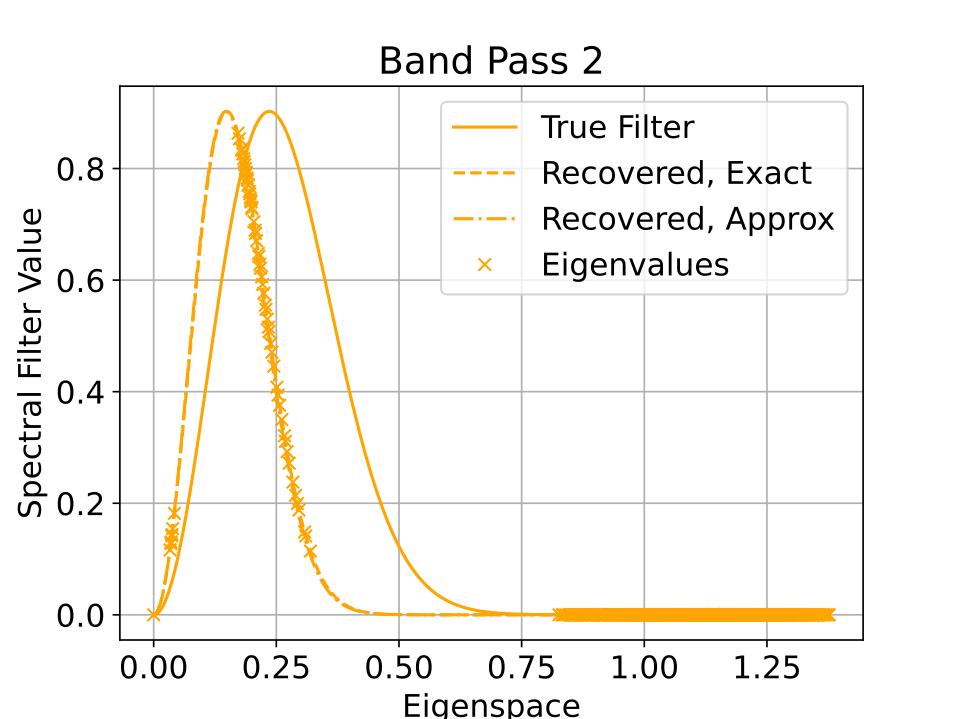}
\end{subfigure}
\begin{subfigure}[b]{0.24\linewidth}
\includegraphics[width=\linewidth, height = 3.cm]{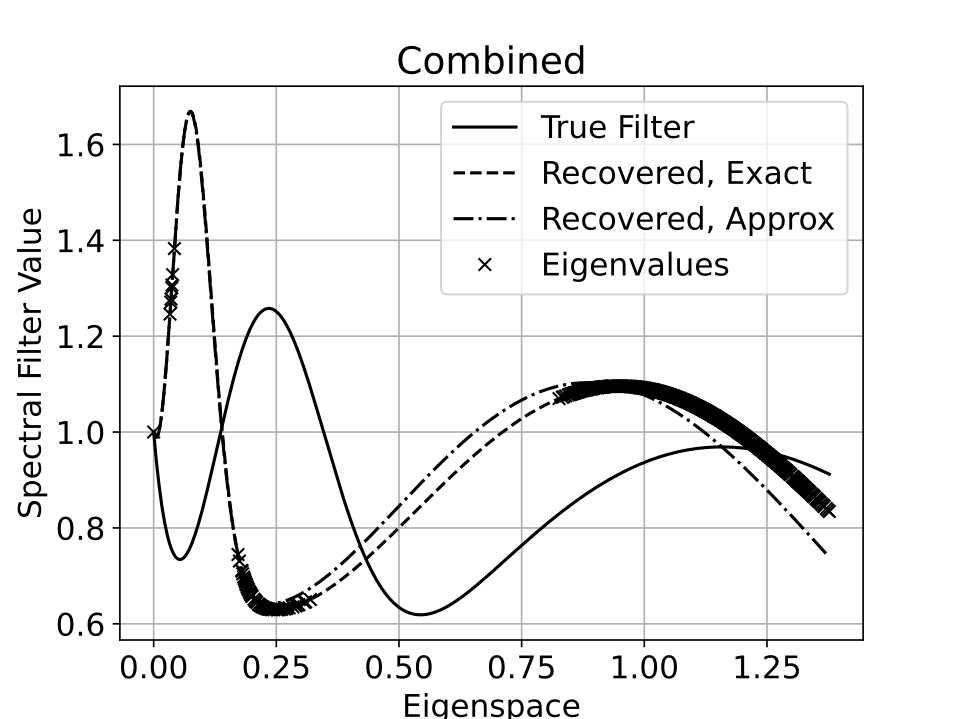}
\end{subfigure}
\begin{subfigure}[b]{0.24\linewidth}
\includegraphics[width=\linewidth, height = 3.cm]{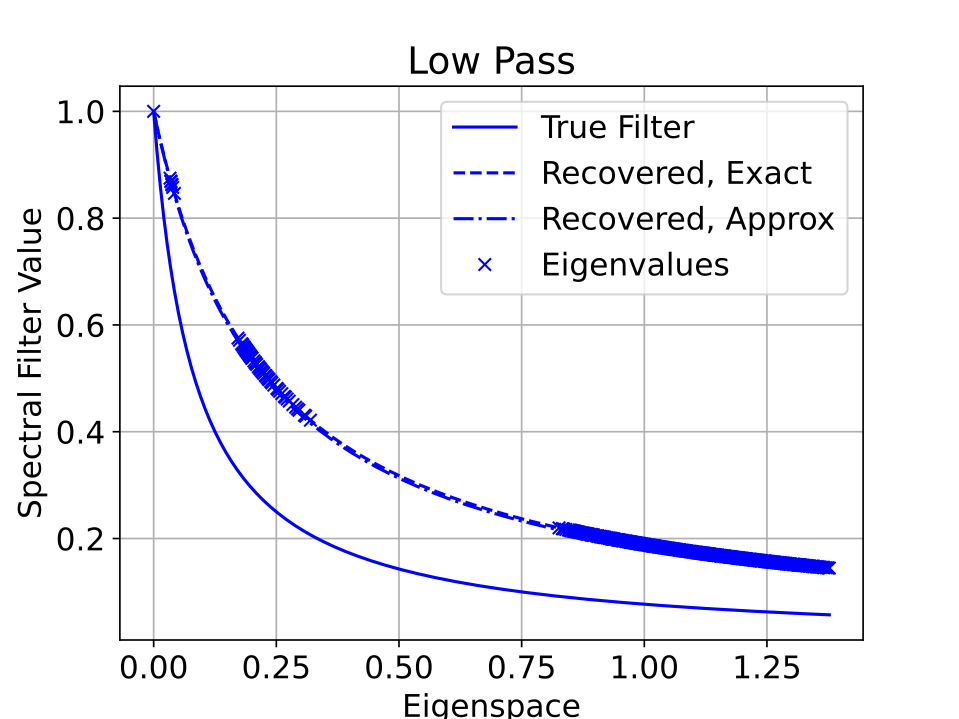}
\end{subfigure}
\begin{subfigure}[b]{0.24\linewidth}
\includegraphics[width=\linewidth, height = 3.cm]{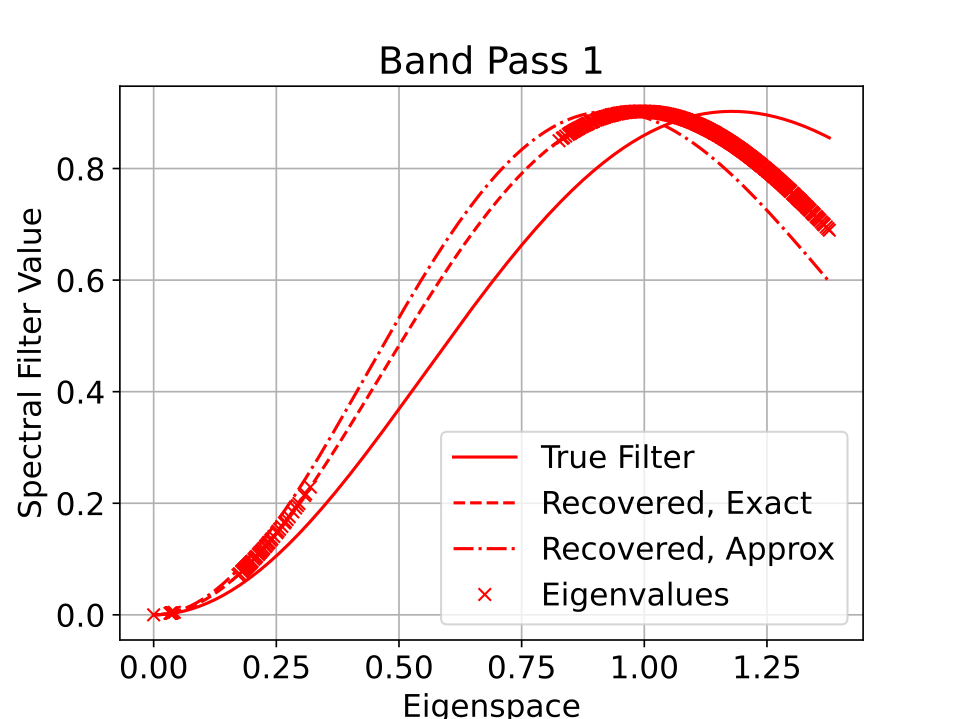}
\end{subfigure}
\begin{subfigure}[b]{0.24\linewidth}
\includegraphics[width=\linewidth, height = 3.cm]{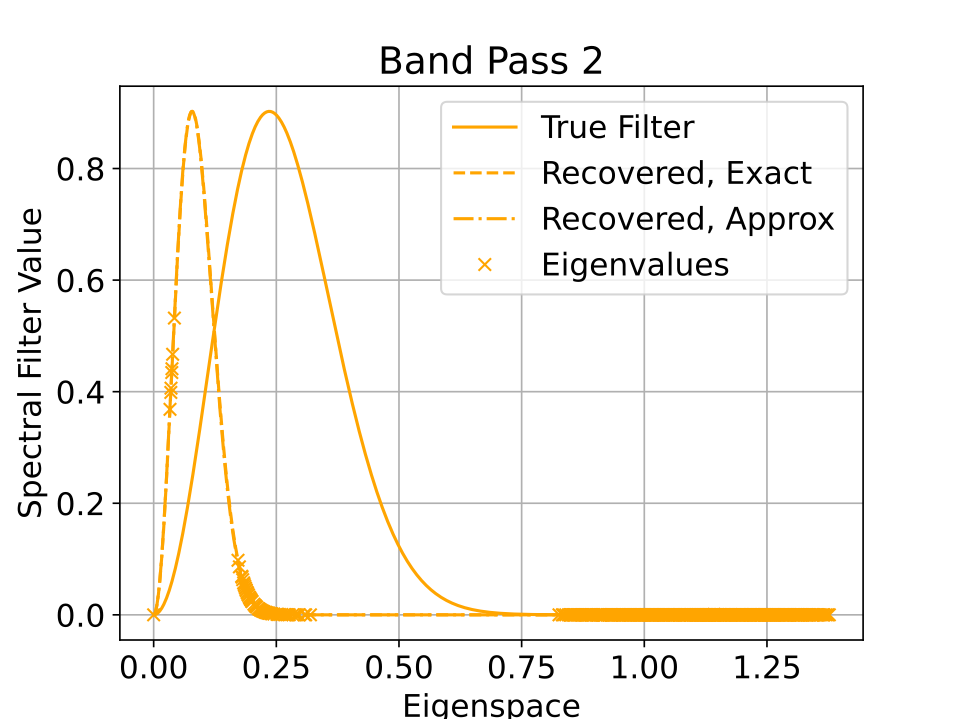}
\end{subfigure}
\begin{subfigure}[b]{0.24\linewidth}
\includegraphics[width=\linewidth, height = 3.cm]{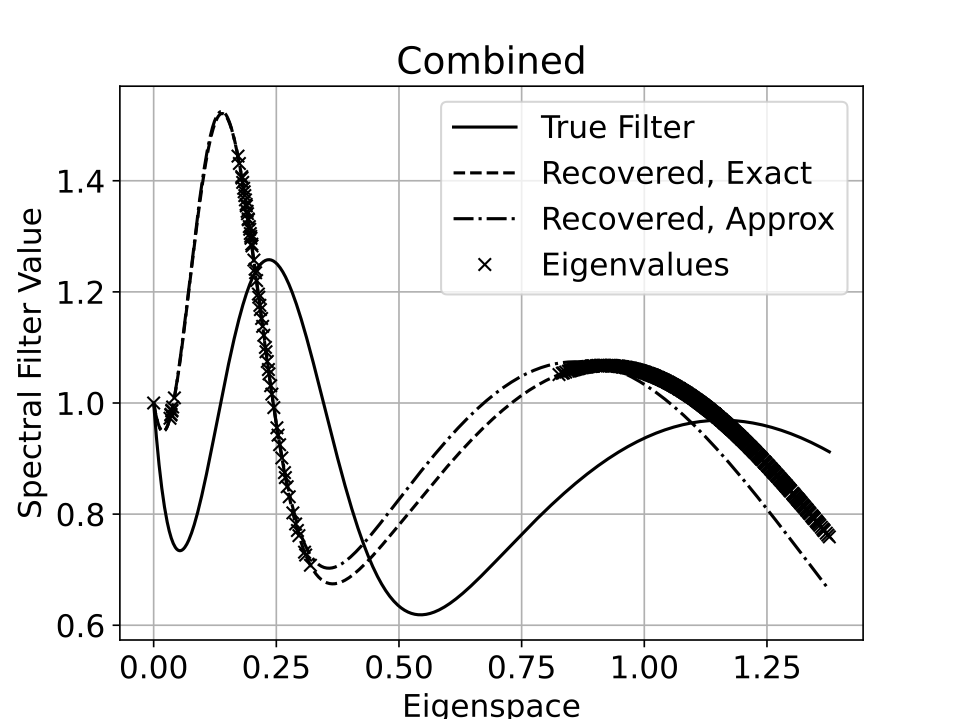}
\end{subfigure}
\begin{subfigure}[b]{0.24\linewidth}
\includegraphics[width=\linewidth, height = 3.cm]{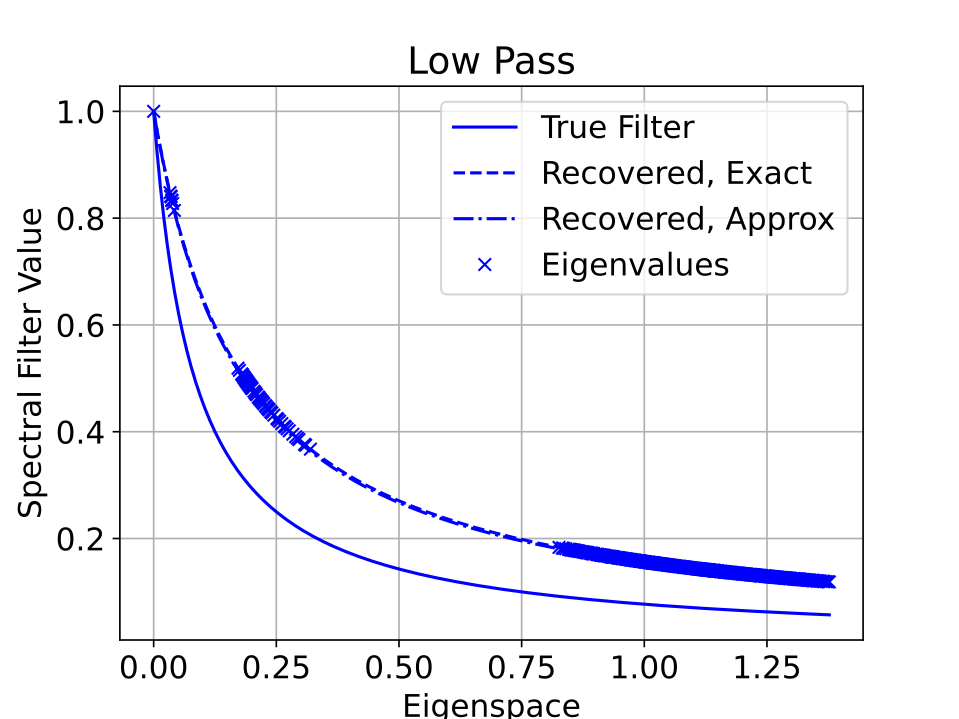}
\end{subfigure}
\begin{subfigure}[b]{0.24\linewidth}
\includegraphics[width=\linewidth, height = 3.cm]{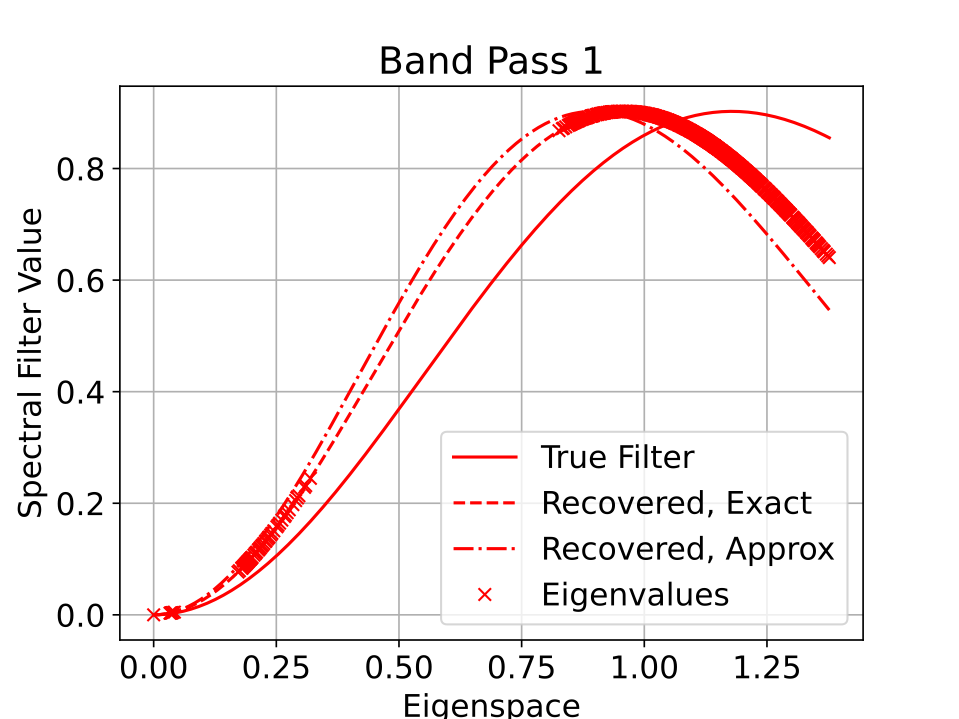}
\end{subfigure}
\begin{subfigure}[b]{0.24\linewidth}
\includegraphics[width=\linewidth, height = 3.cm]{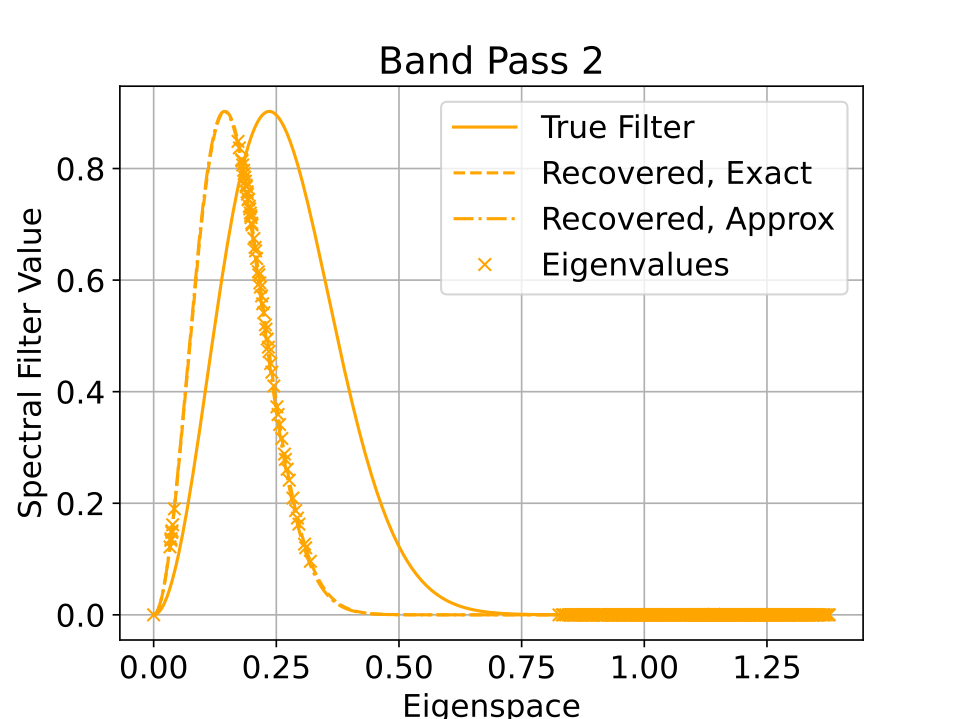}
\end{subfigure}
\caption{Scale recoveries from synthetic experiments using 30\% of nodes as training. Each row is a different random selections of training nodes.}\label{wavelet_transforms_add30}
\end{figure}

\begin{figure}[h]
\centering
\begin{subfigure}[b]{0.24\linewidth}
\includegraphics[width=\linewidth, height = 3.cm]{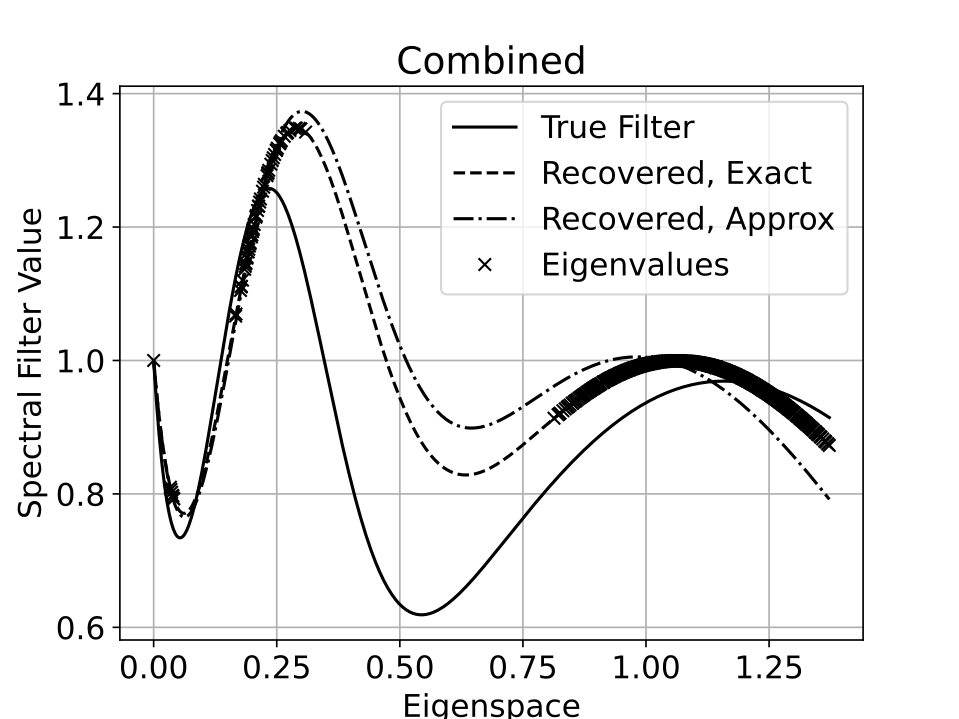}
\end{subfigure}
\begin{subfigure}[b]{0.24\linewidth}
\includegraphics[width=\linewidth, height = 3.cm]{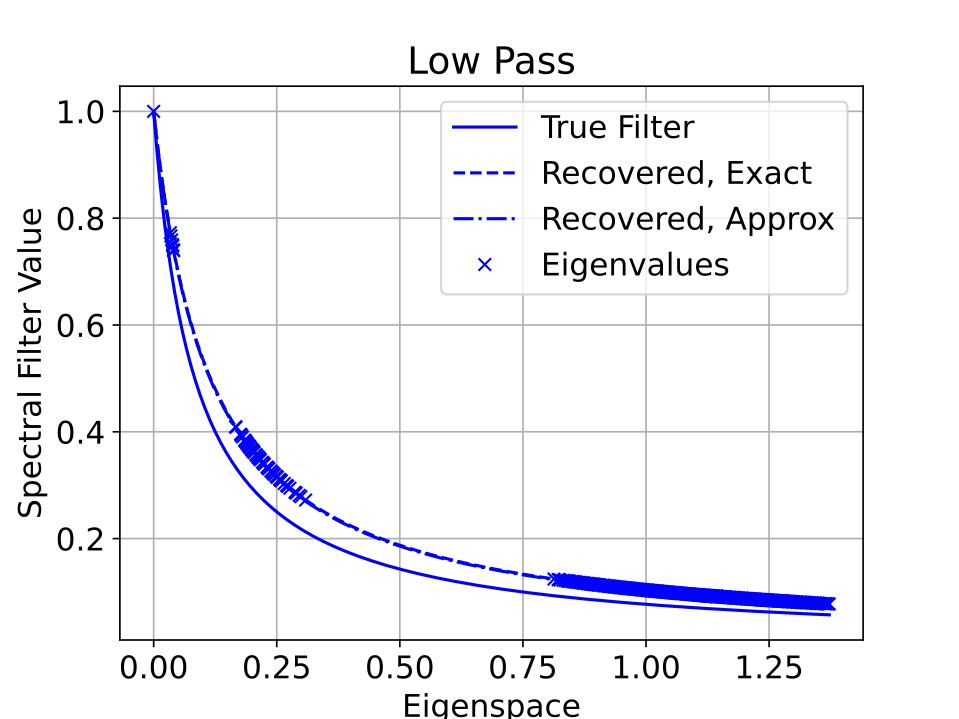}
\end{subfigure}
\begin{subfigure}[b]{0.24\linewidth}
\includegraphics[width=\linewidth, height = 3.cm]{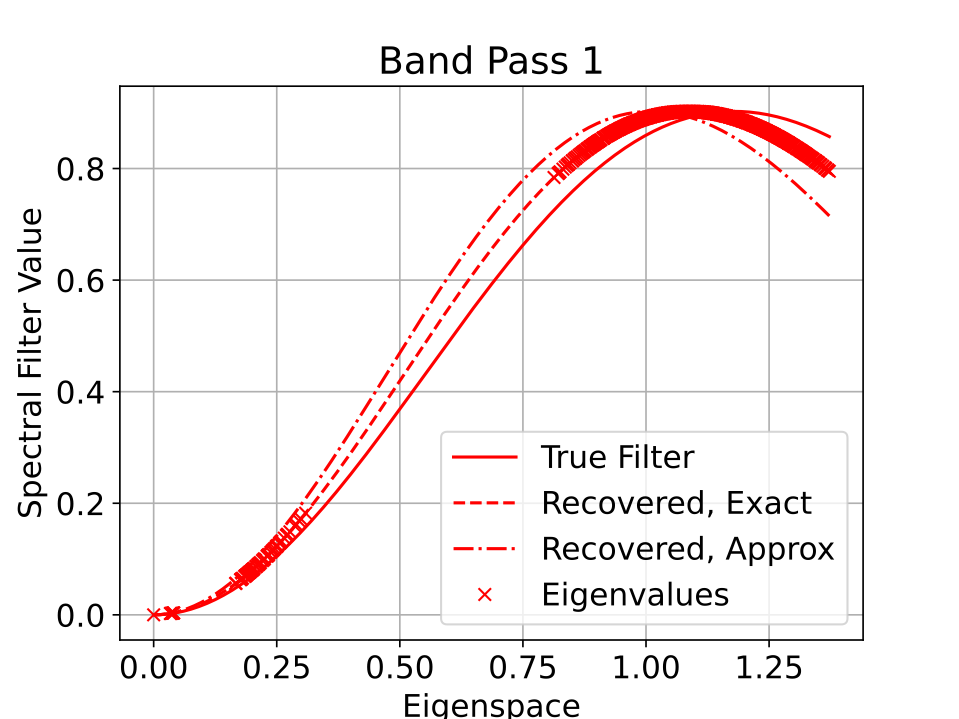}
\end{subfigure}
\begin{subfigure}[b]{0.24\linewidth}
\includegraphics[width=\linewidth, height = 3.cm]{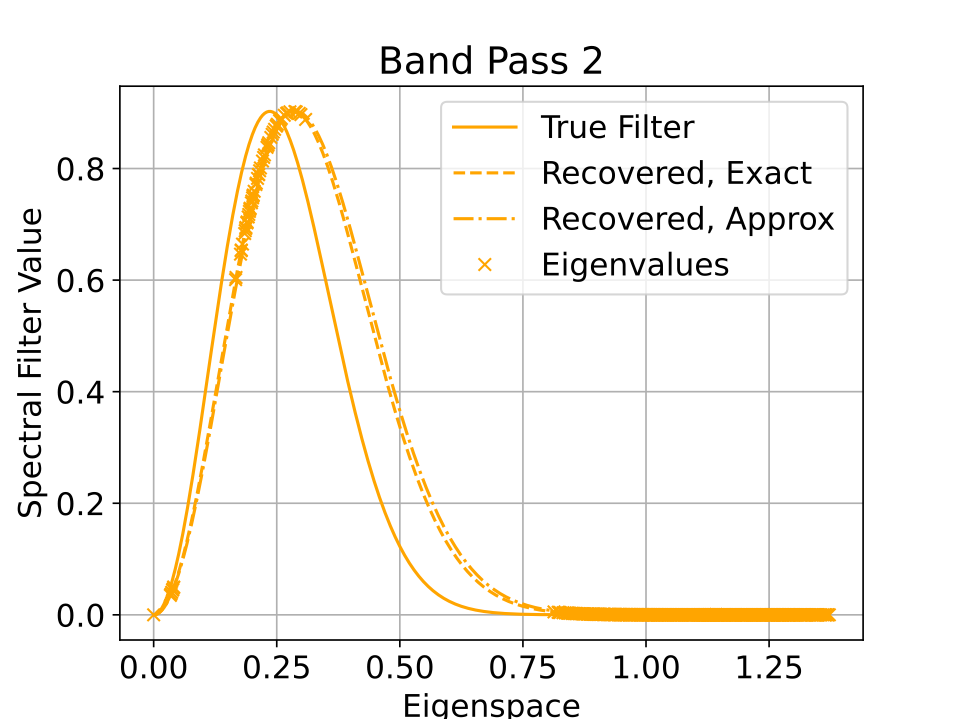}
\end{subfigure}
\begin{subfigure}[b]{0.24\linewidth}
\includegraphics[width=\linewidth, height = 3.cm]{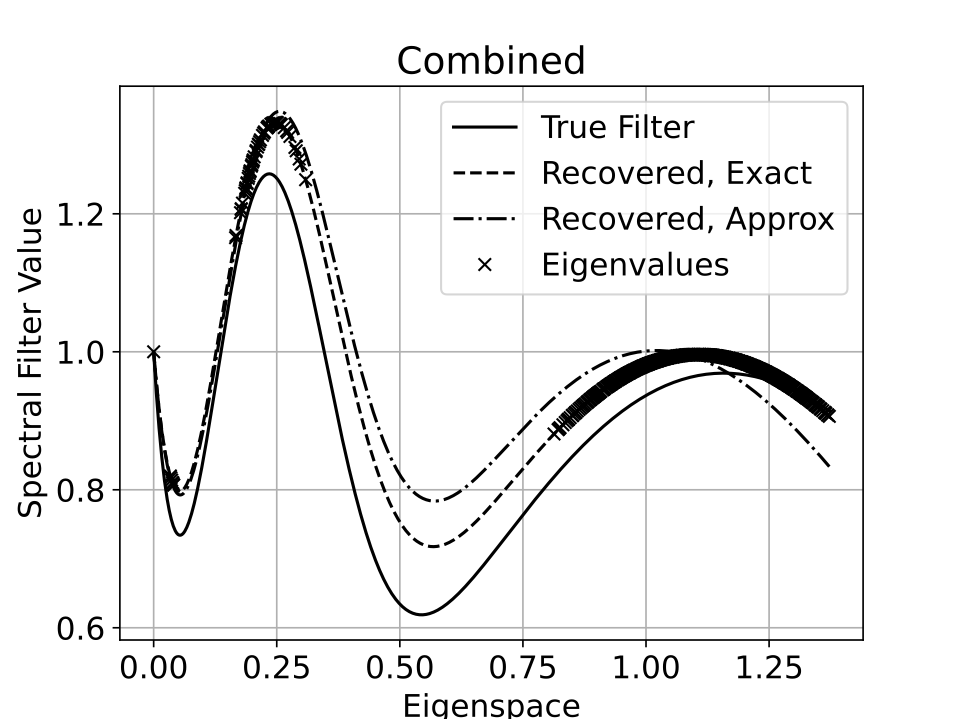}
\end{subfigure}
\begin{subfigure}[b]{0.24\linewidth}
\includegraphics[width=\linewidth, height = 3.cm]{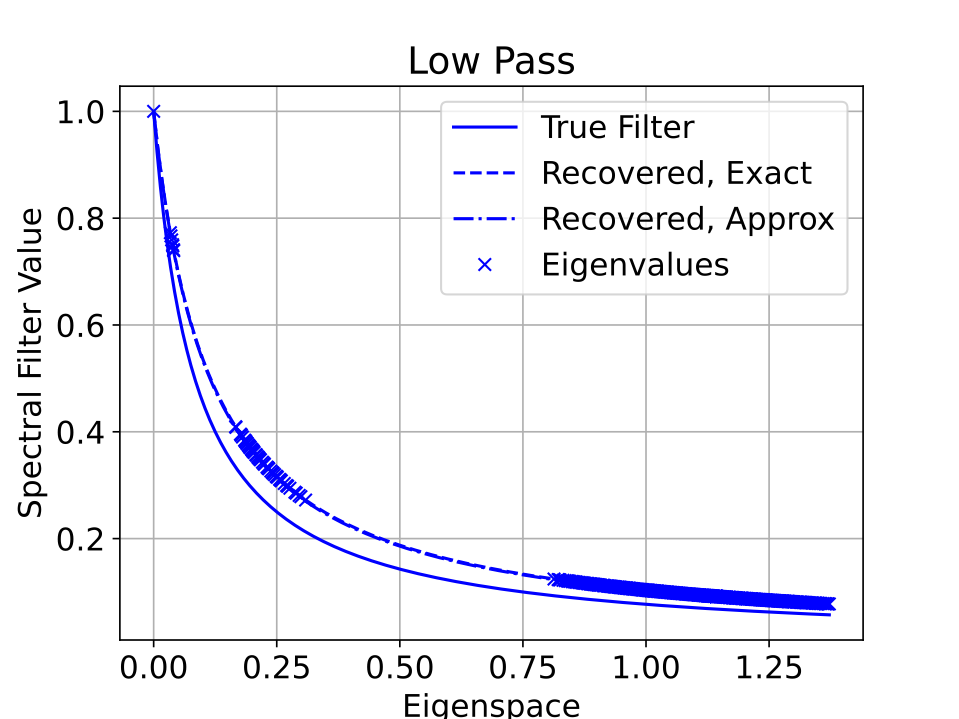}
\end{subfigure}
\begin{subfigure}[b]{0.24\linewidth}
\includegraphics[width=\linewidth, height = 3.cm]{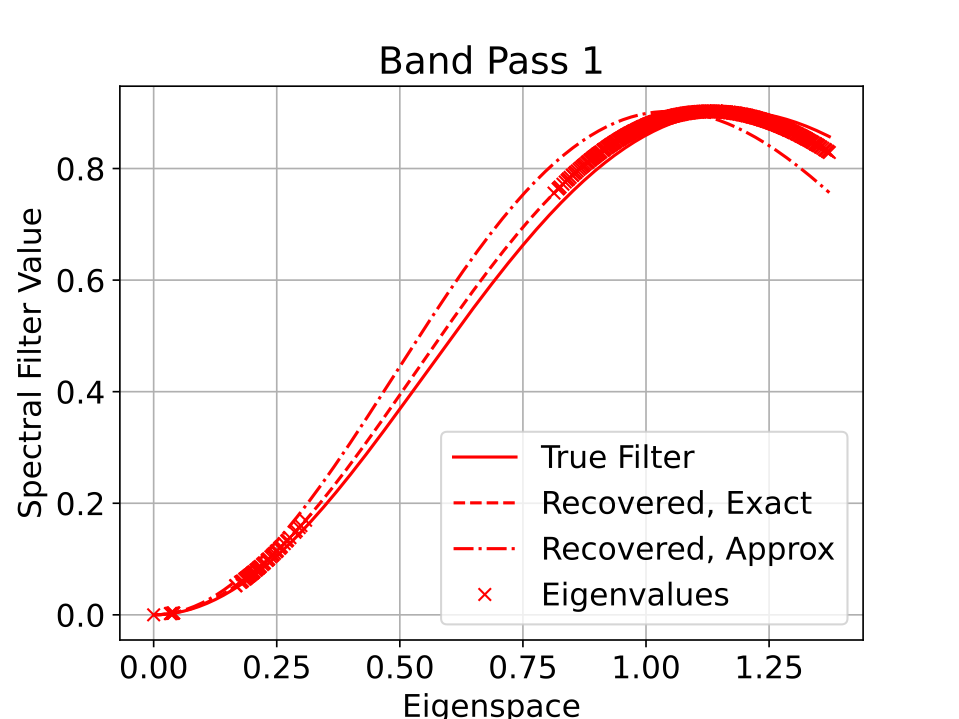}
\end{subfigure}
\begin{subfigure}[b]{0.24\linewidth}
\includegraphics[width=\linewidth, height = 3.cm]{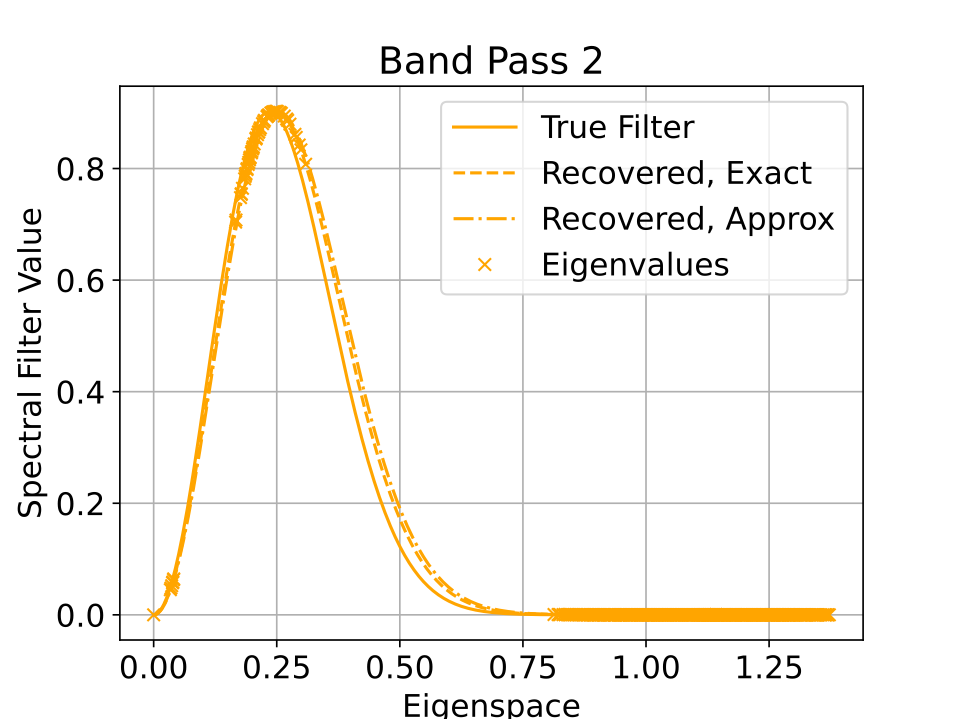}
\end{subfigure}
\begin{subfigure}[b]{0.24\linewidth}
\includegraphics[width=\linewidth, height = 3.cm]{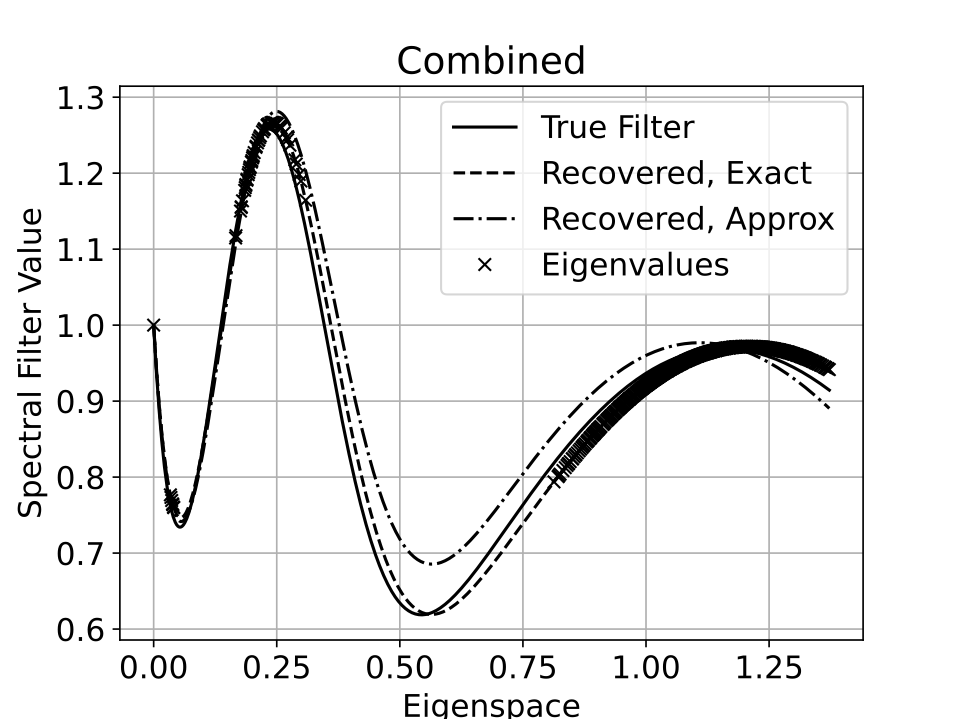}
\end{subfigure}
\begin{subfigure}[b]{0.24\linewidth}
\includegraphics[width=\linewidth, height = 3.cm]{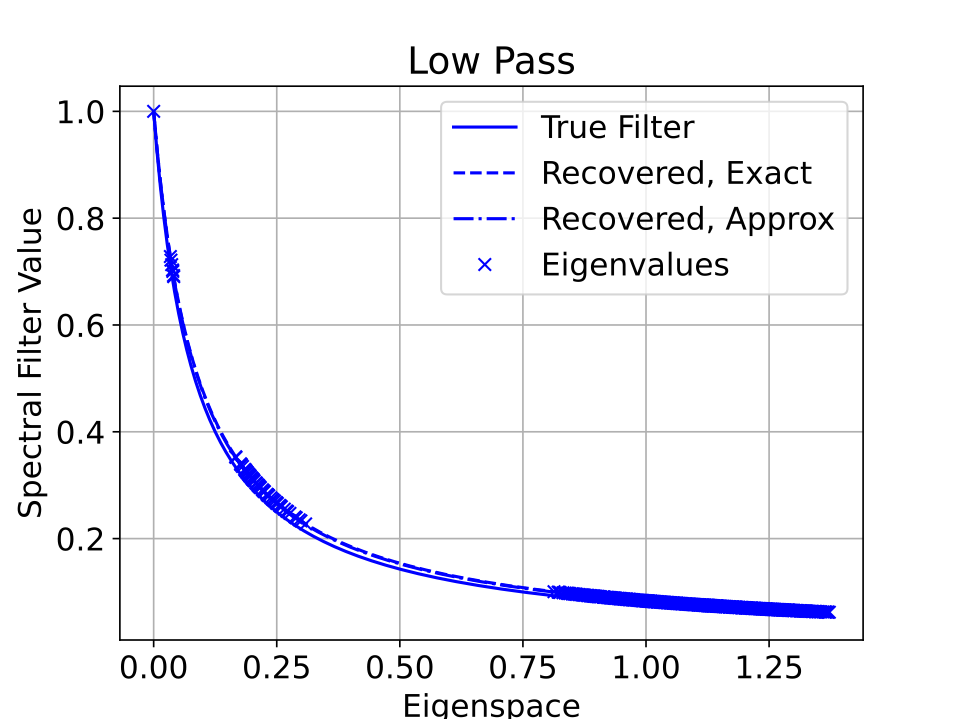}
\end{subfigure}
\begin{subfigure}[b]{0.24\linewidth}
\includegraphics[width=\linewidth, height = 3.cm]{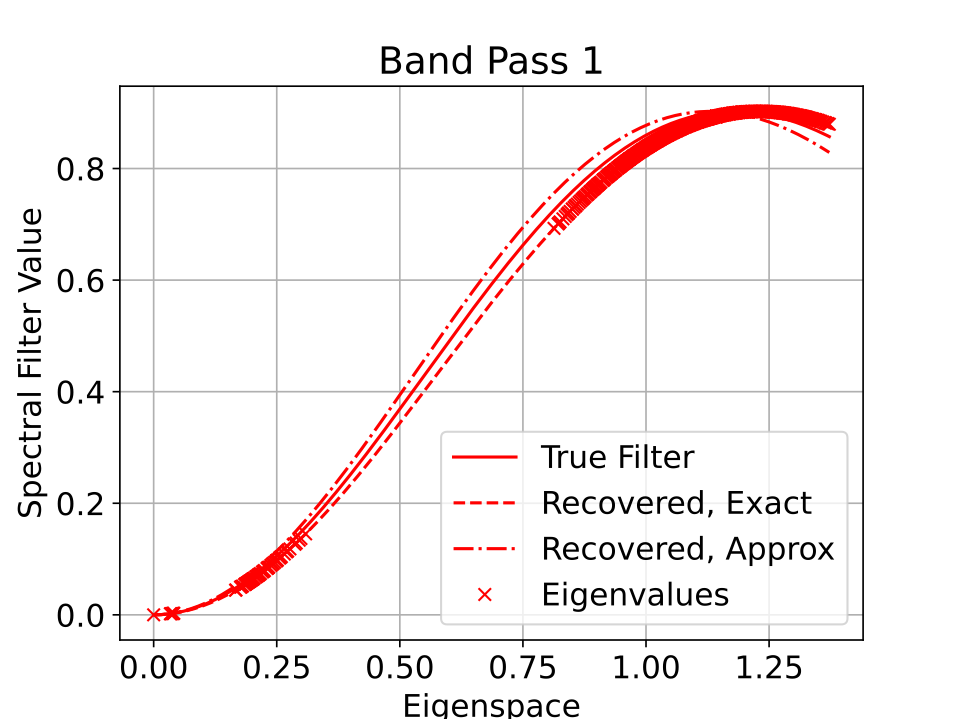}
\end{subfigure}
\begin{subfigure}[b]{0.24\linewidth}
\includegraphics[width=\linewidth, height = 3.cm]{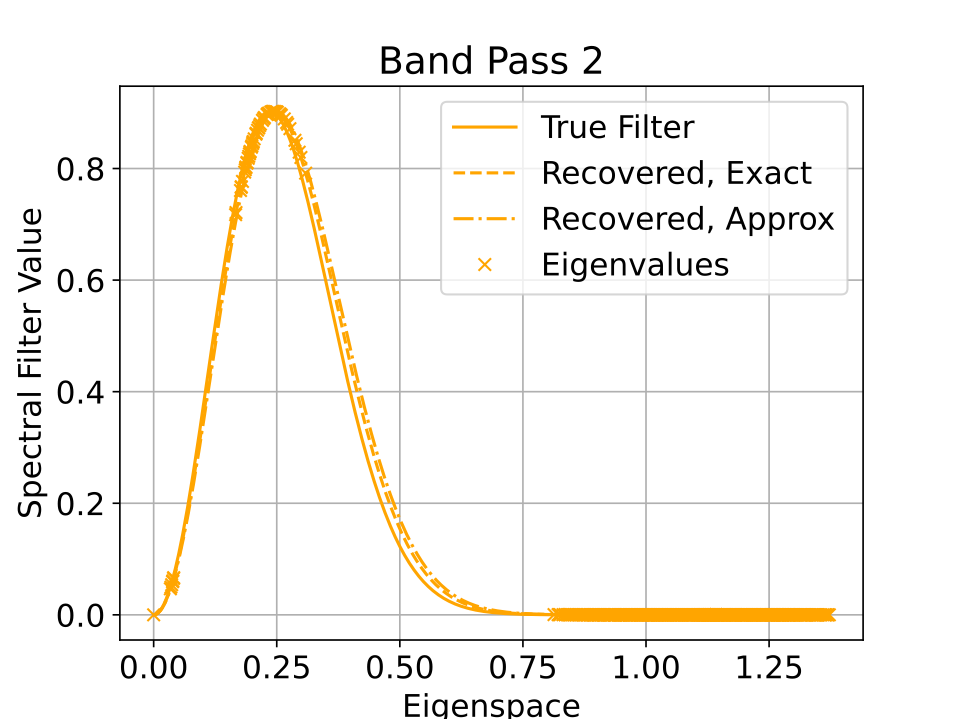}
\end{subfigure}
\caption{Scale recoveries from synthetic experiments using 70\% of nodes as training. Each row is a different random selections of training nodes.} \label{wavelet_transforms_add70}
\end{figure}

The selection of nodes for training will have an effect on the scales we recover. We have presented one particular random split for $50\%$ of nodes used for training in Figure~\ref{fig:mexicanhat} of the main text; In Figures \ref{wavelet_transforms_add10}, \ref{wavelet_transforms_add30} and \ref{wavelet_transforms_add70}, we present the scale recovery results for $10\%$, $30\%$ and $70\%$ of nodes selected for training for three random splits each. We can see the quality of the recoveries improves as the percentage of training nodes increases. Additionally, the approximate recoveries are consistently very close to the exact recoveries, showing the accuracy of our polynomial approximation.

\subsection{Baseline GP Models on Synthetic Data}

\begin{figure}[h]
\centering
\begin{subfigure}[b]{0.32\linewidth}
\includegraphics[width=\linewidth, height = 4.cm]{Figures/mae_mh_mh_norm.png}
\caption{WGGP}\label{mae2}
\end{subfigure}
\begin{subfigure}[b]{0.32\linewidth}
\includegraphics[width=\linewidth, height = 4.cm]{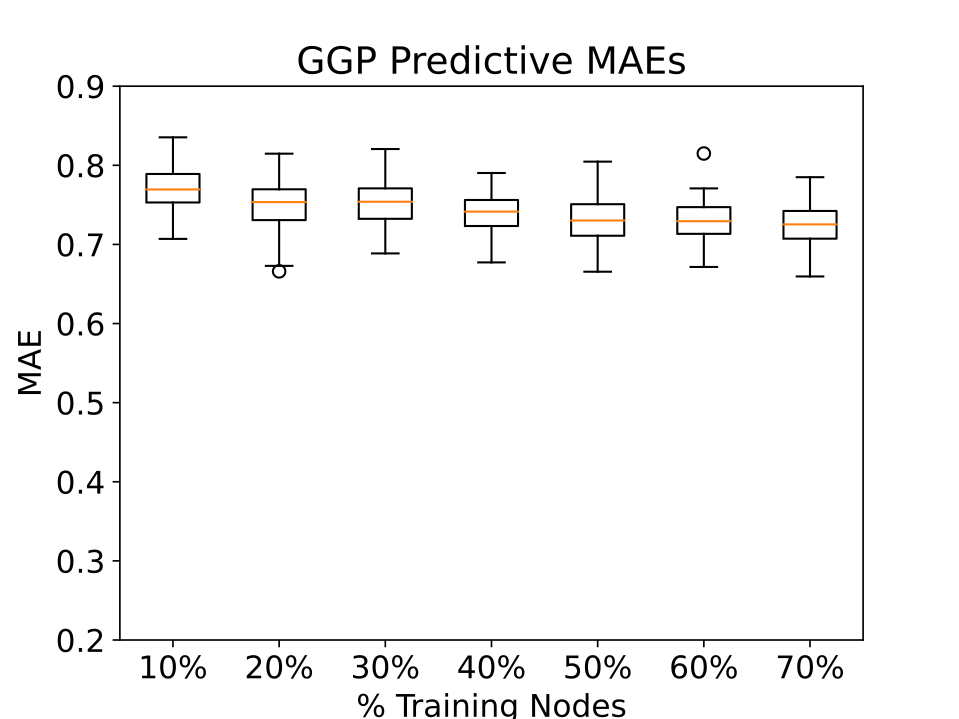}
\caption{GGP}\label{mae1}
\end{subfigure}
\begin{subfigure}[b]{0.32\linewidth}
\includegraphics[width=\linewidth, height = 4.cm]{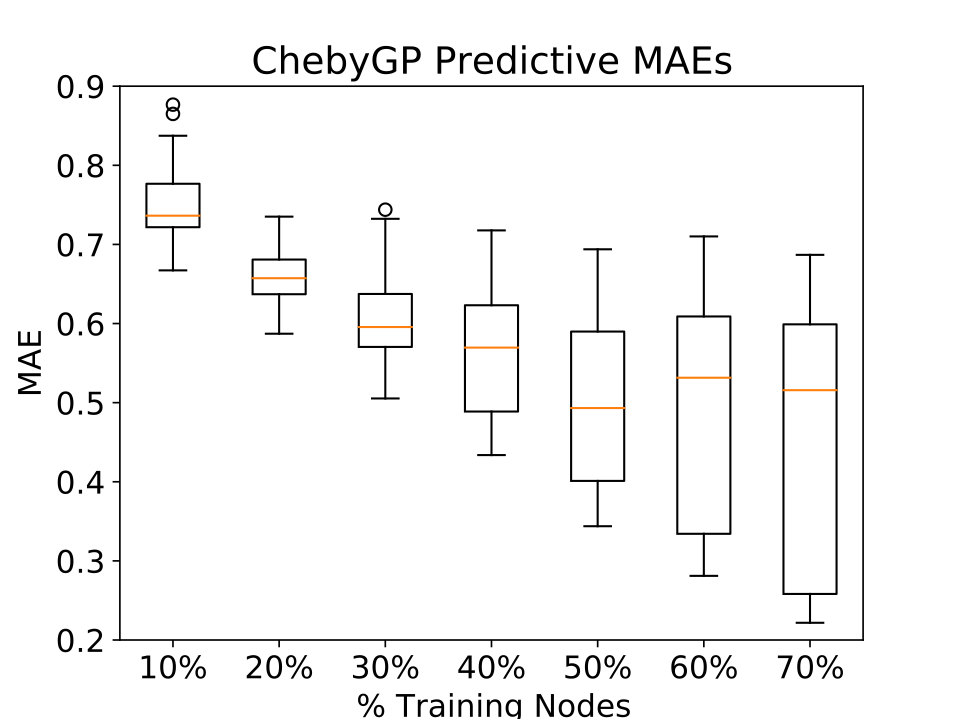}
\caption{ChebGP}\label{mae2}
\end{subfigure}
\caption{WGGP prediction MAE on synthetic data (a and identical to Figure 3b in main text) compared to MAEs of baseline GP models, GGP (b) and ChebGP (c).}\label{maes_ggp_cheby}
\end{figure}

We also evaluate the baseline GP models from Section~\ref{sec:experiments} on synthetic data. The graph neural network models were not compared against as they require a validation set of nodes, which are not assigned and would make an unfair comparison. We use GGP and ChebGP to make predictions on the synthetically generated signals, with the MAEs presented in Figure \ref{maes_ggp_cheby}.

The results in Figure \ref{maes_ggp_cheby} show that the GGP only improves marginally with additional training data, indicating the model's inability to capture multi-scale information. ChebGP, which uses Chebyshev polynomials for approximations, does approximate a multi-scale spectral wavelet function, but we can see by the means and quantiles of the boxplots that they are less consistent in producing low MAEs compared to our polynomial approximation. As the number of training nodes increases, the model should be able to capture the different scales more accurately; however, the wider quantiles indicate the Chebyshev approximation is less consistent in producing accurate recoveries.

\subsection{Performance on Synthetic Data Generated Using Different Ground Truth Wavelets}

\begin{figure}[h]
\centering
\begin{subfigure}[b]{0.45\linewidth}
\centering
\includegraphics[height=4.cm]{Figures/mae_mh_mh_norm.png}
\caption{MAE between predicted and ground truth synthetic signal when using a Mexican Hat wavelet in the inference GP and a \textbf{Mexican Hat} wavelet in the data generating GP.}\label{subfig:mh_pred}
\end{subfigure}
\hfill
\begin{subfigure}[b]{0.45\linewidth}
\centering
\includegraphics[height=4.cm]{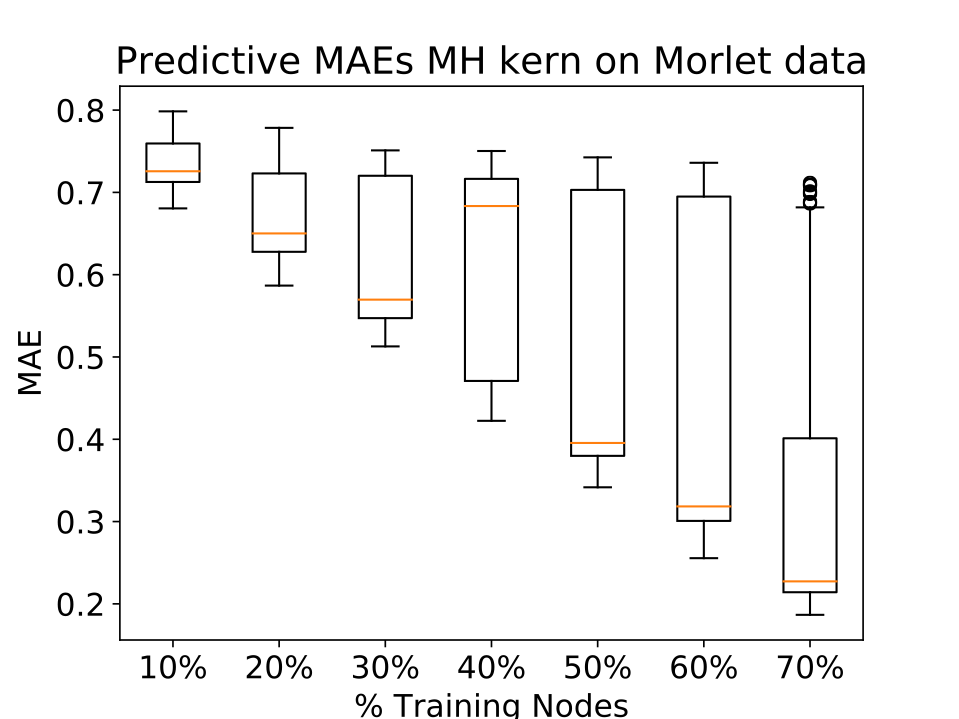}
\caption{MAE between predicted and ground truth synthetic signal when using a Mexican Hat wavelet in the inference GP and a \textbf{Morlet} wavelet in the data generating GP.}\label{subfig:morlet_pred}
\end{subfigure}

\begin{subfigure}[b]{0.45\linewidth}
\centering
\includegraphics[height=4.cm]{Figures/mae_mh_mh_eig_norm.png}
\caption{MAE between reconstructed and ground truth filter when using a Mexican Hat wavelet in the inference GP and a \textbf{Mexican Hat} wavelet in the data generating GP.}\label{subfig:mh_eigvals}
\end{subfigure}
\hfill
\begin{subfigure}[b]{0.45\linewidth}
\centering
\includegraphics[height=4.cm]{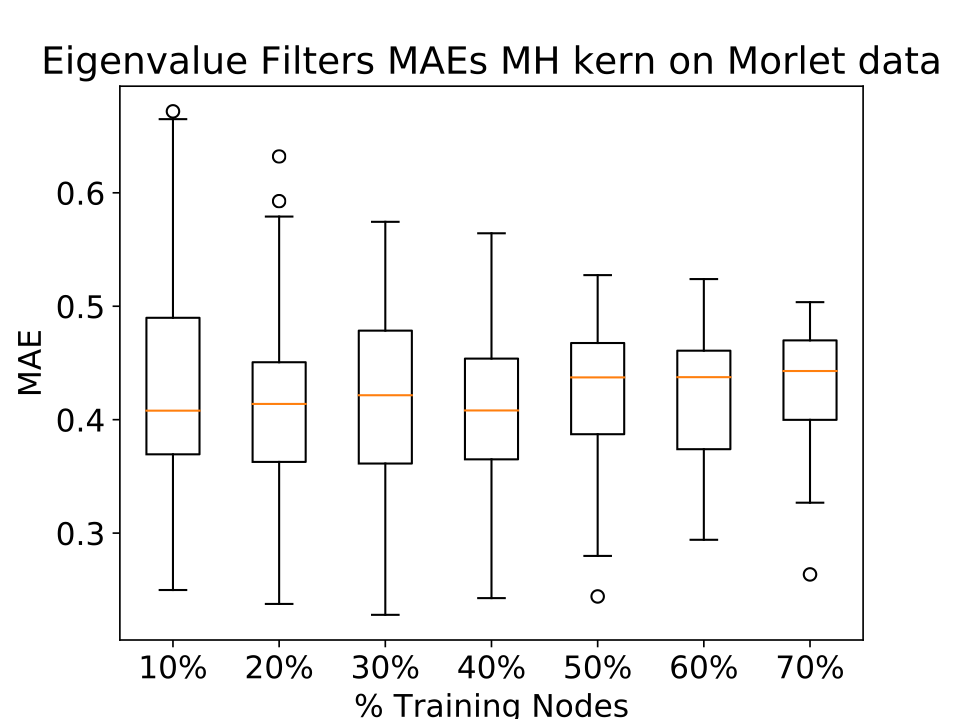}
\caption{MAE between reconstructed and ground truth filter when using a Mexican Hat wavelet in the inference GP and a \textbf{Morlet} wavelet in the data generating GP.}\label{subfig:morlet_eigvals}
\end{subfigure}
\caption{Comparing the prediction and filter MAEs for different fractions of nodes used for trainig when the choice of mother wavelet match or do not match between the inference GP and the data generating GP.}\label{fig:different_wavelet}
\end{figure}

The synthetic setting described in Section~\ref{sec:experiments} uses the Mexican Hat kernel in both the inference GP and the data generating model. We now study the case where there is a mismatch in mother wavelet between the inference GP and the data generating model. In Figure\ref{fig:different_wavelet} we always use a Mexican Hat wavelet for the inference GP and compare the case of using a Mexican Hat wavelet (Figures \ref{subfig:mh_pred} and \ref{subfig:mh_eigvals}) versus a Morlet wavelet (Figures \ref{subfig:morlet_pred} and \ref{subfig:morlet_eigvals}) in the data generating GP both in terms of prediction MAE (Figures \ref{subfig:mh_pred} and \ref{subfig:morlet_pred}) and MAE of reconstructed filter compared to ground truth filter (Figures \ref{subfig:mh_eigvals} and \ref{subfig:morlet_eigvals}).

\subsection{WGGP without Feature Space Kernel}

To measure the importance of the feature space kernel, we repeat experiments with WGGP on Cora and Citeseer with the feature space kernel $K_\Psi(\bd{x}_i, \bd{x}_j) = \delta_{ij}$ set to the identity. As a result, the model classifies nodes no longer also based on node features but on graph structure alone. We compare the results to those of the full WGGP model in Table~\ref{tab:no_feat_kernel}.
As expected, the performance of the model drops decisively when removing the dependence on the node features, demonstrating the importance of the node feature kernel for the predictive performance.

\begin{table}[H]
    \centering
    \begin{tabular}{lcccccc}
        \toprule
        \textbf{Method} & \textbf{Cora} & \textbf{Citeseer}\\
        \midrule
        \textbf{WGGP} & $84.7$ & $70.8$ \\
        \textbf{WGGP without node features} & $71.9$ & $47.7$ \\
        \bottomrule
        \vspace{0.5mm}
    \end{tabular}
    \caption{Classification accuracy of the WGGP model with and without the node feature kernel. When removing the node feature kernel, the predictive performance drops by more than $10\%$ for both data sets.}\label{tab:no_feat_kernel}
\end{table}

\subsection{ELBO Plots}

\begin{figure*}[h]
\centering
\begin{subfigure}[b]{0.3\textwidth}
    \centering
    \includegraphics[width=\textwidth]{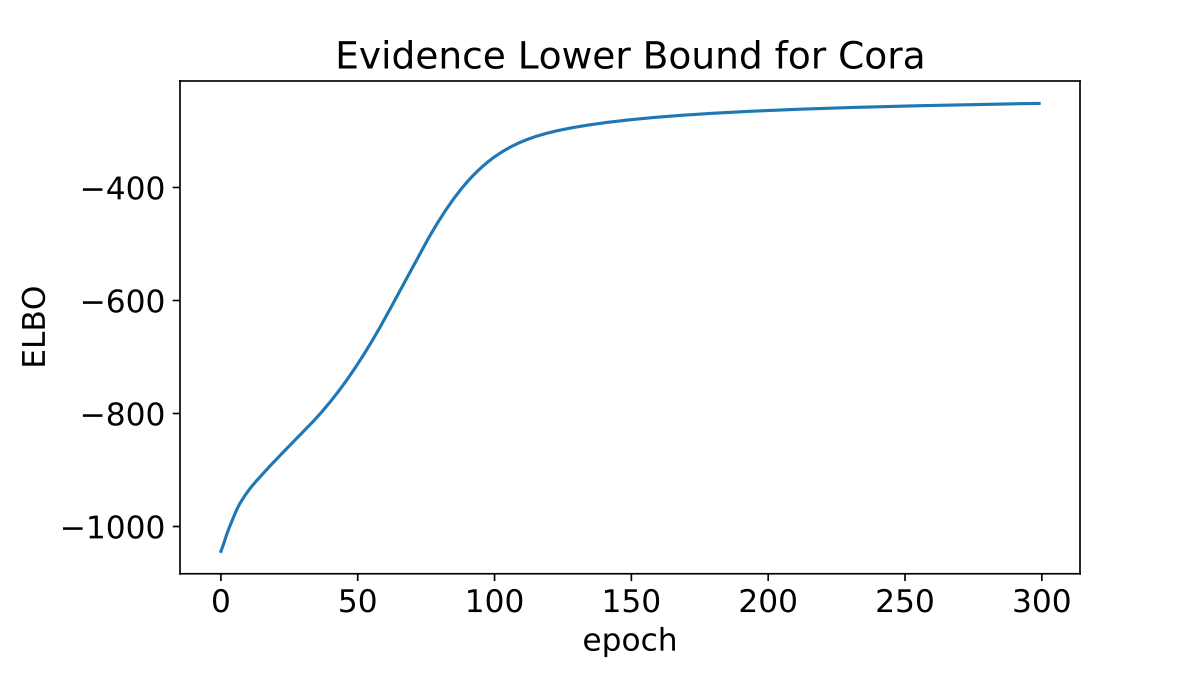}
    \caption{Cora}
    \label{fig:elbo_cora} 
\end{subfigure}
\begin{subfigure}[b]{0.3\textwidth}
    \centering
    \includegraphics[width=\textwidth]{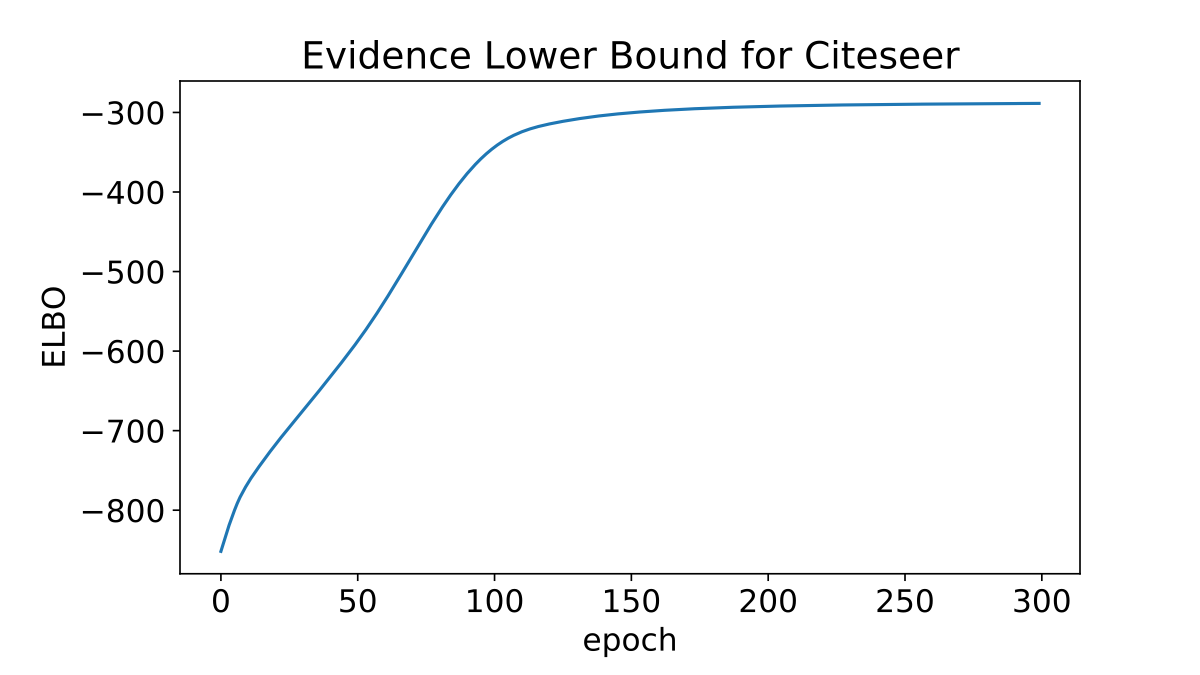}
    \caption{Citeseer}
    \label{fig:elbo_citeseer}
\end{subfigure}
\begin{subfigure}[b]{0.3\textwidth}
    \centering
    \includegraphics[width=\textwidth]{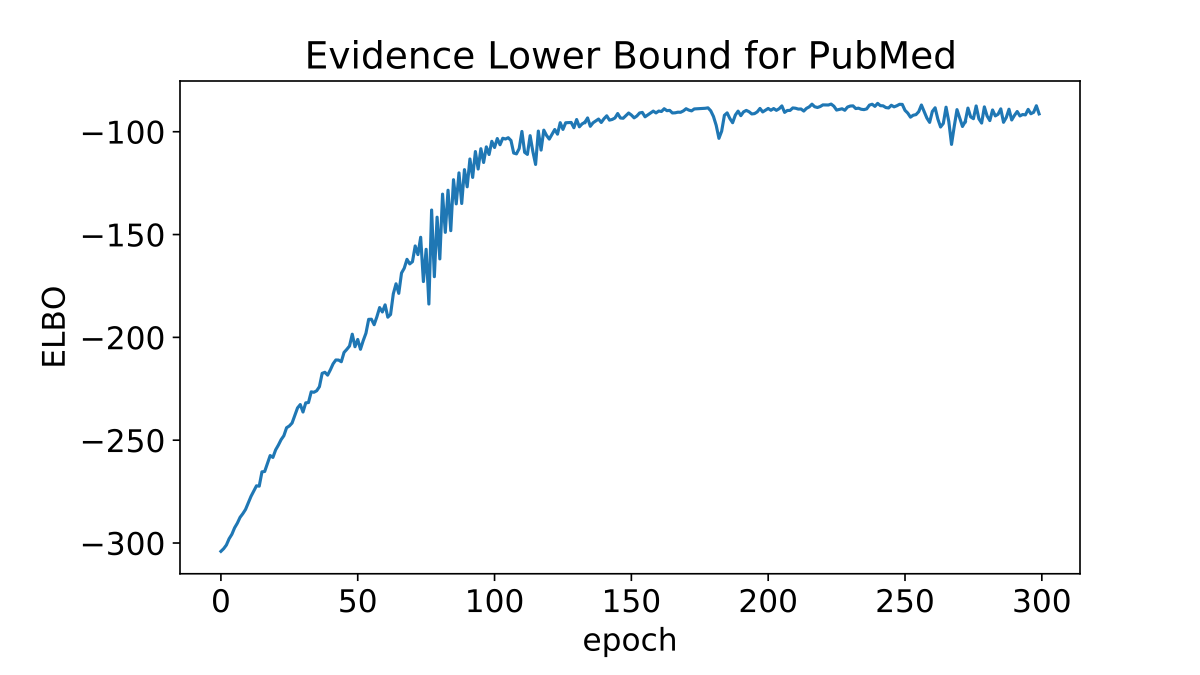}
    \caption{Pubmed}
    \label{fig:elbo_pubmed}
\end{subfigure}
\caption{Value of the ELBO during training over time.}
\label{fig:elbo_plots}
\end{figure*}

As described in Section 5, early stopping is performed based on the ELBO. To check convergence, we show how the ELBO varies from epoch to epoch in Figure~\ref{fig:elbo_plots}. Note that the ELBO curve for the PubMed data set is non-monotonic as stochastic optimisation is employed during training.

\subsection{Robustness Analysis}

We perform a robustness analysis examining how the model performance changes as we vary different parts of the model or training setup, while keeping everything else as described in Section 5. 

\paragraph{Data Split} 
In a first experiment, we use $10$ different data splits for Cora and Citeseer that retain the uniform distribution of classes and re-run the model with otherwise equal hyperparameters. The average performance across the $10$ data splits is reported in Table~\ref{tab:robustness} together with one standard deviation. We find that the model perofrmance only varies modestly for both data sets and the performance remains comparable to the one achieved on the public data split.

\paragraph{Number of scales} 
We also analyse how the model performance varies when using different number of scales in the model, ranging from using only a low-pass filter to also including $4$ band pass filters. The results are again reported in Table~\ref{tab:robustness} with the standard deviation over the $4$ different setups ($0$-$4$ scales), showing that the model accuracy varies only slightly when using different number of filters.

\paragraph{Hyperparameter initialisations}
Finally, we repeat the experiments with random initialisations of the scale hyperparameters.
The results with their standard deviation over $10$ different initialisations (Table~\ref{tab:robustness}) demonstrate the model's robustness to different hyperparameter initialisations.

\begin{table}[H]
    \centering
    \begin{tabular}{lcccccc}
        \toprule
        \textbf{Method} & \textbf{Cora} & \textbf{Citeseer}\\
        \midrule
        \textbf{WGGP with varying data splits} & $82.4 \pm 1.1$ & $67.8 \pm 2.7$ \\
        \textbf{WGGP with varying number of scales} & $84.7 \pm 0.2$ & $70.6 \pm 0.2$ \\
        \textbf{WGGP with varying hyperparameter initalisations} & $84.2 \pm 0.4$ & $71.0 \pm 0.6 $ \\
        \bottomrule
        \vspace{0.5mm}
    \end{tabular}
    \caption{Results of the robustness analysis of the WGGP model when varying the data split, the number of scales, or the scale hyperparameter initialisations.}\label{tab:robustness}
\end{table}


\section{DATA SET STATISTICS}

\begin{table}[H]
    \centering
    \begin{tabular}{lcccccc}
        \toprule
        \textbf{Data} & Type & $\mathbf{N_{nodes}}$ & $\mathbf{N_{edges}}$ & $\mathbf{N_{label\_cat}}$ & $\mathbf{D_{features}}$ & \textbf{Label Rate}\\
        \midrule
        \textbf{Cora} & Citation & 2,708 & 5,429 & 7 &  1,433 &  0.052\\
        \textbf{Citeseer} & Citation & 3,327 & 4,732 & 6 &  3,703 &  0.036\\
        \textbf{PubMed} & Citation & 19,717 & 44,338 & 3 & 500 &  0.003\\
        \bottomrule
        \vspace{0.5mm}
    \end{tabular}
    \caption{Summary of citation networks for node classification experiments.}
\end{table}

\section{COMPUTING PLATFORM AND CODE IMPLEMENTATION}

The experiments were performed using 
Xeon W-2133 12GB NVIDIA GTX 1080 Ti and 48GB NVIDIA Quadro RTX 8000.

The code for reproducing the results in the paper has been submitted as part of the supplementary material.

\section{FIGURES}

\end{document}


%

%

\onecolumn
\aistatstitle{Adaptive Gaussian Processes on Graphs via Spectral Graph Wavelets --- 
Supplementary Materials}
\appendix

\section{Wavelet Transform}

\begin{figure}[h]
\centering
\begin{subfigure}[b]{0.33\linewidth}
\includegraphics[width=\linewidth, height = 3.8cm]{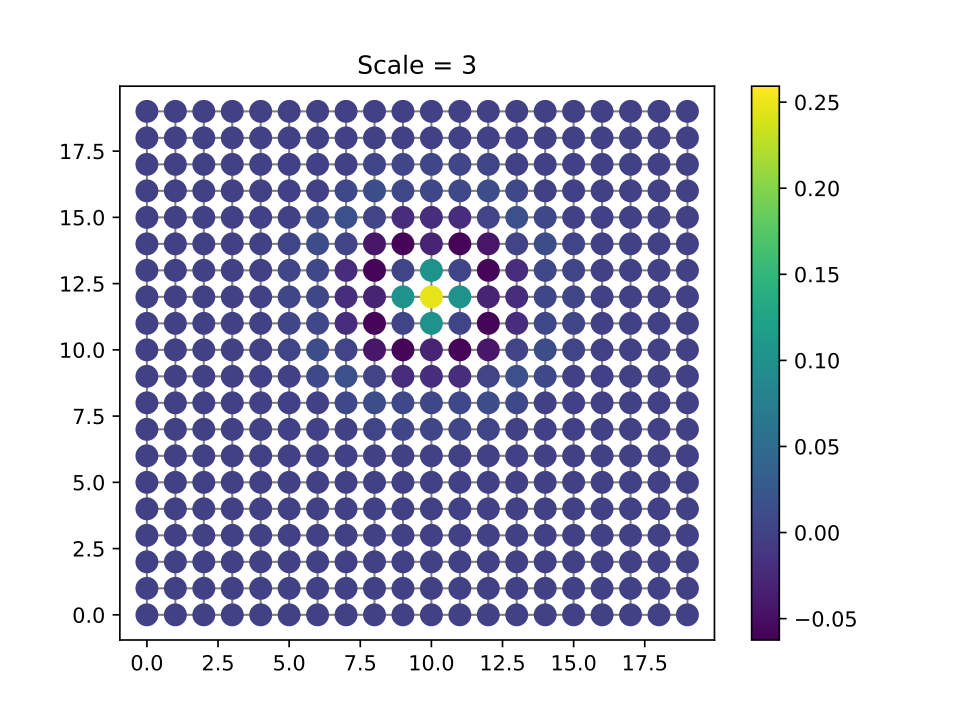}
\caption{}
\end{subfigure}
\hspace{-3mm}
\begin{subfigure}[b]{0.33\linewidth}
\includegraphics[width=\linewidth, height = 3.8cm]{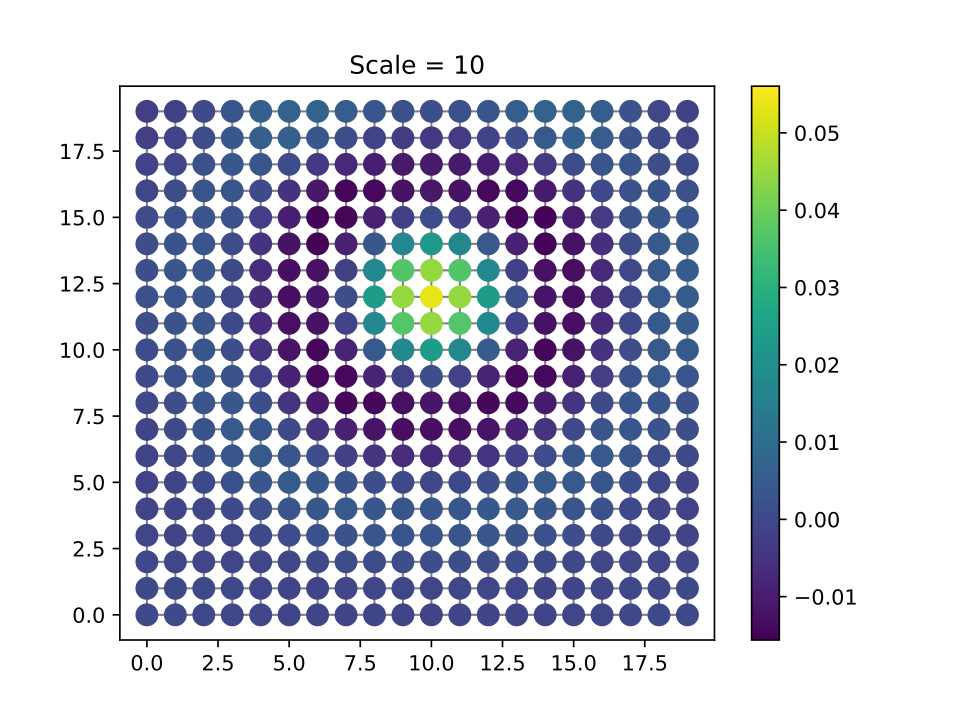}
\caption{}
\end{subfigure}
\hspace{-3mm}
\begin{subfigure}[b]{0.33\linewidth}
\includegraphics[width=\linewidth, height = 3.8cm]{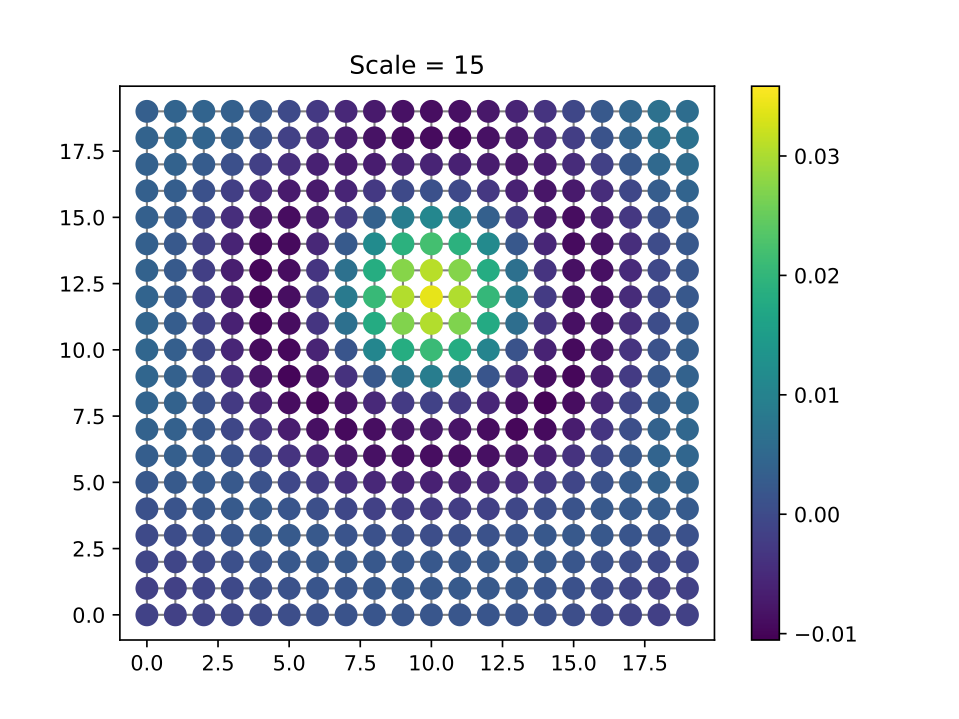}
\caption{}
\end{subfigure}
\caption{The Mexican Hat wavelet transform of a $\delta$ signal on a regular grid graph. The grid simulates a Euclidean domain to demonstrate the neighbourhoods more clearly at different scales.}\label{wavelet_transforms_grid}
\end{figure}

By applying the wavelet transforms to a impulse function (centred at yellow node) on the graph, we can see how the wavelets at different scales spread around the centre node, capturing different ranges of neighbourhoods. When applied to a regular grid graph, the pattern resembles that of the Euclidean domain. In Figure \ref{wavelet_transforms_grid} we apply the Mexican Hat wavelet transform with various scales to show the different ranges of neighbourhoods. The neighbours are weighted continuously with intensity becoming zero once beyond a certain proximity. Thus, by using different scales, we can capture different ranges of neighbourhood information.

\section{Additional Experimental Results}

\subsection{Synthetic Scale Recovery Experiments and Implementation Details}

We run our synthetic experiments multiple times to show the overall behaviour of the model. We sample the labels for the nodes by Eq. (17) 
100 times; for each sample, we also randomly select a set of nodes to use as training. The hyperparameters are optimised as part of the training process. For each set of training labels, we test 20 different initializations and use the converged values that lead to the lowest loss using the GPflow package.

\begin{figure}[h]
\centering
\begin{subfigure}[b]{0.24\linewidth}
\includegraphics[width=\linewidth, height = 3.2cm]{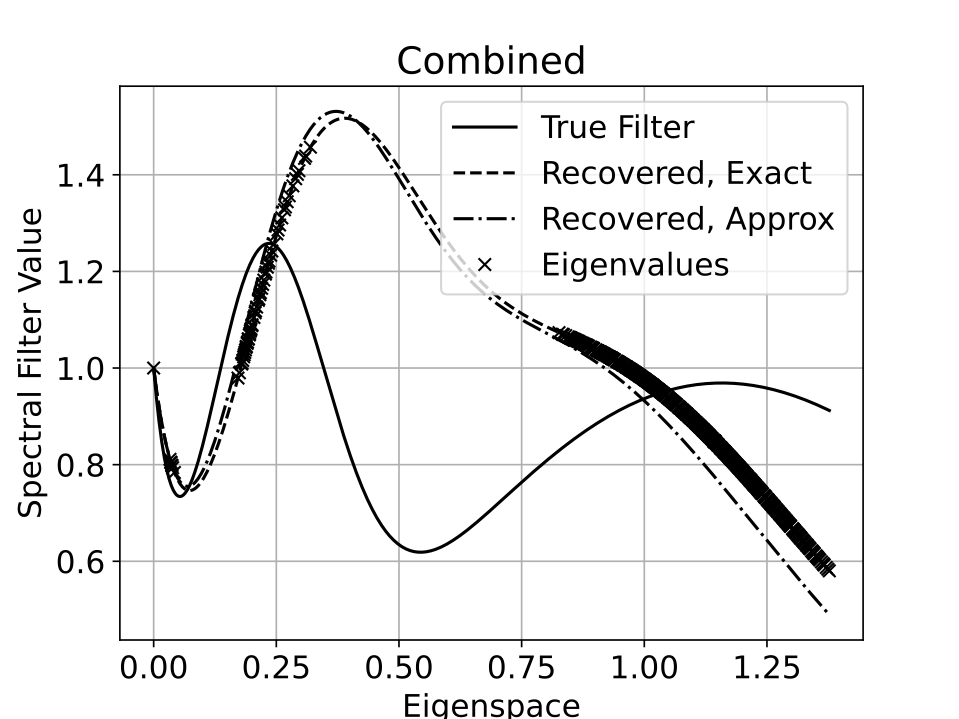}
\end{subfigure}
\begin{subfigure}[b]{0.24\linewidth}
\includegraphics[width=\linewidth, height = 3.2cm]{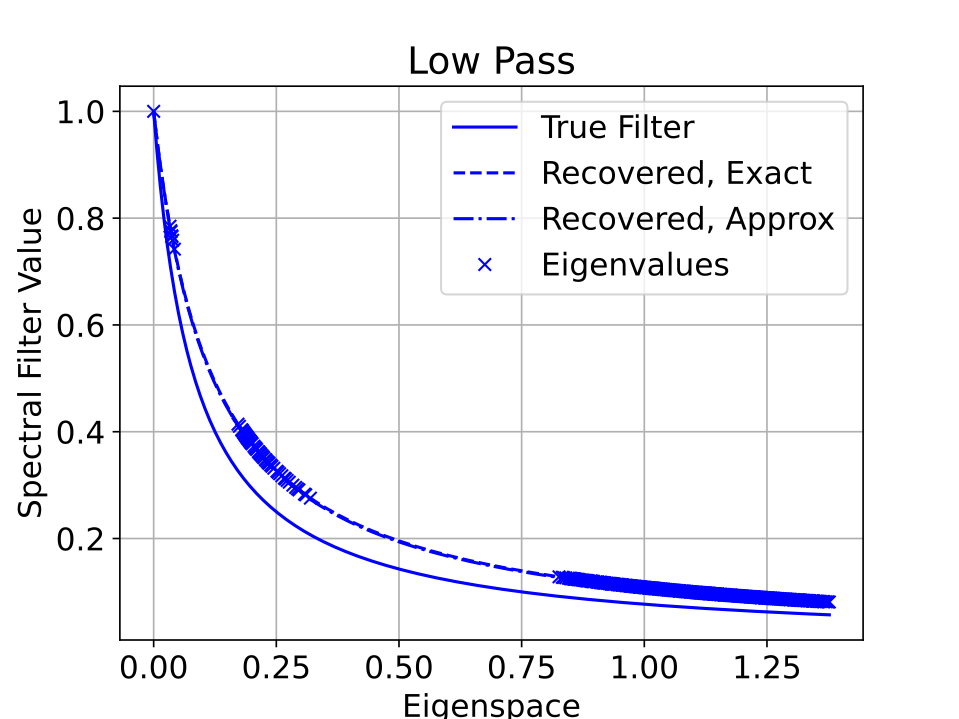}
\end{subfigure}
\begin{subfigure}[b]{0.24\linewidth}
\includegraphics[width=\linewidth, height = 3.2cm]{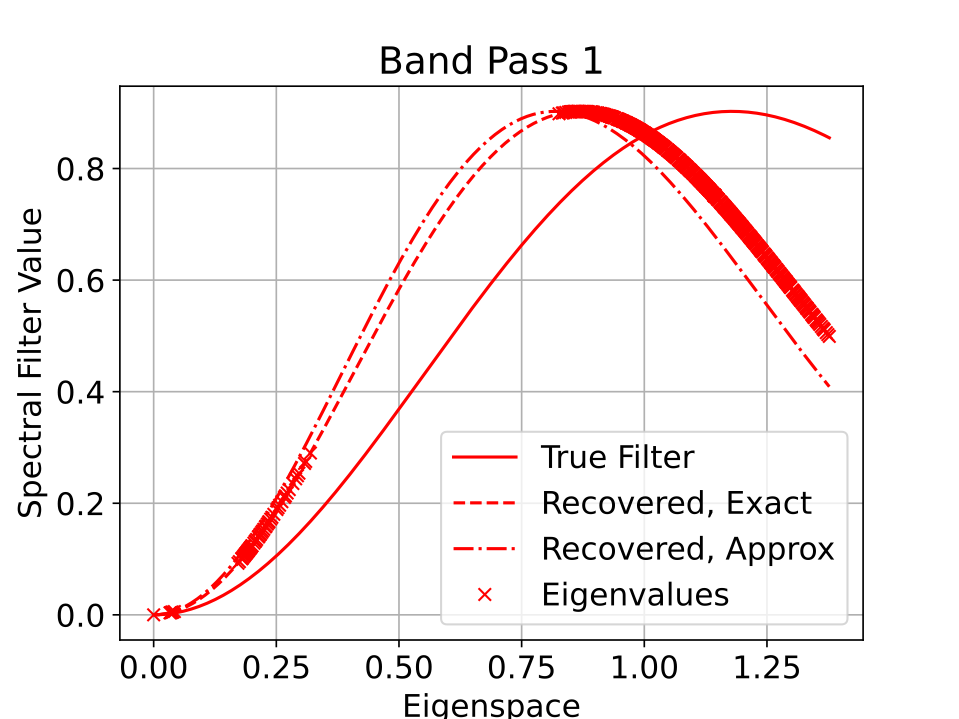}
\end{subfigure}
\begin{subfigure}[b]{0.24\linewidth}
\includegraphics[width=\linewidth, height = 3.2cm]{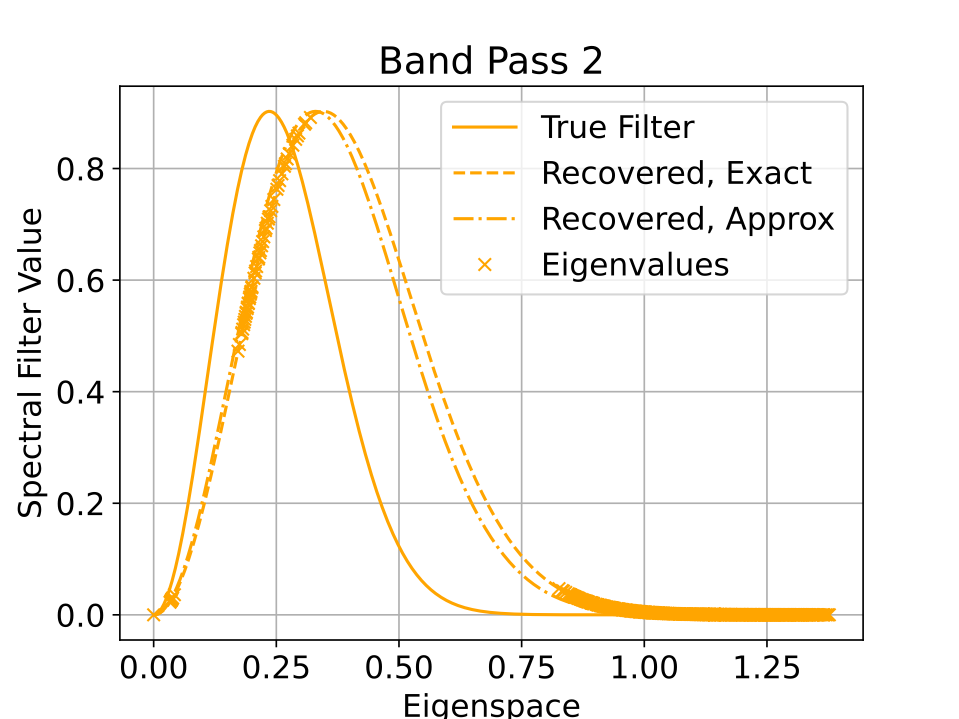}
\end{subfigure}
\begin{subfigure}[b]{0.24\linewidth}
\includegraphics[width=\linewidth, height = 3.2cm]{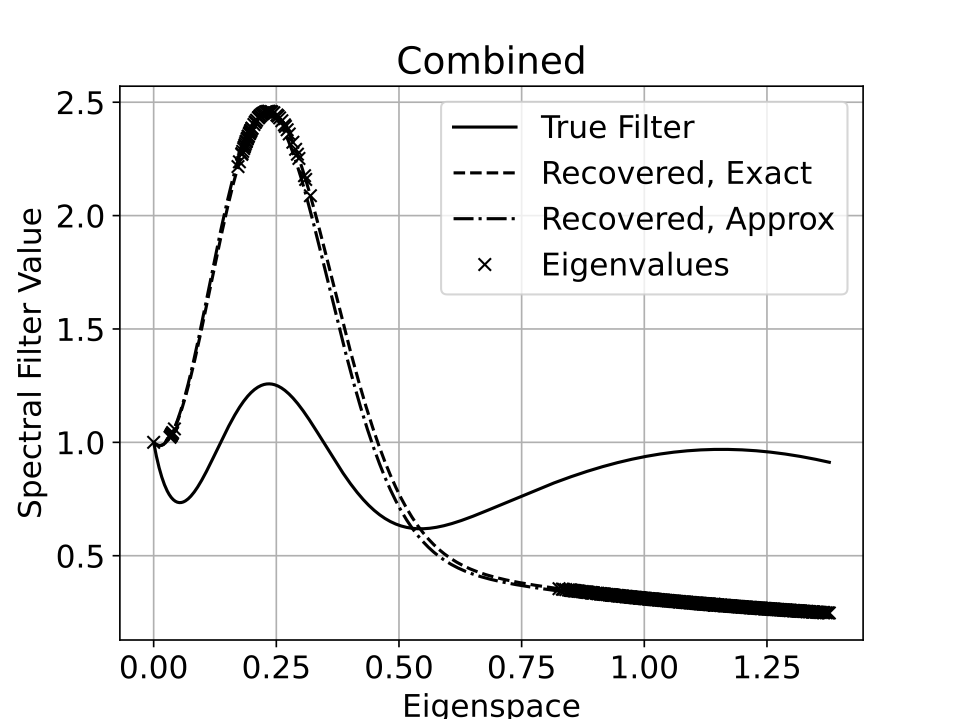}
\end{subfigure}
\begin{subfigure}[b]{0.24\linewidth}
\includegraphics[width=\linewidth, height = 3.2cm]{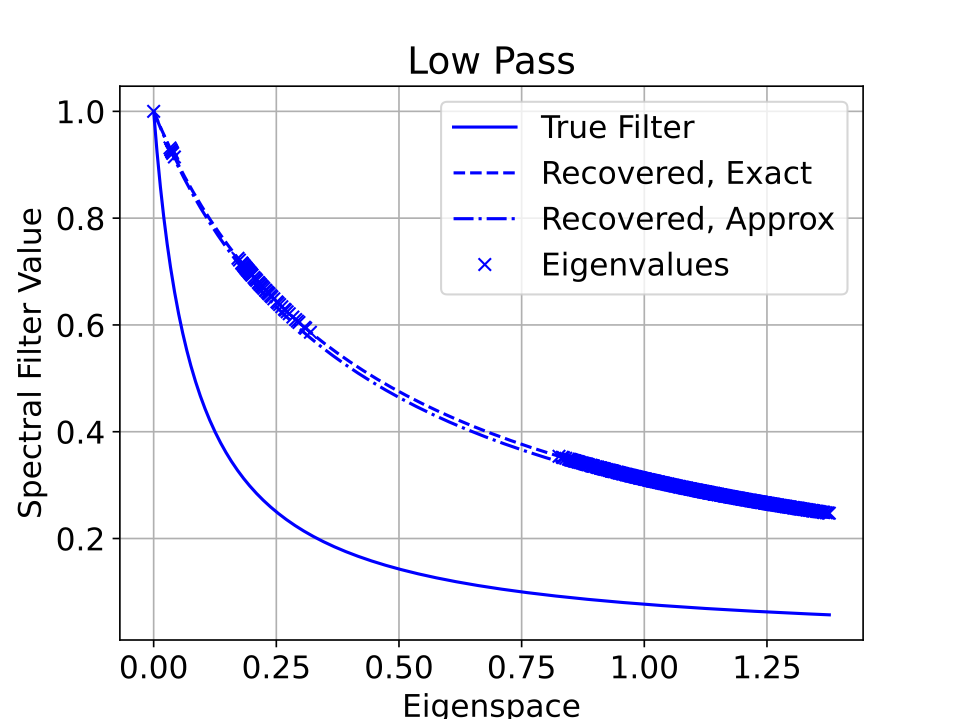}
\end{subfigure}
\begin{subfigure}[b]{0.24\linewidth}
\includegraphics[width=\linewidth, height = 3.2cm]{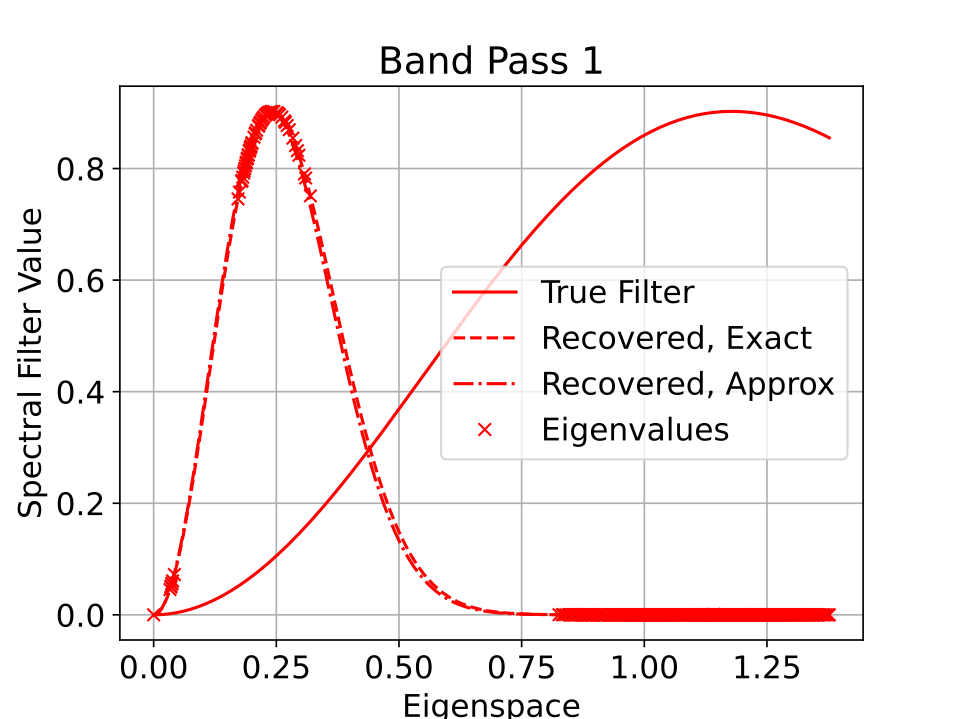}
\end{subfigure}
\begin{subfigure}[b]{0.24\linewidth}
\includegraphics[width=\linewidth, height = 3.2cm]{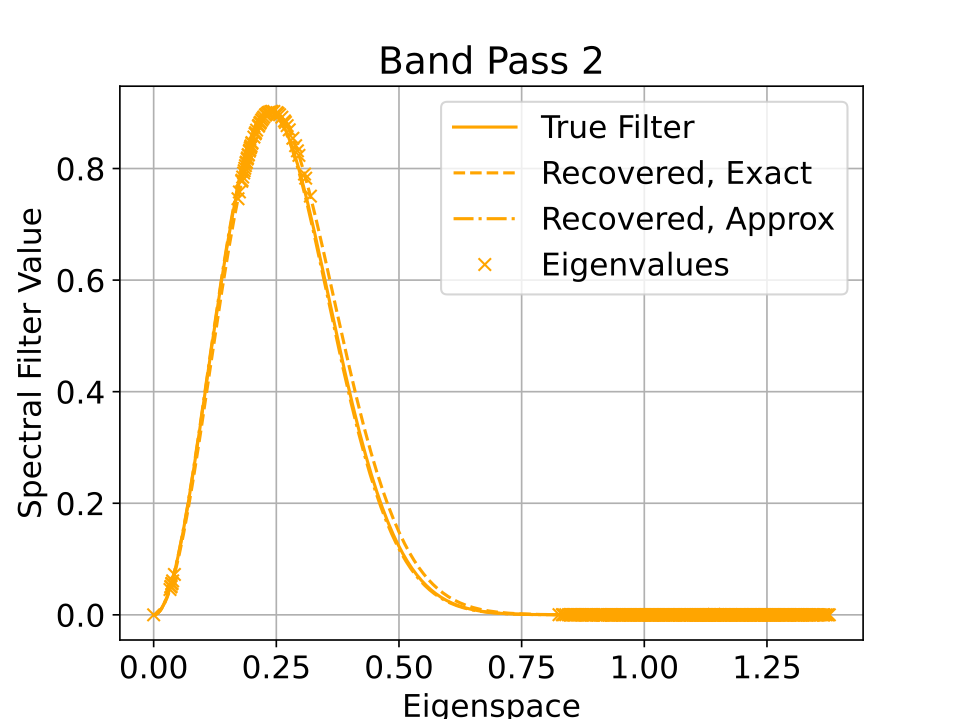}
\end{subfigure}
\begin{subfigure}[b]{0.24\linewidth}
\includegraphics[width=\linewidth, height = 3.2cm]{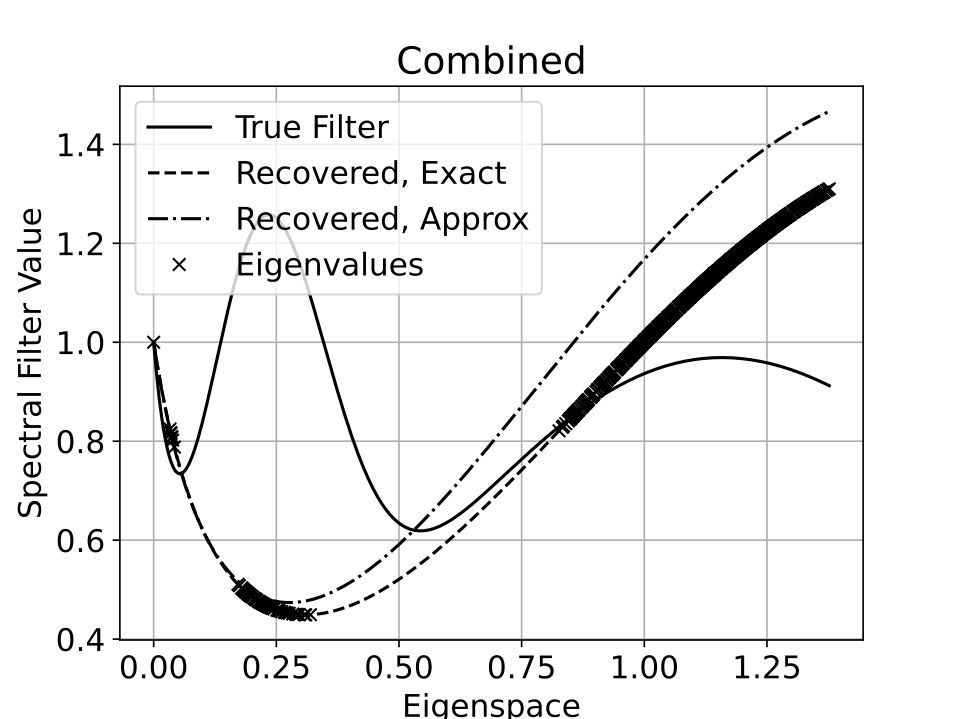}
\end{subfigure}
\begin{subfigure}[b]{0.24\linewidth}
\includegraphics[width=\linewidth, height = 3.2cm]{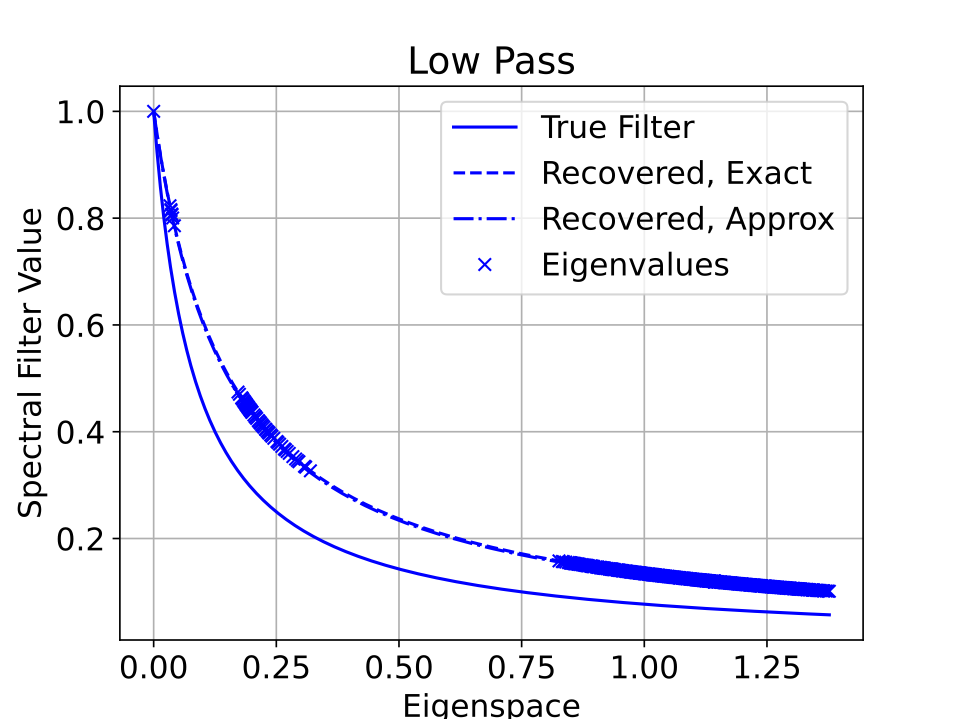}
\end{subfigure}
\begin{subfigure}[b]{0.24\linewidth}
\includegraphics[width=\linewidth, height = 3.2cm]{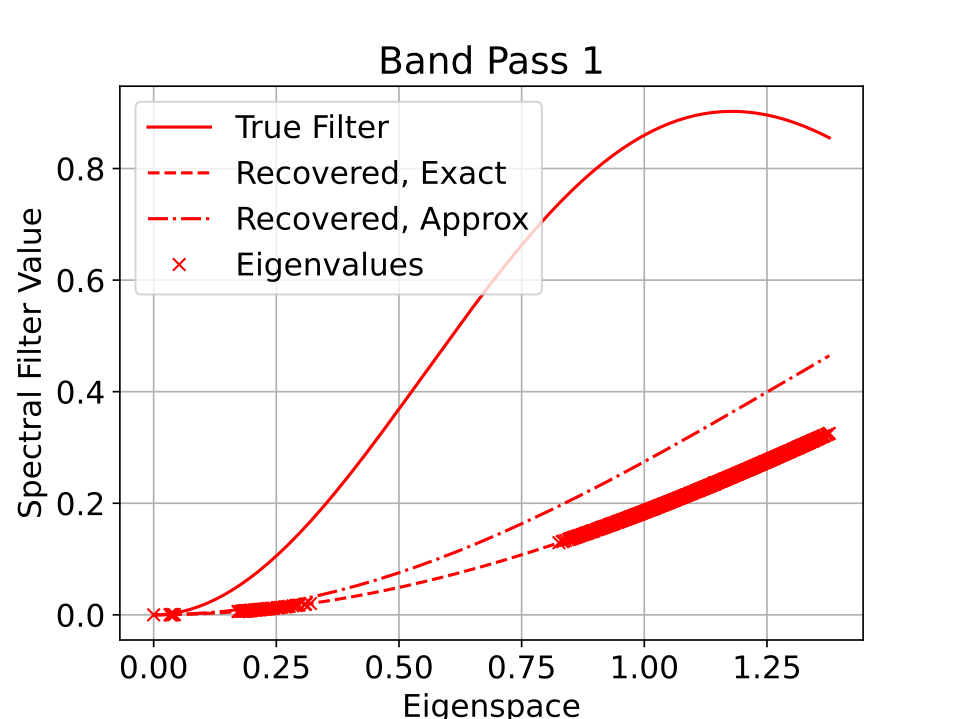}
\end{subfigure}
\begin{subfigure}[b]{0.24\linewidth}
\includegraphics[width=\linewidth, height = 3.2cm]{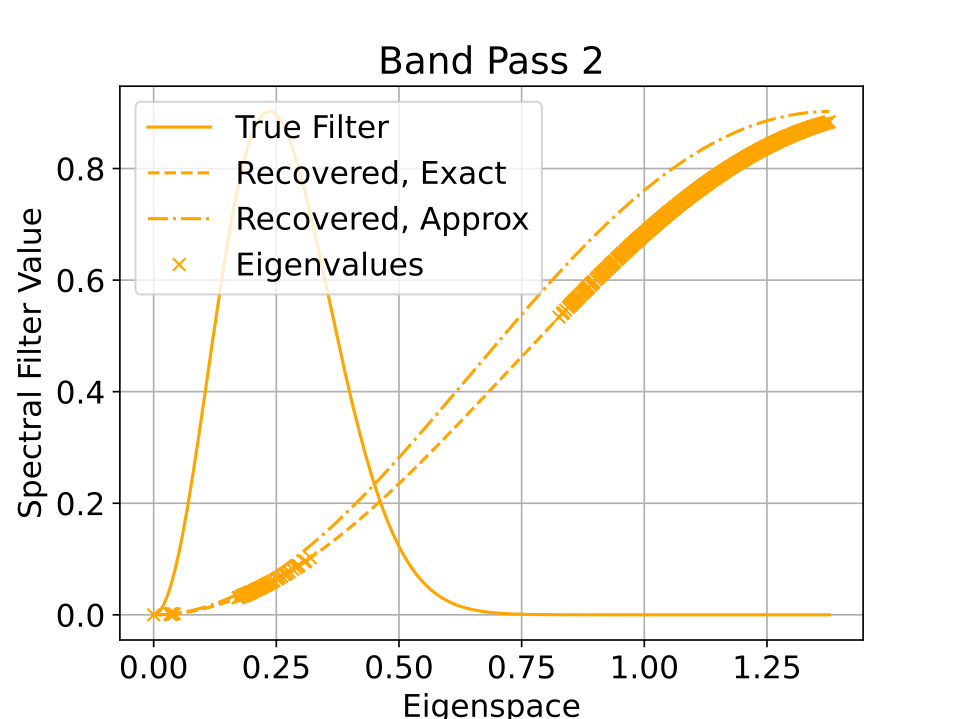}
\end{subfigure}
\caption{Scale recoveries from synthetic experiments using 10\% of nodes as training. Each row is a different random selections of training nodes.}\label{wavelet_transforms_add10}
\end{figure}

\begin{figure}[h]
\centering
\begin{subfigure}[b]{0.24\linewidth}
\includegraphics[width=\linewidth, height = 3.cm]{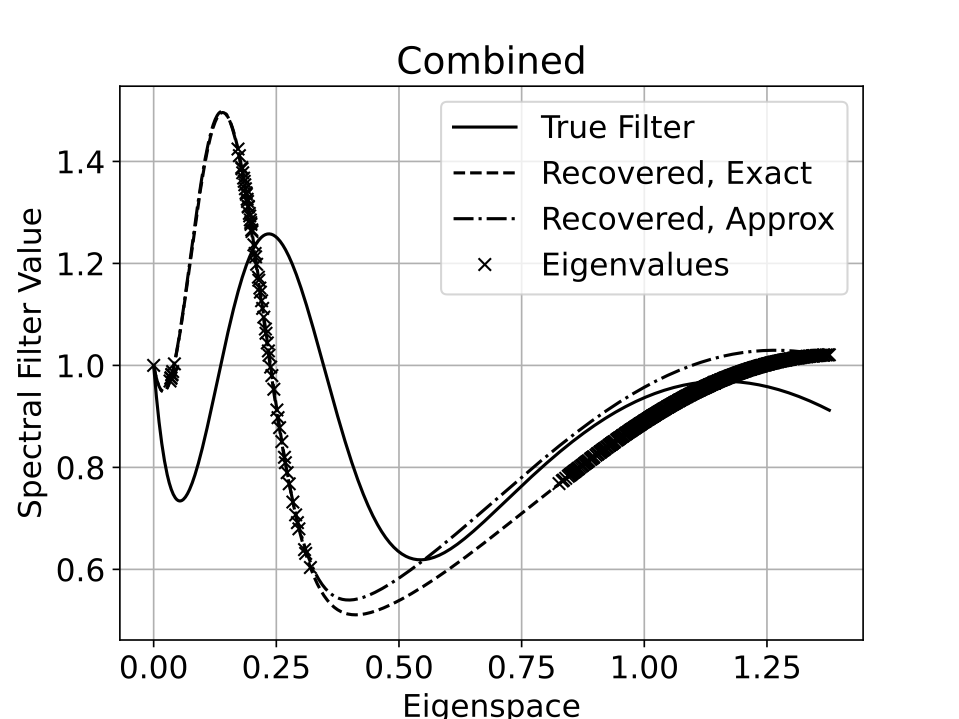}
\end{subfigure}
\begin{subfigure}[b]{0.24\linewidth}
\includegraphics[width=\linewidth, height = 3.cm]{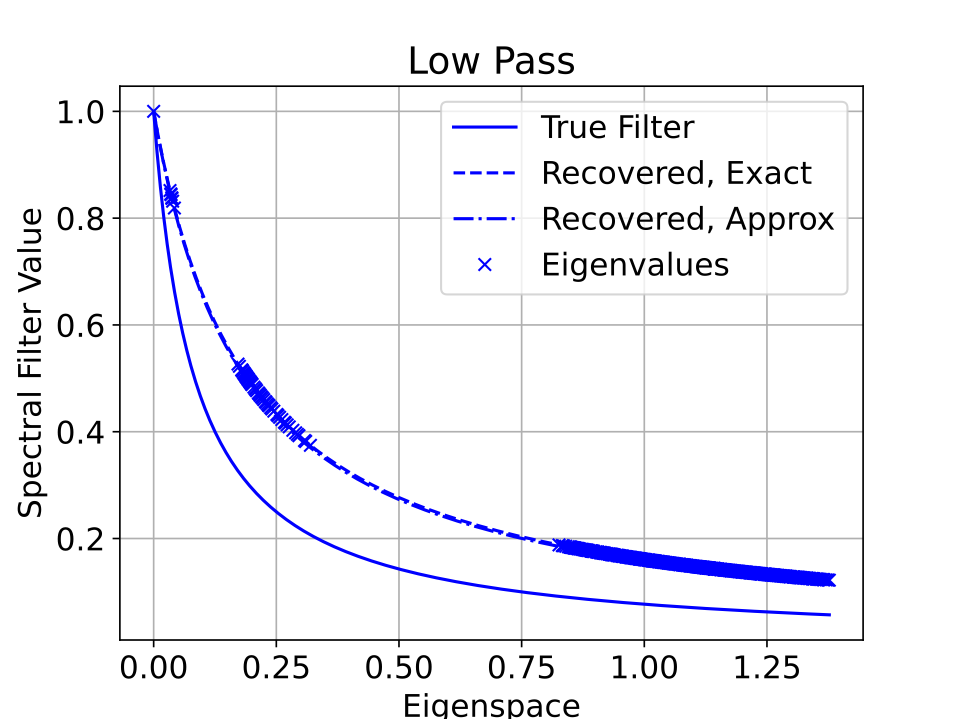}
\end{subfigure}
\begin{subfigure}[b]{0.24\linewidth}
\includegraphics[width=\linewidth, height = 3.cm]{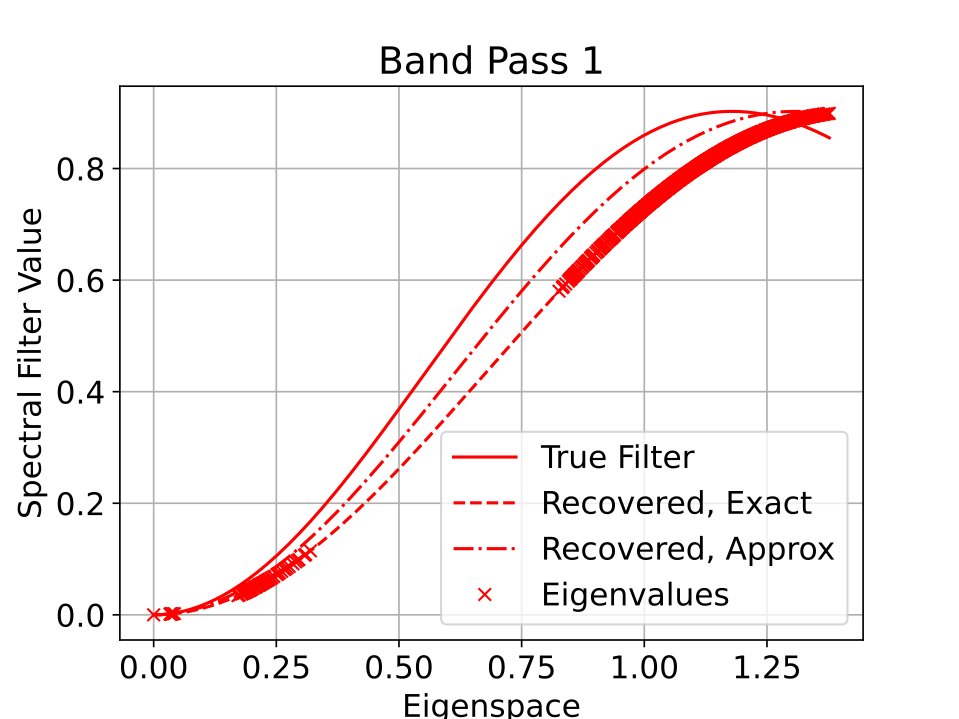}
\end{subfigure}
\begin{subfigure}[b]{0.24\linewidth}
\includegraphics[width=\linewidth, height = 3.cm]{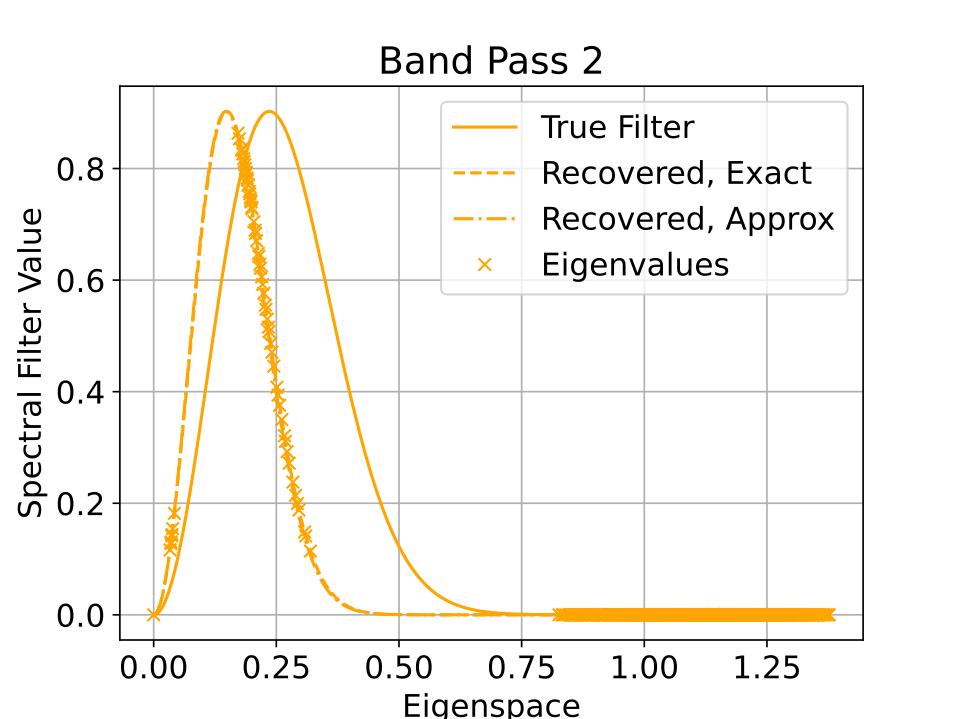}
\end{subfigure}
\begin{subfigure}[b]{0.24\linewidth}
\includegraphics[width=\linewidth, height = 3.cm]{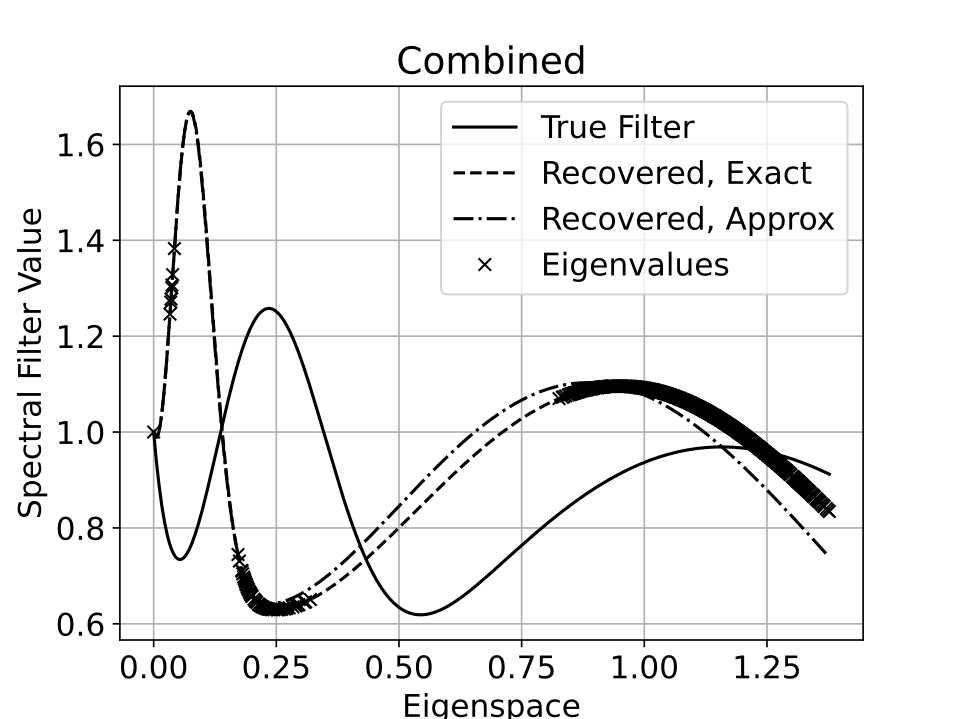}
\end{subfigure}
\begin{subfigure}[b]{0.24\linewidth}
\includegraphics[width=\linewidth, height = 3.cm]{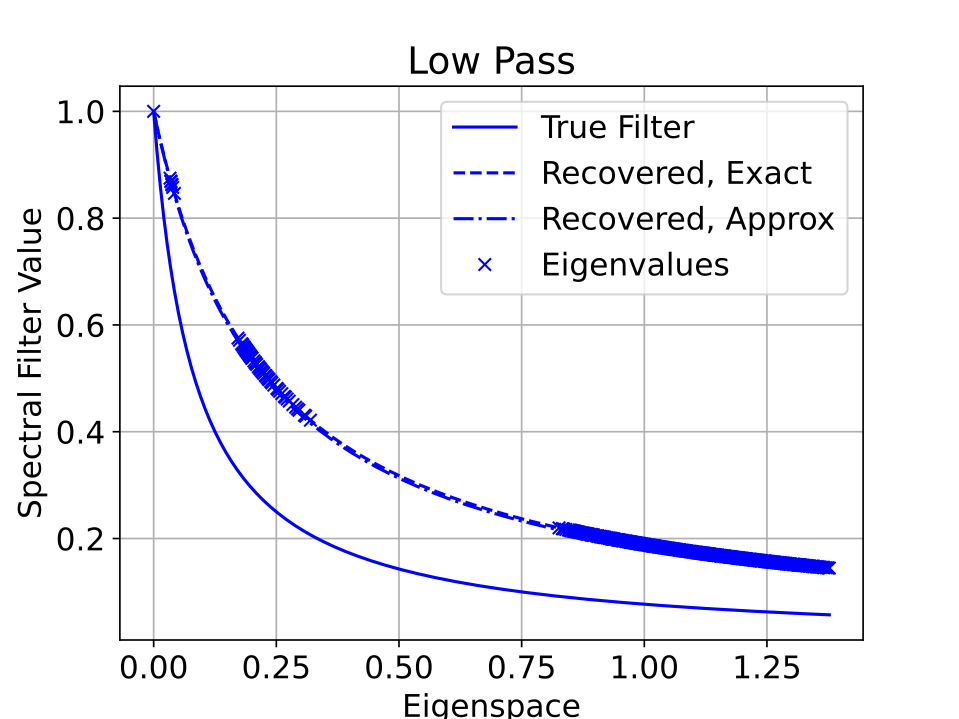}
\end{subfigure}
\begin{subfigure}[b]{0.24\linewidth}
\includegraphics[width=\linewidth, height = 3.cm]{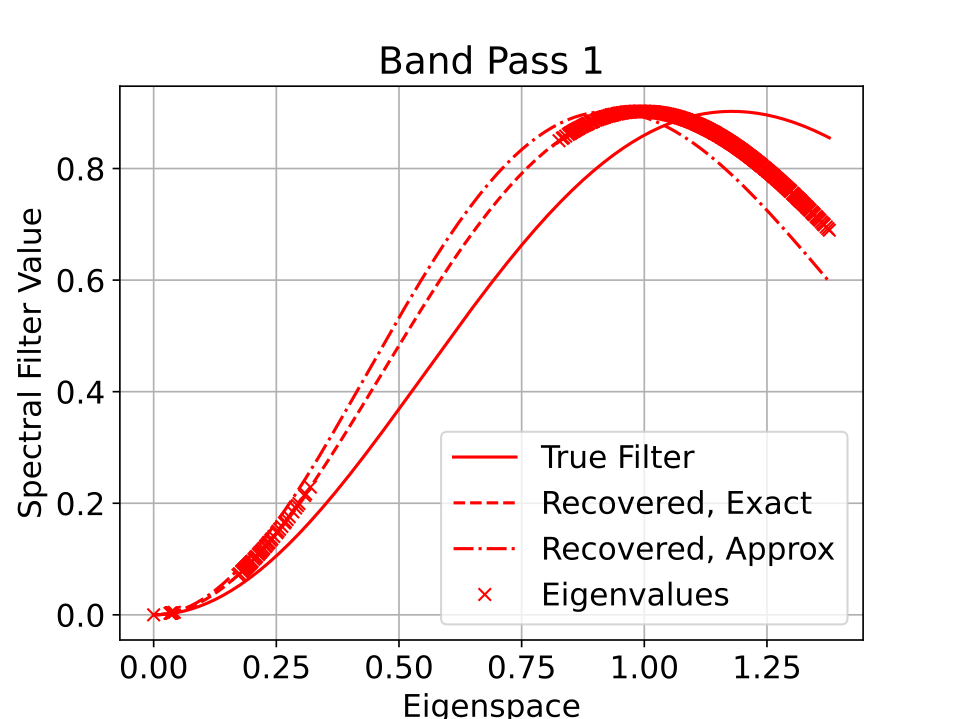}
\end{subfigure}
\begin{subfigure}[b]{0.24\linewidth}
\includegraphics[width=\linewidth, height = 3.cm]{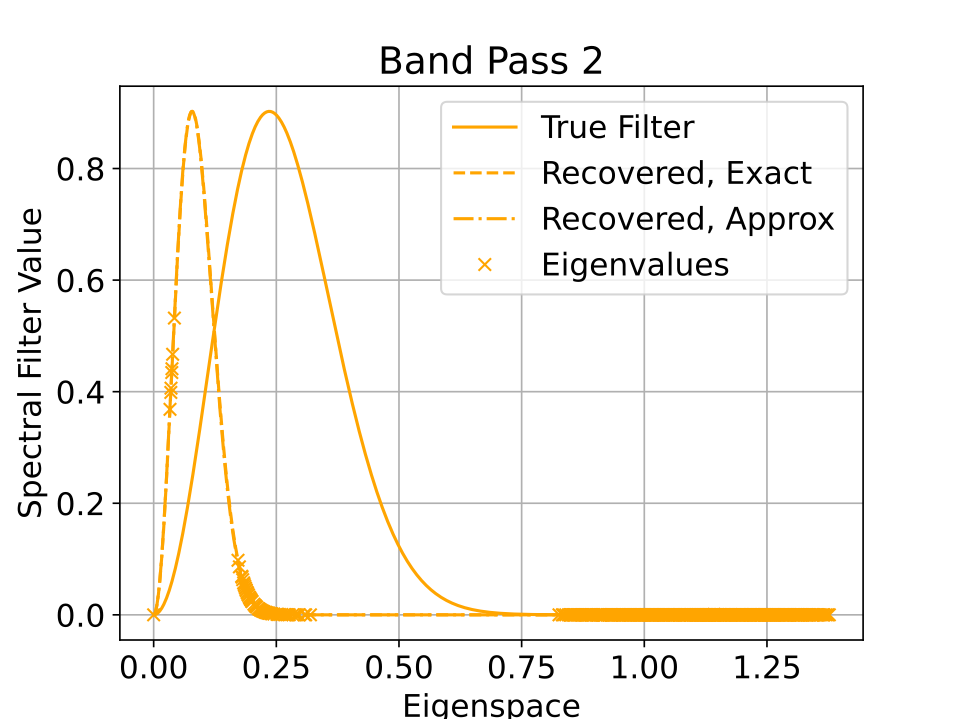}
\end{subfigure}
\begin{subfigure}[b]{0.24\linewidth}
\includegraphics[width=\linewidth, height = 3.cm]{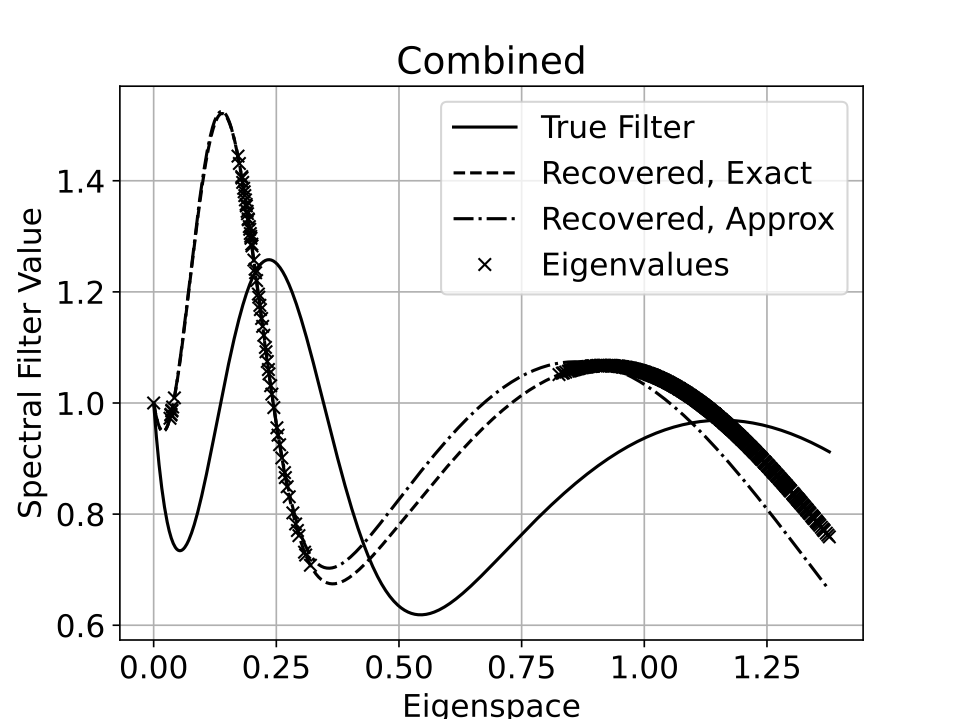}
\end{subfigure}
\begin{subfigure}[b]{0.24\linewidth}
\includegraphics[width=\linewidth, height = 3.cm]{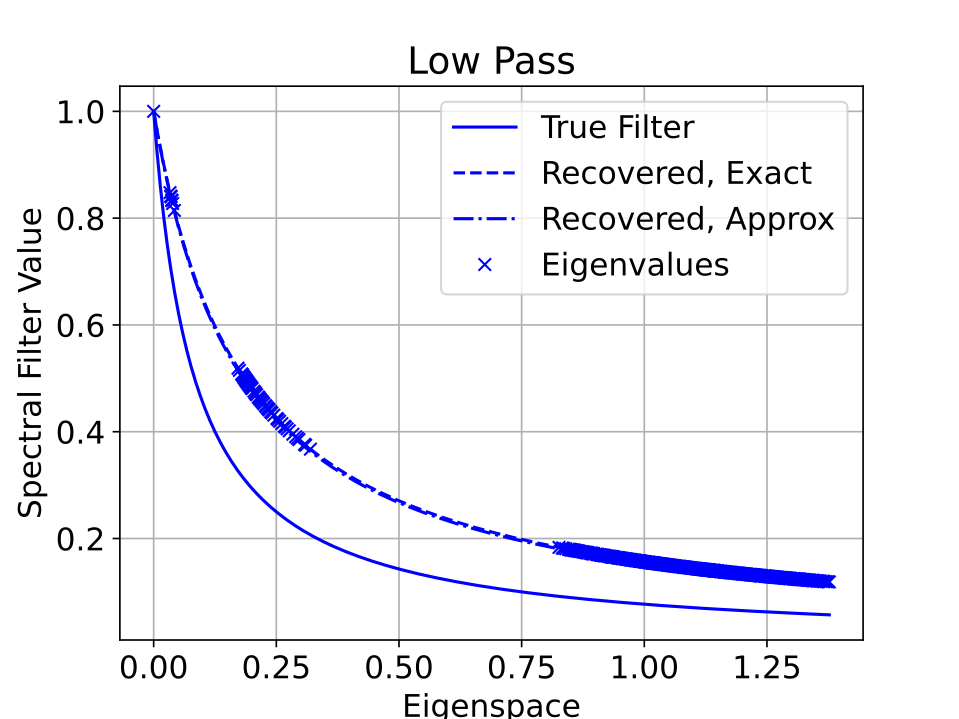}
\end{subfigure}
\begin{subfigure}[b]{0.24\linewidth}
\includegraphics[width=\linewidth, height = 3.cm]{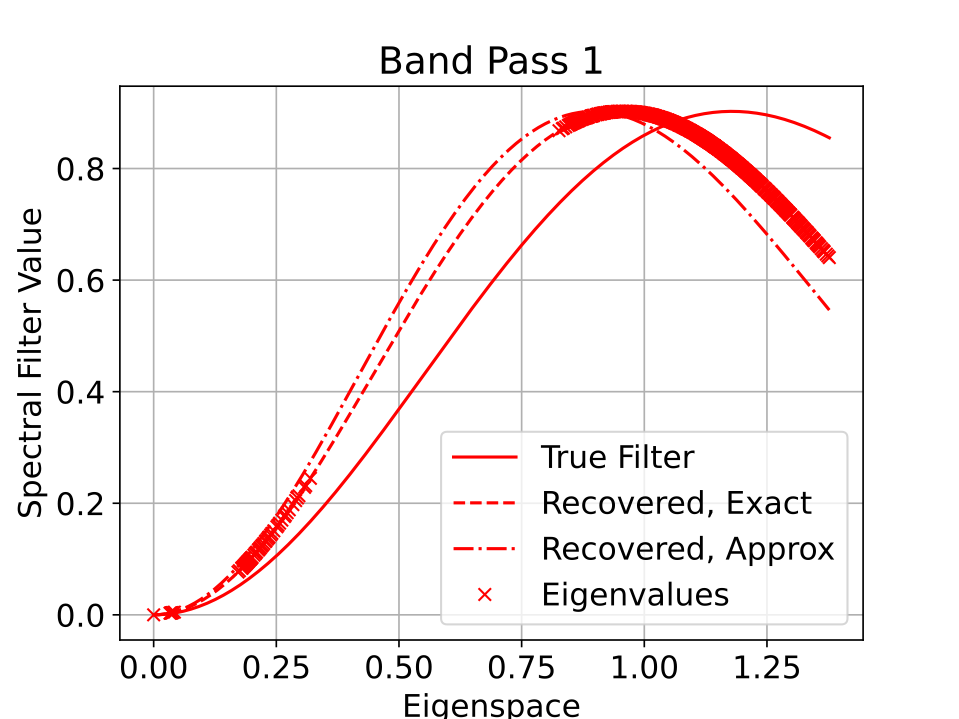}
\end{subfigure}
\begin{subfigure}[b]{0.24\linewidth}
\includegraphics[width=\linewidth, height = 3.cm]{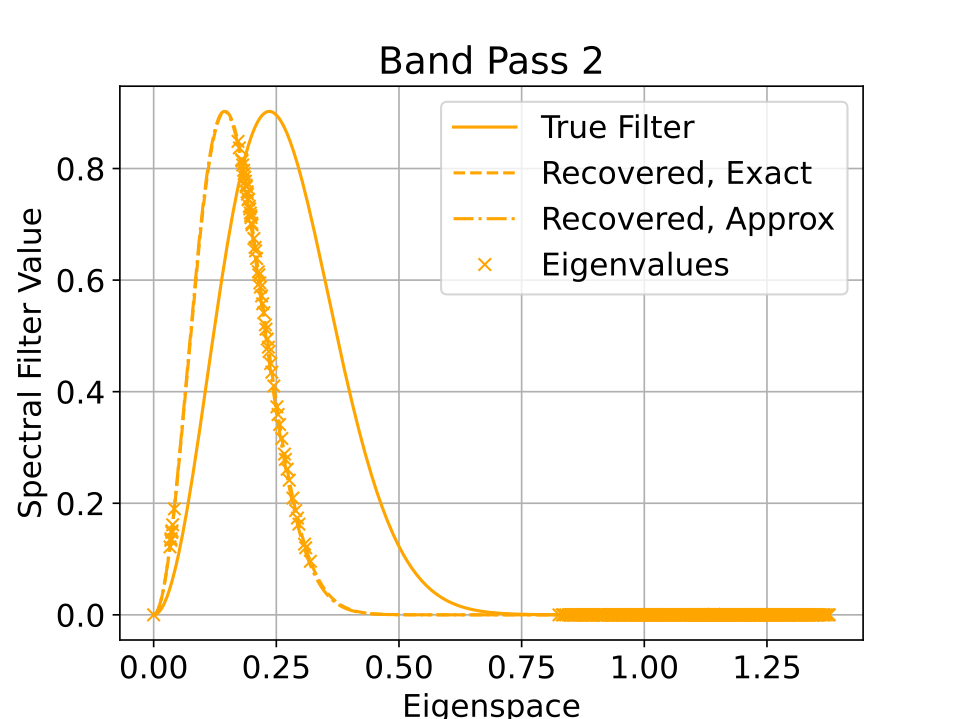}
\end{subfigure}
\caption{Scale recoveries from synthetic experiments using 30\% of nodes as training. Each row is a different random selections of training nodes.}\label{wavelet_transforms_add30}
\end{figure}

\begin{figure}[h]
\centering
\begin{subfigure}[b]{0.24\linewidth}
\includegraphics[width=\linewidth, height = 3.cm]{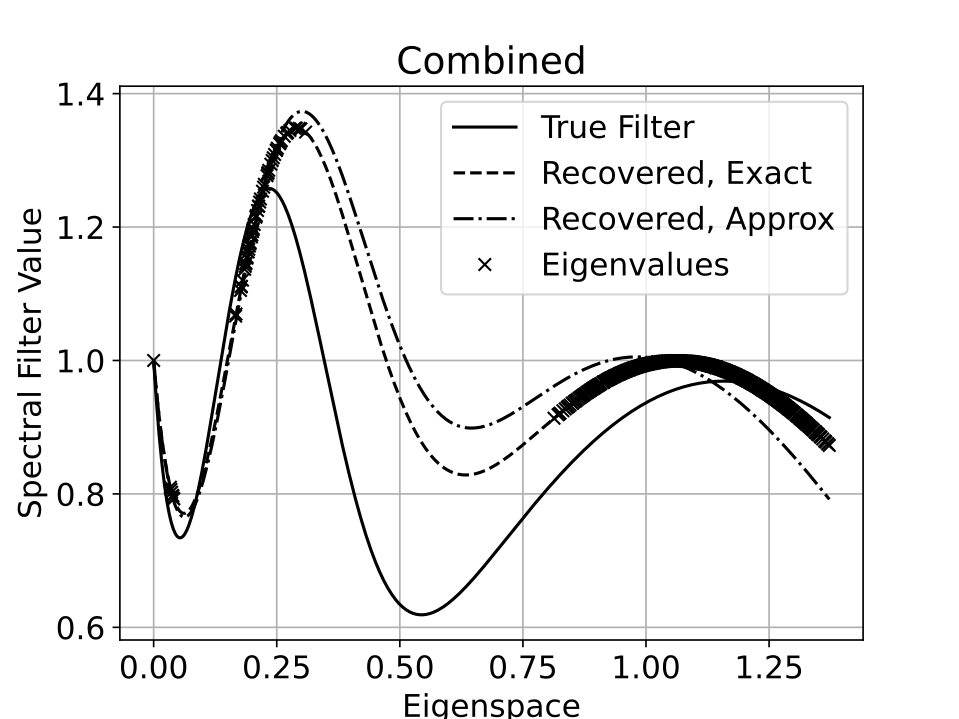}
\end{subfigure}
\begin{subfigure}[b]{0.24\linewidth}
\includegraphics[width=\linewidth, height = 3.cm]{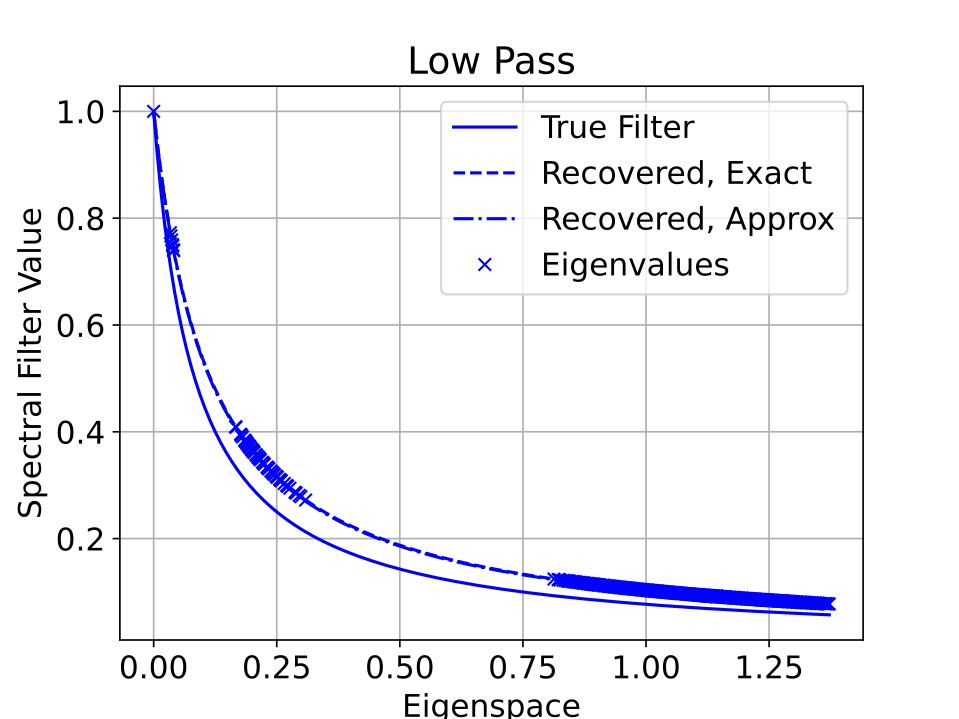}
\end{subfigure}
\begin{subfigure}[b]{0.24\linewidth}
\includegraphics[width=\linewidth, height = 3.cm]{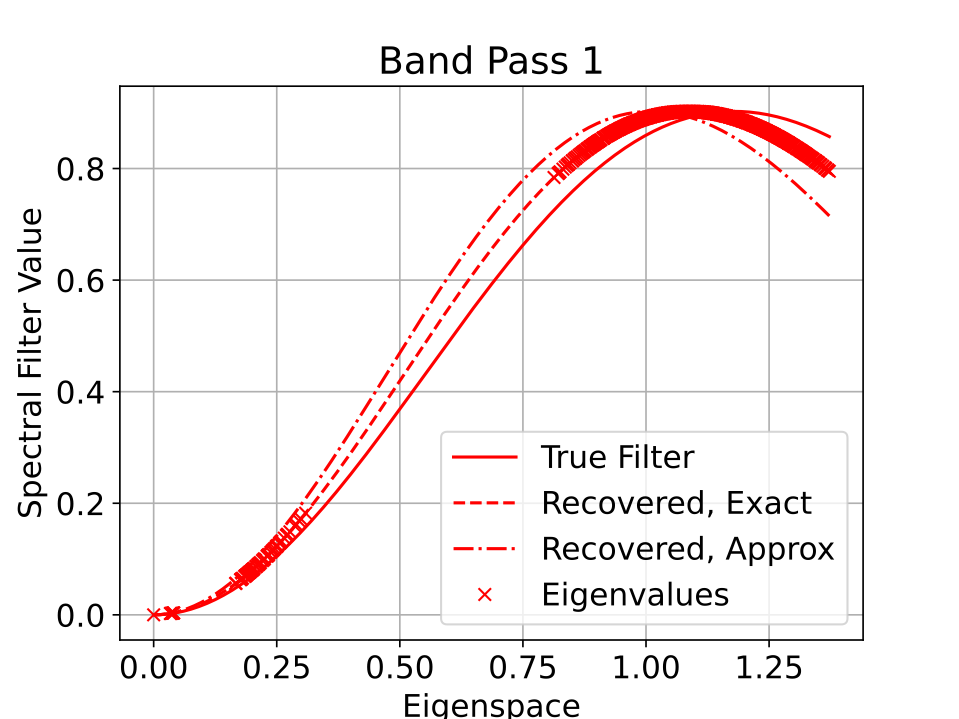}
\end{subfigure}
\begin{subfigure}[b]{0.24\linewidth}
\includegraphics[width=\linewidth, height = 3.cm]{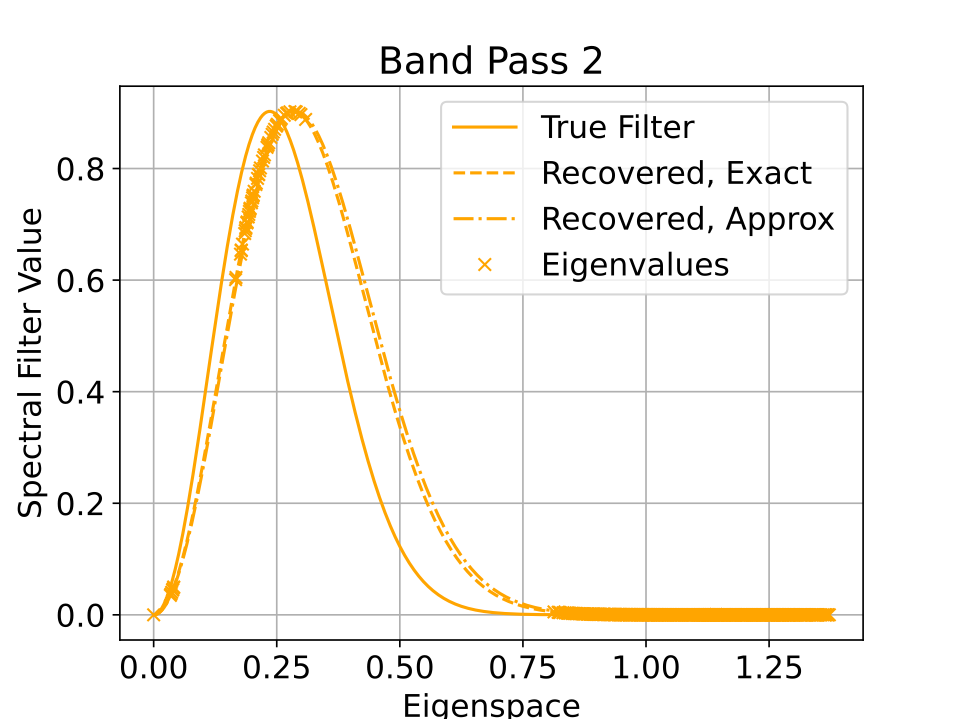}
\end{subfigure}
\begin{subfigure}[b]{0.24\linewidth}
\includegraphics[width=\linewidth, height = 3.cm]{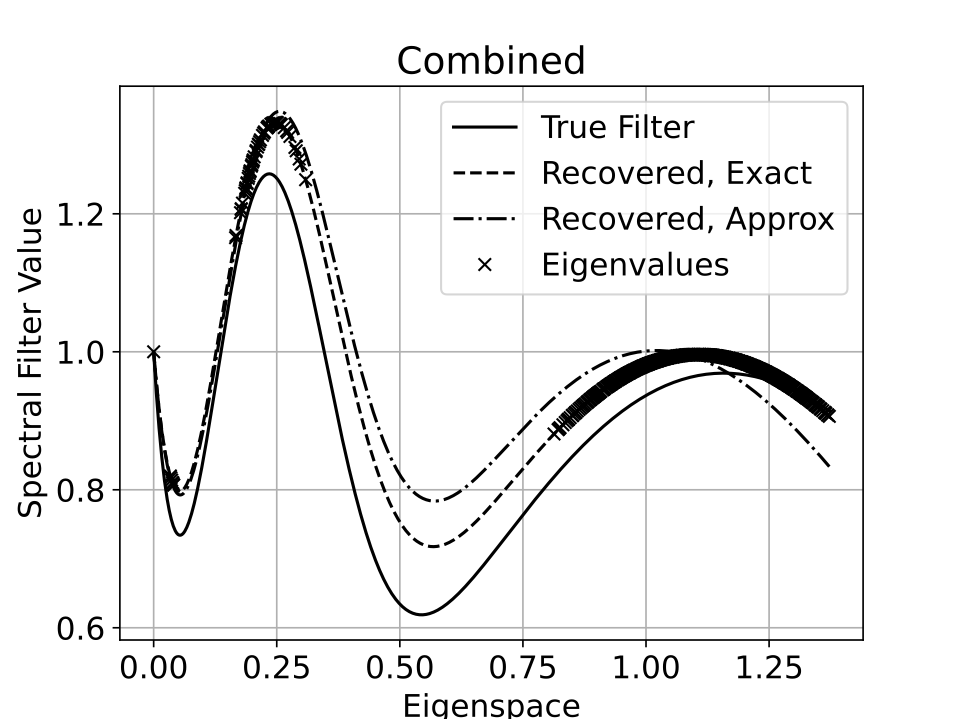}
\end{subfigure}
\begin{subfigure}[b]{0.24\linewidth}
\includegraphics[width=\linewidth, height = 3.cm]{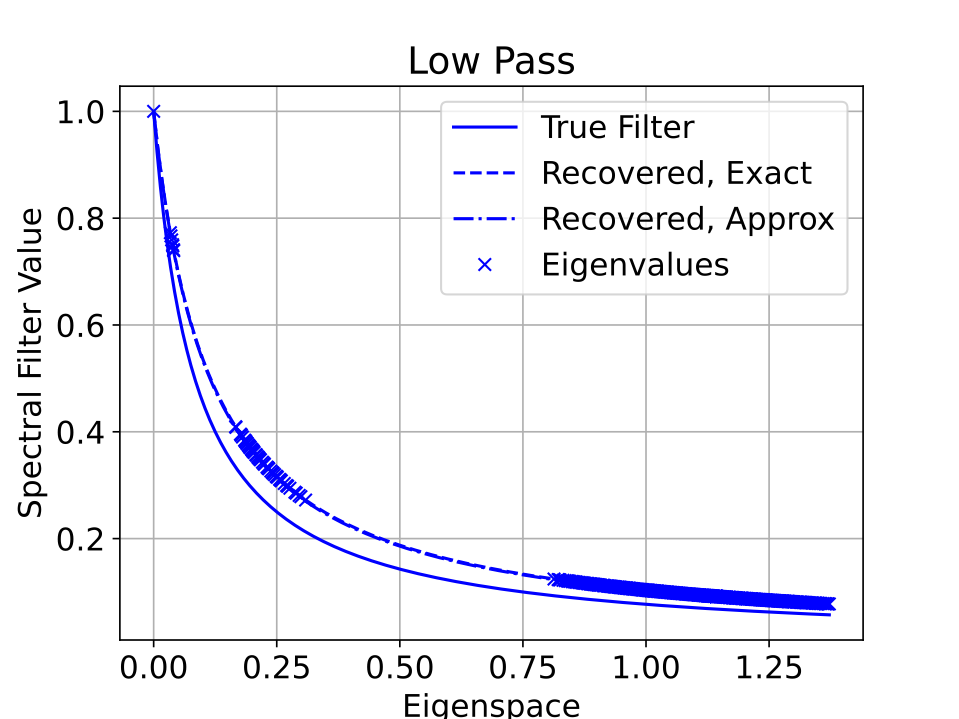}
\end{subfigure}
\begin{subfigure}[b]{0.24\linewidth}
\includegraphics[width=\linewidth, height = 3.cm]{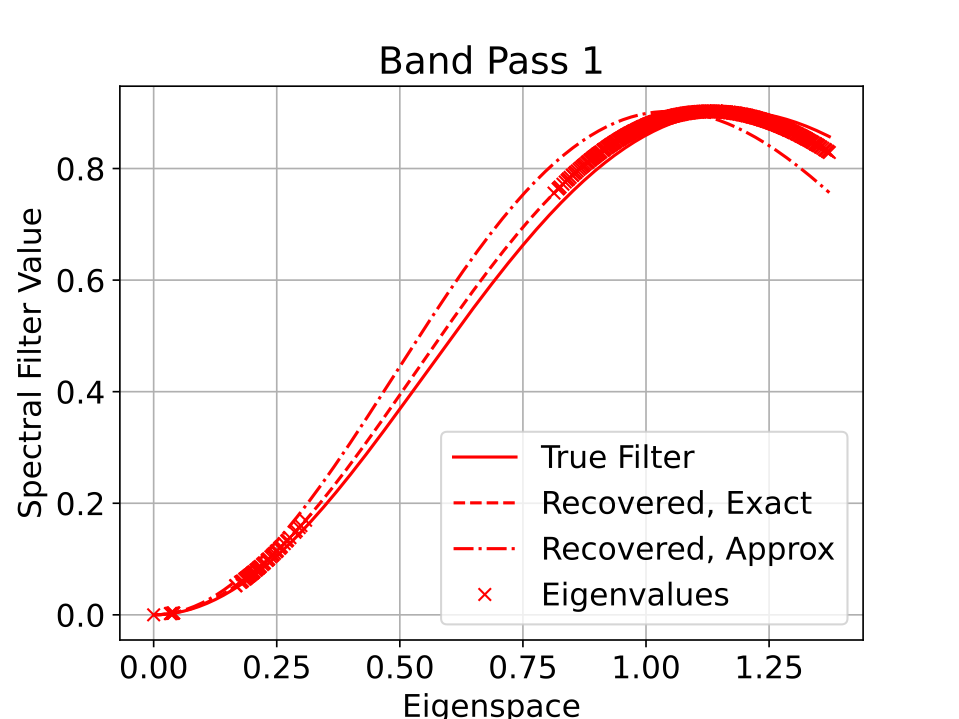}
\end{subfigure}
\begin{subfigure}[b]{0.24\linewidth}
\includegraphics[width=\linewidth, height = 3.cm]{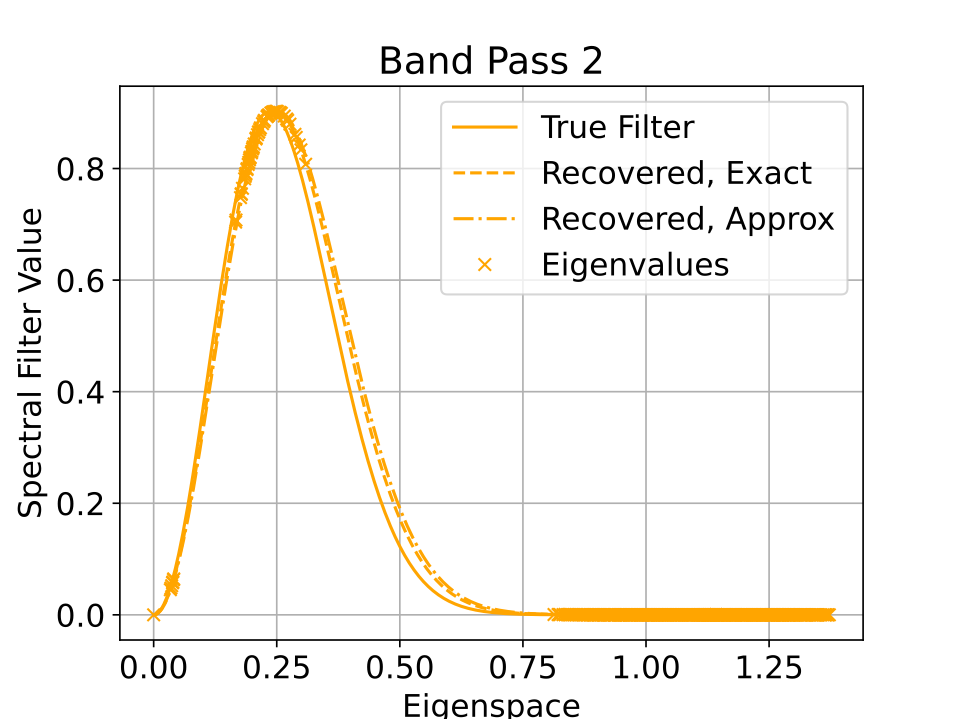}
\end{subfigure}
\begin{subfigure}[b]{0.24\linewidth}
\includegraphics[width=\linewidth, height = 3.cm]{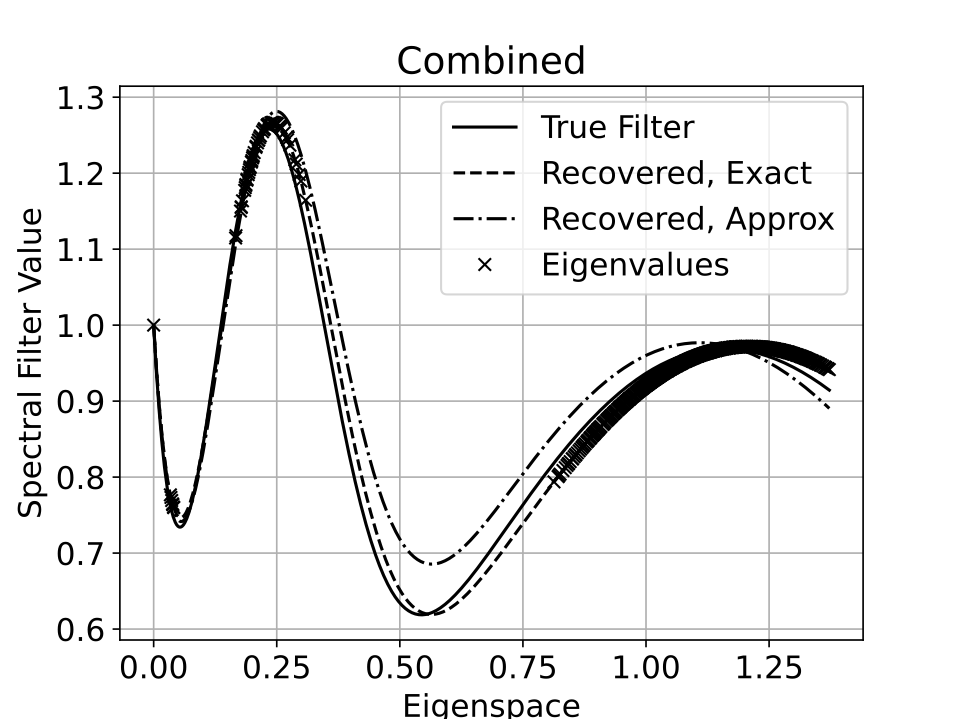}
\end{subfigure}
\begin{subfigure}[b]{0.24\linewidth}
\includegraphics[width=\linewidth, height = 3.cm]{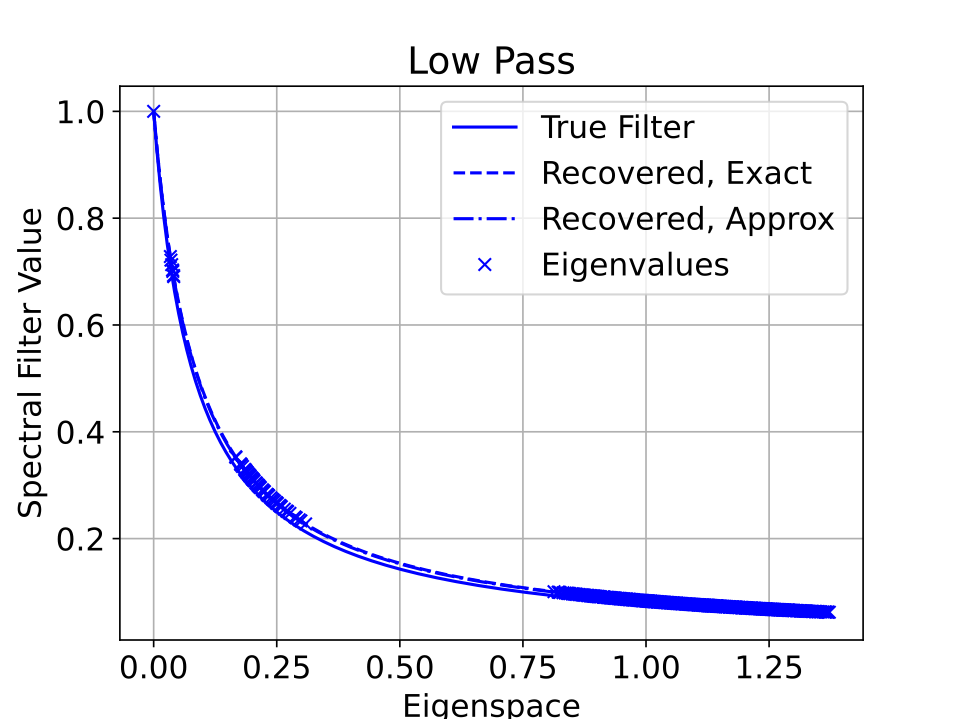}
\end{subfigure}
\begin{subfigure}[b]{0.24\linewidth}
\includegraphics[width=\linewidth, height = 3.cm]{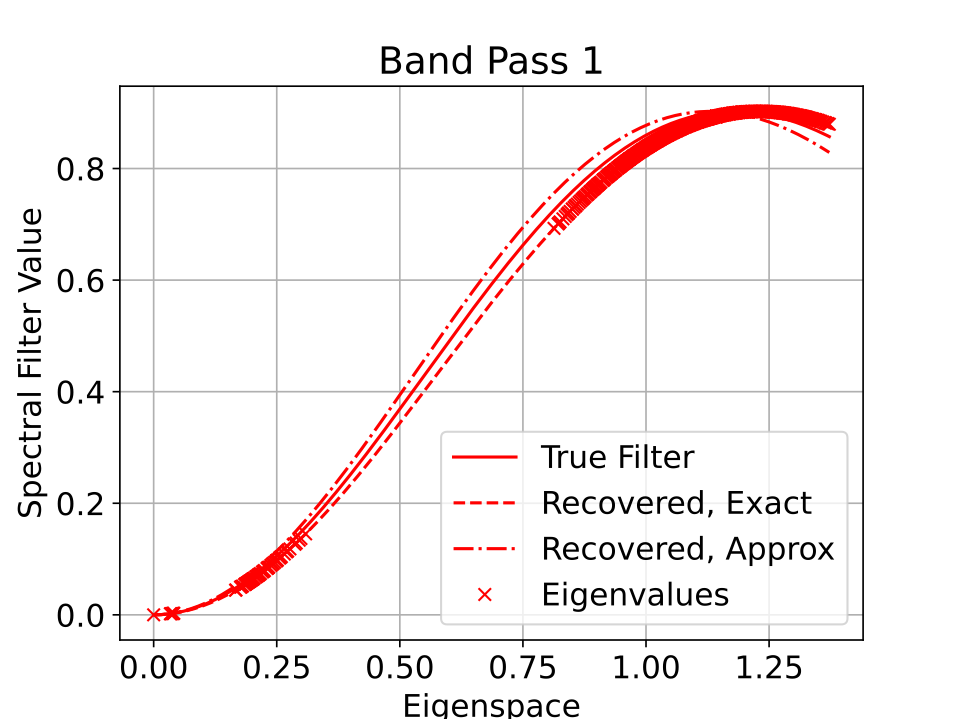}
\end{subfigure}
\begin{subfigure}[b]{0.24\linewidth}
\includegraphics[width=\linewidth, height = 3.cm]{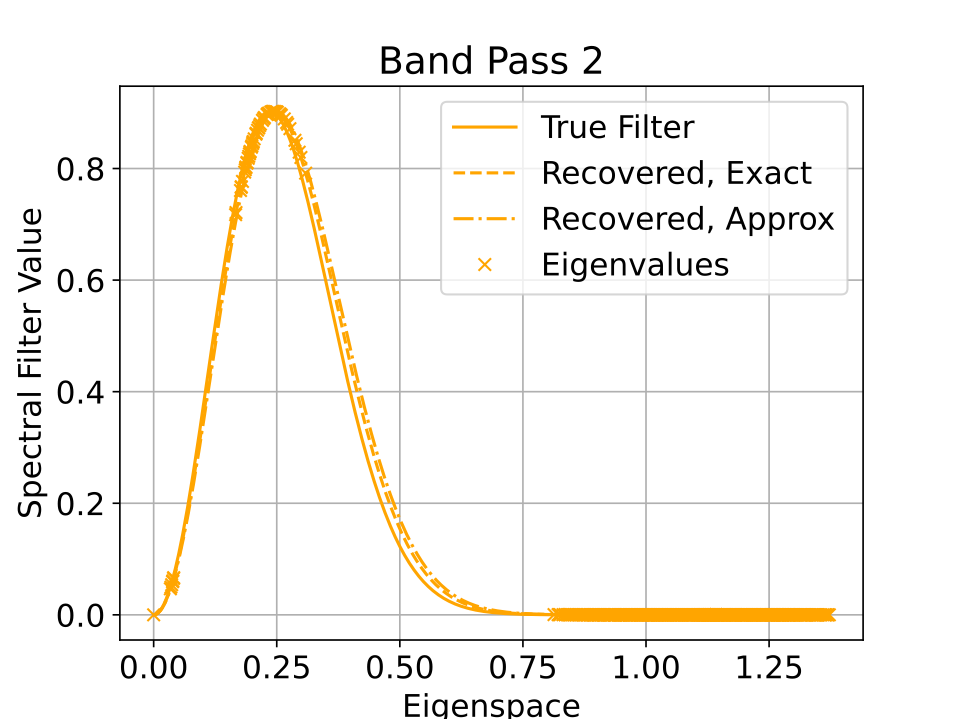}
\end{subfigure}
\caption{Scale recoveries from synthetic experiments using 70\% of nodes as training. Each row is a different random selections of training nodes.} \label{wavelet_transforms_add70}
\end{figure}

The selection of nodes as training will have an effect on the scales we recover. We have presented one particular random split in the main text; here, we present the scale recovery results based on 10\%, 30\% and 70\% of nodes in the graph as training nodes. For each setting, we present results from three random splits, and these are presented in Figure \ref{wavelet_transforms_add10}, \ref{wavelet_transforms_add30} and \ref{wavelet_transforms_add70}. Visually, we can see the quality of the recoveries improves as the number of training nodes we use increases. Additionally, the approximate recoveries are consistently very close to the exact recoveries, showing the accuracy of our approximations.

\subsection{Baseline GP Models on Synthetic Data}

We also test the GP models that we compared against as baselines on synthetic data. The graph neural network models were not compared against as they require a validation set of nodes which are not assigned and would make an unfair comparison. We use GGP and ChebGP to make predictions on the synthetically generated signals, with the MAEs presented in Figure \ref{maes_ggp_cheby}.

The results in Figure \ref{maes_ggp_cheby} showed that the GGP only improves marginally with additional training data, indicating the model's inability to capture multi-scale information. ChebGP, which uses Chebyshev polynomials for approximations, does approximate a multi-scale spectral wavelet function, but we can see by the means and quantiles of the boxplots that they are less consistent in producing low MAEs as our polynomial approximation. As the number of training nodes increases, the model should be able to capture the different scales more accurately; however, the wider quantiles indicates Chebyshev approximations are less consistent in producing accurate recoveries.

\subsection{Performance on Synthetic Data Generated Using Different Ground Truth Wavelets}

The synthetic setting uses the Mexican Hat kernel in both the GP and the generating model. We now test when the data generating model is different from the GP, where the data is generated from the same procedure but with a Morlet wavelet. 

\subsection{WGGP without Feature Space Kernel}

To measure the importance of the feature space kernel, we repeat experiments with WGGP on Cora and Citeseer with the feature space kernel $K_\Psi(\bd{x}_i, \bd{x}_j) = \delta_{ij}$ set to the identity. As a result, the model classifies nodes no longer also based on node features but on graph structure alone. We compare the results to those of the full WGGP model in Table~\ref{tab:no_feat_kernel}.
As expected, the performance of the model drops decisively when removing the dependence on the node features, demonstrating the importance of the node feature kernel for the predictive performance.

\begin{table}[H]
    \centering
    \begin{tabular}{lcccccc}
        \toprule
        \textbf{Method} & \textbf{Cora} & \textbf{Citeseer}\\
        \midrule
        \textbf{WGGP} & $84.7$ & $70.8$ \\
        \textbf{WGGP without node features} & $71.9$ & $47.7$ \\
        \bottomrule
        \vspace{0.5mm}
    \end{tabular}
    \caption{Classification accuracy of the WGGP model with and without the node feature kernel. When removing the node feature kernel, the predictive performance drops by more than $10\%$ for both data sets.}\label{tab:no_feat_kernel}
\end{table}

\subsection{Robustness Analysis}

We perform a robustness analysis examining how the model performance changes as we vary different parts of the model or training setup, while keeping everything else as described in Section 5. 

\paragraph{Data Split} 
In a first experiment, we use $10$ different data splits for Cora and Citeseer that retain the uniform distribution of classes and re-run the model with otherwise equal hyperparameters. The average performance across the $10$ data splits is reported in Table~\ref{tab:robustness} together with one standard deviation. We find that the model perofrmance only varies modestly for both data sets and the performance remains comparable to the one achieved on the public data split.

\paragraph{Number of scales} 
We also analyse how the model performance varies when using different number of scales in the model, ranging from using only a low-pass filter to also including $4$ band pass filters. The results are again reported in Table~\ref{tab:robustness} with the standard deviation over the $4$ different setups ($0$-$4$ scales), showing that the model accuracy varies only slightly when using different number of filters.

\paragraph{Hyperparameter initialisations}
Finally, we repeat the experiments with random initialisations of the scale hyperparameters.
The results with their standard deviation over $10$ different initialisations (Table~\ref{tab:robustness}) demonstrate the model's robustness to different hyperparameter initialisations.

\begin{table}[H]
    \centering
    \begin{tabular}{lcccccc}
        \toprule
        \textbf{Method} & \textbf{Cora} & \textbf{Citeseer}\\
        \midrule
        \textbf{WGGP with varying data splits} & $82.4 \pm 1.1$ & $67.8 \pm 2.7$ \\
        \textbf{WGGP with varying number of scales} & $84.7 \pm 0.2$ & $70.6 \pm 0.2$ \\
        \textbf{WGGP with varying hyperparameter initalisations} & $84.2 \pm 0.4$ & $71.0 \pm 0.6 $ \\
        \bottomrule
        \vspace{0.5mm}
    \end{tabular}
    \caption{Classification accuracy of the WGGP model with and without the node feature kernel. When removing the node feature kernel, the predictive performance drops by more than $10\%$ for both data sets.}\label{tab:robustness}
\end{table}


\section{Statistics of Data Sets}

\begin{table}[H]
    \centering
    \begin{tabular}{lcccccc}
        \toprule
        \textbf{Data} & Type & $\mathbf{N_{nodes}}$ & $\mathbf{N_{edges}}$ & $\mathbf{N_{label\_cat}}$ & $\mathbf{D_{features}}$ & \textbf{Label Rate}\\
        \midrule
        \textbf{Cora} & Citation & 2,708 & 5,429 & 7 &  1,433 &  0.052\\
        \textbf{Citeseer} & Citation & 3,327 & 4,732 & 6 &  3,703 &  0.036\\
        \textbf{PubMed} & Citation & 19,717 & 44,338 & 3 & 500 &  0.003\\
        \bottomrule
        \vspace{0.5mm}
    \end{tabular}
    \caption{Summary of citation networks for node classification experiments.}
\end{table}

\section{Computing platform and code implementation}

The experiments were performed using 
Xeon W-2133 12GB NVIDIA GTX 1080 Ti and 48GB NVIDIA Quadro RTX 8000.

The code for reproducing the results in the paper has been submitted as part of the Supplementary Materials.

\section{Figures}

\begin{figure}[h]
\centering
\begin{subfigure}[b]{0.32\linewidth}
\includegraphics[width=\linewidth, height = 4.cm]{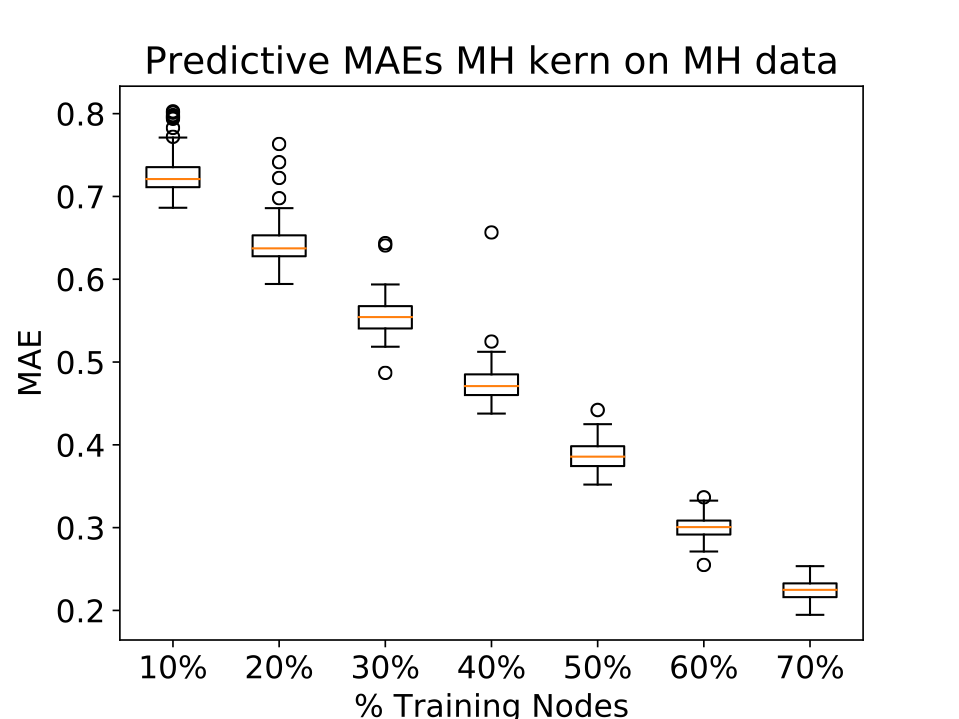}
\caption{}\label{mae2}
\end{subfigure}
\begin{subfigure}[b]{0.32\linewidth}
\includegraphics[width=\linewidth, height = 4.cm]{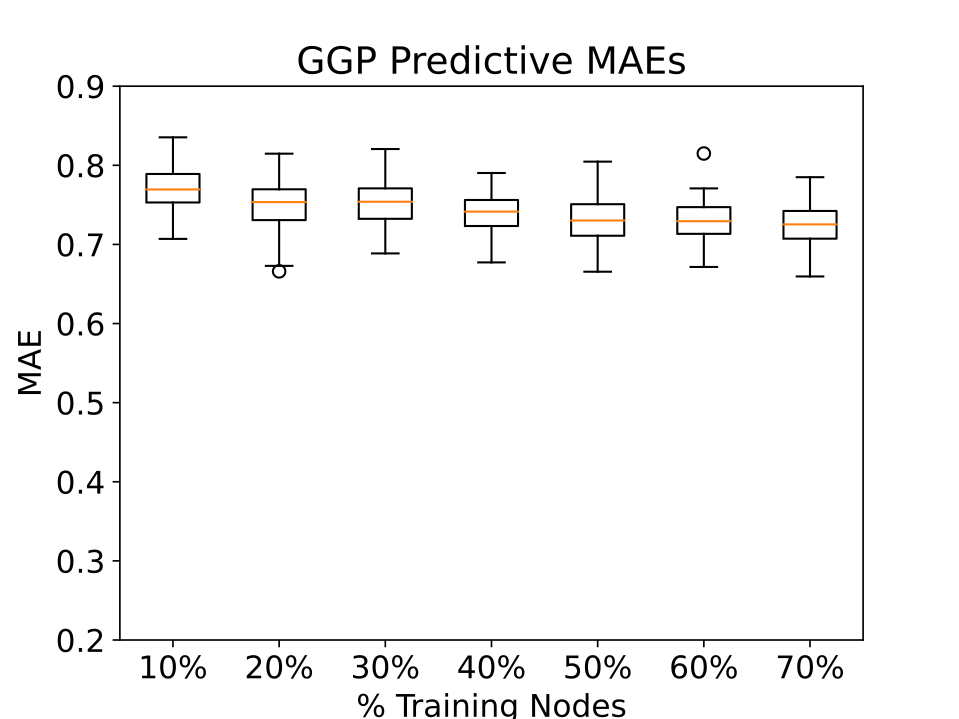}
\caption{}\label{mae1}
\end{subfigure}
\begin{subfigure}[b]{0.32\linewidth}
\includegraphics[width=\linewidth, height = 4.cm]{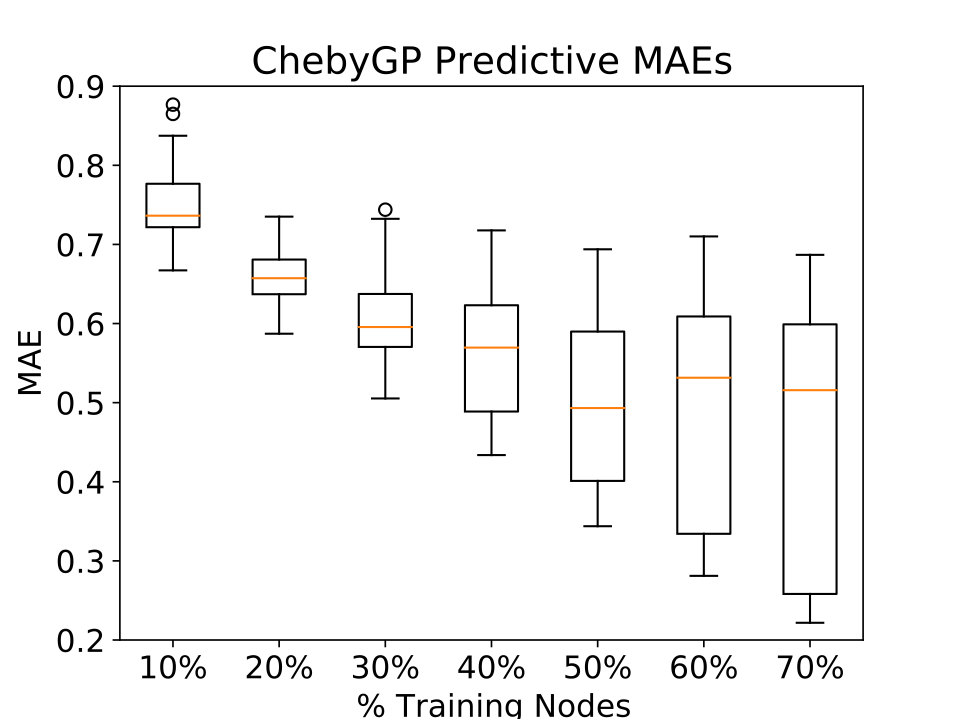}
\caption{}\label{mae2}
\end{subfigure}
\caption{WGGP prediction MAE on synthetic data (Figure 3b from main text) in addition to MAEs of GP models in the baselines, (b) GGP and (c) ChebGP.}\label{maes_ggp_cheby}
\end{figure}

\begin{figure}[h]
\centering
\begin{subfigure}[b]{0.32\linewidth}
\includegraphics[width=\linewidth, height = 4.cm]{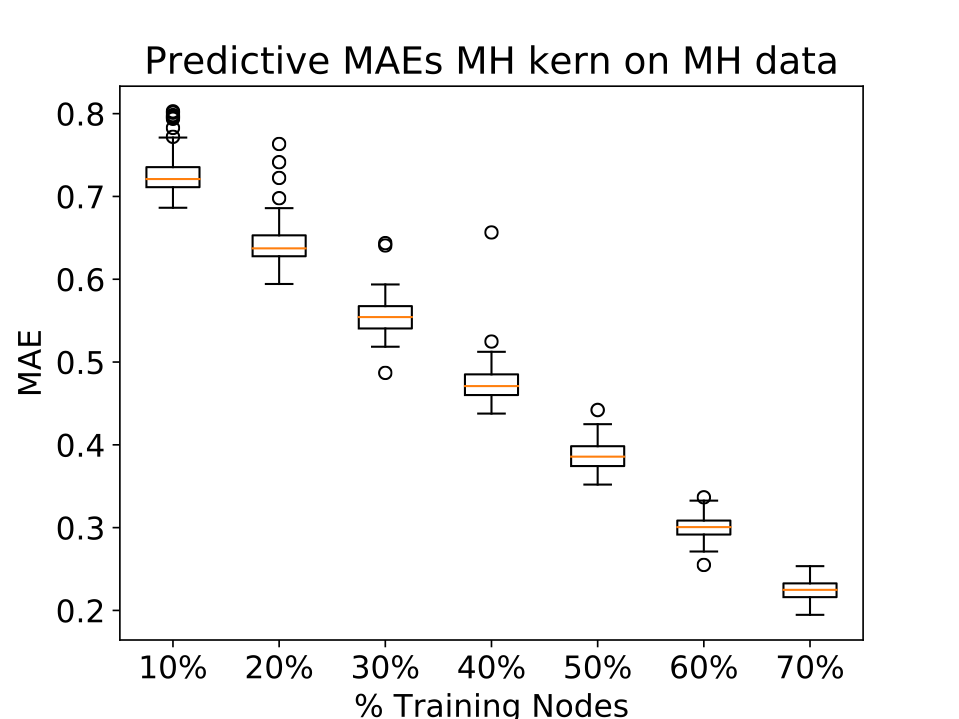}
\caption{}
\end{subfigure}
\begin{subfigure}[b]{0.32\linewidth}
\includegraphics[width=\linewidth, height = 4.cm]{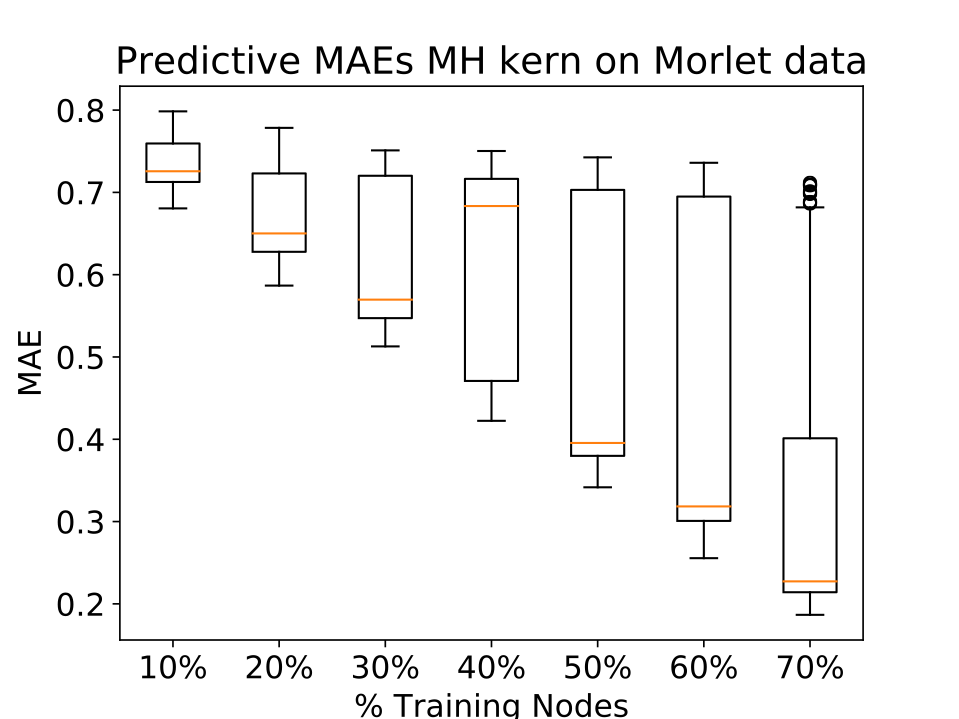}
\caption{}
\end{subfigure}

\begin{subfigure}[b]{0.32\linewidth}
\includegraphics[width=\linewidth, height = 4.cm]{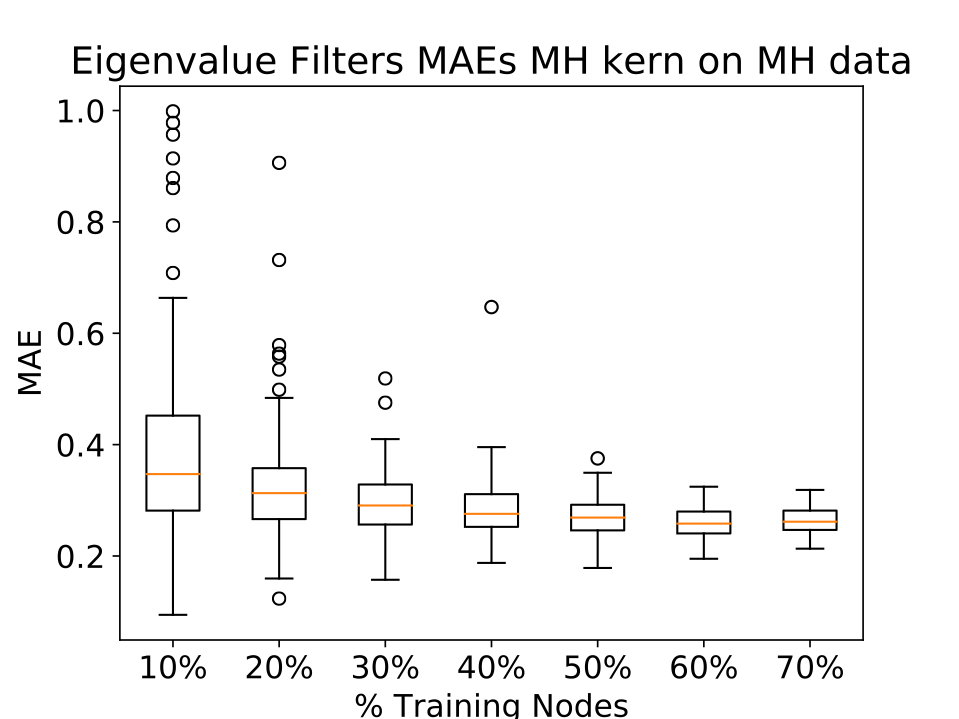}
\caption{}\label{mae2}
\end{subfigure}
\begin{subfigure}[b]{0.32\linewidth}
\includegraphics[width=\linewidth, height = 4.cm]{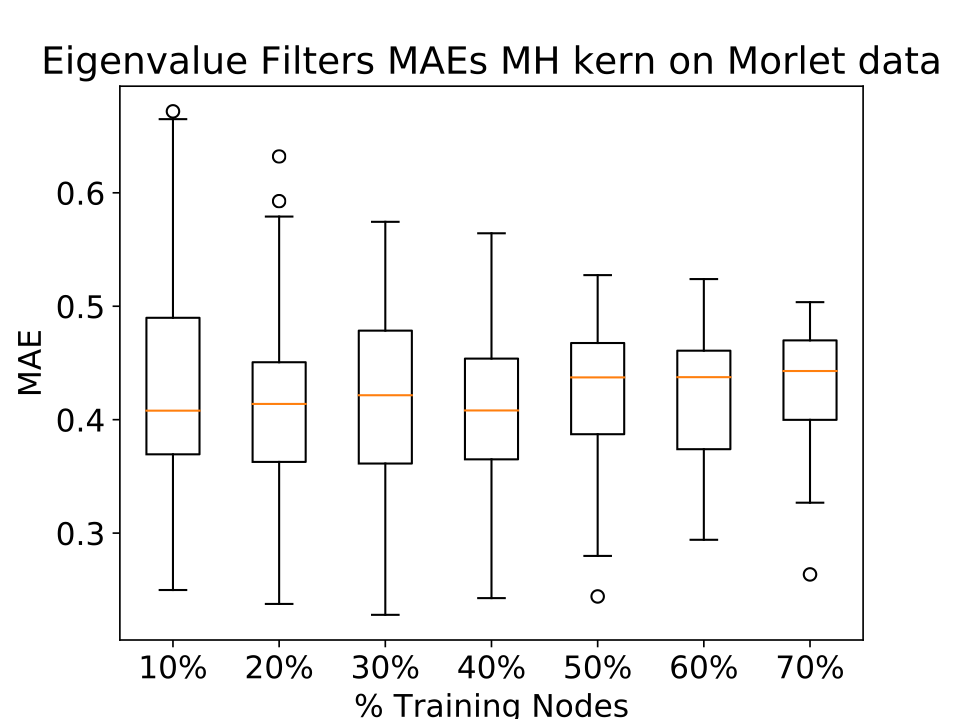}
\caption{}\label{mae2}
\end{subfigure}
\caption{WGGP prediction MAE (Figure 3b from main text) using the Mexican Hat kernel on synthetic data generated using (a) and (c) Mexican Hat wavelet (b) and (d) Morlet wavelet. }\label{maes_ggp_cheby}
\end{figure}

